\documentclass[english]{article}
\usepackage[T1]{fontenc}
\usepackage[latin9]{inputenc}
\usepackage{xcolor}
\usepackage{array}
\usepackage{multirow}
\usepackage{amsmath}
\usepackage{amssymb}
\usepackage{graphicx}

\makeatletter

\providecommand{\tabularnewline}{\\}

\PassOptionsToPackage{numbers, compress}{natbib}

\usepackage[preprint]{neurips_2021}

\usepackage{hyperref}       
\usepackage{url}            
\usepackage{booktabs}       
\usepackage{amsfonts}       
\usepackage{nicefrac}       
\usepackage{microtype}      
\usepackage{xcolor} 

\usepackage{microtype}
\usepackage{graphicx}
\usepackage{subfigure}
\usepackage{mathtools}

\makeatother

\usepackage{babel}
\begin{document}
\renewcommand{\contentsname}{Table of Content for Appendix}

\title{Revisiting the Dataset Bias Problem from a Statistical Perspective}

\author{\textbf{Kien Do, Dung Nguyen, Hung Le, Thao Le, Dang Nguyen, Haripriya
Harikumar, }\\
\textbf{Truyen Tran, Santu Rana, Svetha Venkatesh}\\
Applied Artificial Intelligence Institute (A2I2), Deakin University,
Australia\\
\emph{\{k.do, dung.nguyen, thai.le, thao.le, d.nguyen, h.harikumar,
}\\
\emph{truyen.tran, santu.rana, svetha.venkatesh\}@deakin.edu.au}}

\maketitle
\global\long\def\Expect{\mathbb{E}}
\global\long\def\Real{\mathbb{R}}
\global\long\def\Data{\mathcal{D}}
\global\long\def\Loss{\mathcal{L}}
\global\long\def\Normal{\mathcal{N}}
\global\long\def\softmax{\text{softmax}}
\global\long\def\ELBO{\text{ELBO}}
\global\long\def\argmin#1{\underset{#1}{\text{argmin}}}
\global\long\def\argmax#1{\underset{#1}{\text{argmax}}}
\global\long\def\do{\text{do}}

\begin{abstract}
In this paper, we study the ``dataset bias'' problem from a statistical
standpoint, and identify the main cause of the problem as the strong
correlation between a class attribute $u$ and a non-class attribute
$b$ in the input $x$, represented by $p(u|b)$ differing significantly
from $p(u)$. Since $p(u|b)$ appears as part of the sampling distributions
in the standard maximum log-likelihood (MLL) objective, a model trained
on a biased dataset via MLL inherently incorporates such correlation
into its parameters, leading to poor generalization to unbiased test
data. From this observation, we propose to mitigate dataset bias via
either weighting the objective of each sample $n$ by $\frac{1}{p(u_{n}|b_{n})}$
or sampling that sample with a weight proportional to $\frac{1}{p(u_{n}|b_{n})}$.
While both methods are statistically equivalent, the former proves
more stable and effective in practice. Additionally, we establish
a connection between our debiasing approach and causal reasoning,
reinforcing our method's theoretical foundation. However, when the
bias label is unavailable, computing $p(u|b)$ exactly is difficult.
To overcome this challenge, we propose to approximate $\frac{1}{p(u|b)}$
using a biased classifier trained with ``bias amplification'' losses.
Extensive experiments on various biased datasets demonstrate the superiority
of our method over existing debiasing techniques in most settings,
validating our theoretical analysis.

\end{abstract}
\addtocontents{toc}{\protect\setcounter{tocdepth}{-1}}

\section{Introduction}

In recent years, Deep Neural Networks (DNNs) have achieved remarkable
performance in Computer Vision and Natural Language Processing tasks.
This success can be attributed to their capability of capturing various
patterns in the training data that are indicative of the target class.
However, when the training data exhibits strong correlation between
a non-class attribute and the target class (often referred to as \emph{``dataset
bias''} \cite{Ahn2023,Bahng2020,LeBras2020,Nam2020}), DNNs may overly
rely on the non-class attribute instead of the actual class attribute,
especially if the non-class attribute is easier to learn \cite{Nam2020}.
This leads to biased models that struggle to generalize to new scenarios
where the training bias is absent. For instance, consider a dataset
of human face images where men typically have black hair and women
usually have blond hair. If we train a DNN on this dataset for gender
classification, the model might take a ``shortcut'' and use hair
color (a non-class attribute) as a primary predictor. As a result,
when the model encounters a man with blond hair during testing, it
erroneously predicts the individual as a woman. 

To tackle the dataset bias problem, earlier approaches rely on the
availability of bias labels \cite{Kim2019}. They employ supervised
learning to train a bias prediction model to capture the bias in the
training data, and concurrently learn debiased features that share
the smallest mutual information with the captured bias. The debiased
features are then utilized for predicting the target class. On the
other hand, alternative approaches relax the assumption of bias label
availability and focus on specific types of bias \cite{Bahng2020,wang2019learning}.
They introduce specialized network architectures to capture these
specific types of bias. For example, Bahng et al. \cite{Bahng2020}
leverage convolutional networks with small receptive fields to capture
textural bias in images. However, acquiring human annotations for
bias can be laborious, expensive, and requires expertise in bias identification,
making it challenging in practical scenarios. Furthermore, bias labeling
may not encompass all forms of bias present in the training data,
particularly those that are continuous. As a result, recent approaches
have shifted their attention to settings where no prior knowledge
about bias is available \cite{Ahn2023,Hwang2022,Kim2021,Kim2022,Lee2021,Nam2020}.
Many of these methods exploit knowledge from a ``biased'' model
trained by minimizing a ``bias amplification'' loss \cite{zhang2018generalized}
to effectively mitigate bias \cite{Ahn2023,Lee2021,Nam2020}. They
have achieved significant improvement in bias mitigation, even surpassing
approaches that assume bias labels. However, the heuristic nature
of their bias correction formulas makes it difficult to clearly understand
why these methods perform well in practice.

In this paper, we revisit the dataset bias problem from a statistical
perspective, and present a mathematical representation of this bias,
expressed as either $p(u|b)\neq p(u)$ or $p(b|u)\neq p(b)$ where
$u$, $b$ refer to the \emph{class attribute} and \emph{non-class
(bias) attribute}, respectively. Our representation characterizes
the common understanding of dataset bias as \emph{``high correlation
between the bias attribute and class attribute''} \cite{Bahng2020,Lee2021}.
In addition, we demonstrate that dataset bias arises naturally within
the standard maximum log-likelihood objective as part of the sampling
distribution, alongside the ``imbalance bias''. Building on this
insight, we propose two approaches to mitigate dataset bias: weighting
the loss of each sample $n$ by $\frac{1}{p(u_{n}|b_{n})}$, or sampling
the sample with a weight proportional to $\frac{1}{p(u_{n}|b_{n})}$.
Through empirical analysis, we highlight the distinct behaviors of
these methods, despite their statistical equivalence. Furthermore,
we offer an intriguing perspective on dataset bias as a ``confounding
bias'' in causal reasoning, and theoretically show that our method
actually learns the causal relationship between the target class $y$
and the class attribute $u$ via minimizing an upper bound of the
expected \emph{negative interventional log-likelihood} $\Expect_{n}\left[-\log p_{\theta}(y_{n}|\do(u_{n}))\right]$.
However, accurately computing $\frac{1}{p(u_{n}|b_{n})}$ or $p(u_{n}|b_{n})$
poses a significant challenge when $b_{n}$ is unknown and intertwined
with $u_{n}$ in the input $x_{n}$. To address this issue, we propose
an alternative approach that approximates $p(y_{n}|b_{n})$, a proxy
for $p(u_{n}|b_{n})$, using a biased classifier trained with ``bias
amplification'' losses \cite{Nam2020}. Our intuition is that if
the biased classifier $p_{\psi}(y_{n}|x_{n})$ is properly trained
to use only the bias attribute $b_{n}$ in the input $x_{n}$ for
predicting $y_{n}$, then $p_{\psi}(y_{n}|x_{n})$ can serve as a
reasonable approximation of $p(y_{n}|b_{n})$.

We conduct comprehensive experiments on four popular biased datasets
that encompass various forms of bias: Colored MNIST, Corrupted CIFAR10,
Biased CelebA, and BAR \cite{Nam2020}. Experimental results show
that our method achieves superior bias mitigation results compared
to many existing baselines. This validates the soundness of our theoretical
analysis and demonstrates the effectiveness of our method in mitigating
bias, especially when no bias label is available. Additionally, our
ablation studies reveal surprising alignments between the optimal
configurations of our method and the values indicated by our theoretical
analysis on some simple datasets like Colored MNIST.

\section{A Statistical View of Dataset Bias\label{sec:A-Statistical-View}}

We consider the standard supervised learning problem which involves
learning a classifier $p_{\theta}(y|x)$, parameterized by $\theta$,
that maps an input sample $x$ to the class probability vector. Let
$\Data:=\left\{ (x_{n},y_{n})\right\} _{n=1}^{N}$ denote the training
dataset consisting of $N$ samples. The typical learning strategy
minimizes the expected negative log-likelihood (NLL) of $y$ conditional
on $x$, computed as follows:
\begin{equation}
\Loss_{\theta}^{\text{NLL}}:=\Expect_{p_{\mathcal{D}}(x_{n},y_{n})}\left[-\log p_{\theta}(y_{n}|x_{n})\right]\label{eq:default_xent_y}
\end{equation}
In the above equation, we intentionally include the subscript $n$
to emphasize that $x_{n}$ and $y_{n}$ correspond to a particular
sample $n$ rather than being arbitrary. Without loss of generality,
we assume that each input $x$ consists of two types of attributes:
the \emph{class} attribute (denoted by $u$) and the \emph{non-class}
attribute (denoted by $b$), i.e., $x=(u,b)$\footnote{We use singular nouns for $u$, $b$ for ease of presentation but
we note that $u$/$b$ can represent a set of class/non-class attributes.}. For example, in the ColoredMNIST dataset \cite{Bahng2020,Nam2020},
$u$ represents the digit shape and $b$ represents the background
color. Eq.~\ref{eq:default_xent_y} can be written as:
\begin{align}
\Loss_{\theta}^{\text{NLL}} & =\Expect_{p_{\Data}(x_{n})p_{\Data}(y_{n}|x_{n})}\left[-\log p_{\theta}(y_{n}|x_{n})\right]\label{eq:ERM_1}\\
 & =\Expect_{p_{\Data}(x_{n})}\left[\mathcal{L}_{\theta}^{\text{xent}}(x_{n},y_{n})\right]\label{eq:ERM_2}\\
 & =\Expect_{p(u_{n})p(b_{n}|u_{n})}\left[\mathcal{L}_{\theta}^{\text{xent}}(x_{n},y_{n})\right]\label{eq:ERM_3}\\
 & =\Expect_{p(b_{n})p(u_{n}|b_{n})}\left[\mathcal{L}_{\theta}^{\text{xent}}(x_{n},y_{n})\right]\label{eq:ERM_4}
\end{align}
where $\Loss_{\theta}^{\text{xent}}$ denotes the cross-entropy loss.
Intuitively, if $u$ is distinctive among classes, $u$ can be generally
treated as a categorical random variable. In this case, we will have
$p(u_{n})\approx p(y_{n})$ and $p(b_{n}|u_{n})\approx p(b_{n}|y_{n})$.
From Eq.~\ref{eq:ERM_3}, it is clear that there are two main sources
of bias in the training data. One comes from the non-uniform distribution
of the class attribute (i.e., $p(u)$ is not uniform among classes),
and the other comes from the strong correlation between the non-class
attribute $b$ and the class attribute $u$ (i.e., $p(b|u)$ is very
different from $p(b)$). A highly correlated non-class attribute will
cause the model to depend more on $b$ and less on $u$ to predict
$y$. The former is commonly known as the \emph{``class-imbalance
bias}'', while the latter is often referred to as the \emph{``dataset
bias''} \cite{Ahn2023,Bahng2020,LeBras2020,Nam2020}. In this paper,
we focus exclusively on addressing the dataset bias due to its difficulty,
especially when no prior knowledge about the bias is available. Besides,
we decide to rename the dataset bias as \emph{``feature-correlation
bias''} since we believe this name better characterizes the property
of the bias. We also refer to $b$ as a \emph{bias attribute} because
it is the primary factor contributing to the bias. In the next section,
we will discuss in detail our methods for mitigating the feature-correlation
bias.

\section{Mitigating Dataset Bias from Statistical and Causal Perspectives\label{sec:Mitigating-Dataset-Bias}}

\subsection{Bias mitigation based on $p(u|b)$\label{subsec:Bias-mitigation-based-on-p(u|b)}}

Eq.~\ref{eq:ERM_4} suggests that we can mitigate the dataset bias
(or feature-correlation bias) by either weighting the individual loss
$\mathcal{L}_{\theta}^{\text{xent}}(x_{n},y_{n})$ by $\frac{1}{p(u_{n}|b_{n})}$
during training or sampling each data point $(x_{n},y_{n})$ with
the weight proportional to $\frac{1}{p(u_{n}|b_{n})}$. We refer to
the two techniques as \emph{loss weighting} (LW) and \emph{weighted
sampling} (WS), respectively. The two techniques are statistically
equivalent, and transform the objective in Eq.~\ref{eq:ERM_4} into
$\Expect_{p(b_{n})}\left[\mathcal{L}_{\theta}^{\text{xent}}(x_{n},y_{n})\right]$,
which no longer contains $p(u_{n}|b_{n})$.

We can approximate $p(u_{n}|b_{n})$ using its proxy $p(y_{n}|b_{n})$.
However, explicitly modeling $p(y|b)$ pose challenges due to the
typical unknown nature of $b$. Meanwhile, modeling $p(y|x)$ (or
$p(y|u,b)$) is straightforward. Therefore, we propose to model $p(y|b)$
indirectly through $p(y|x)$ by training a parameterized model $p_{\psi}(y|x)$
in a manner that amplifies the influence of the bias attribute $b$
(in $x$) on $y$. We refer to $p_{\psi}(y|x)$ as the \emph{biased
classifier} and train it for $T_{\text{bias}}$ epochs using a bias
amplification loss. Specifically, we choose the generalized cross-entropy
(GCE) loss $\Loss_{\psi}^{\text{GCE}}(x_{n},y_{n})=\frac{1-{p_{\psi}(y_{n}|x_{n})}^{\tau}}{\tau}$
\cite{zhang2018generalized} where $\tau\in(0,1]$ is a hyperparameter
controlling the degree of amplification. Once $p_{\psi}(y|x)$ has
been trained, we can compute the weight for sample $n$ as follows:
\begin{equation}
w_{n}=\min\left(\frac{1}{p_{\psi}(y_{n}|x_{n})},\gamma\right)\label{eq:weight_clamp}
\end{equation}
where $\gamma>0$ is a clamp hyperparameter that prevents $w_{n}$
from becoming infinite when $p_{\psi}(y_{n}|x_{n})$ is close to 0.
Since $p_{\psi}(y_{n}|x_{n})\in[0,1]$, $w_{n}\in[1,\gamma]$. $w_{n}$
can be considered as an approximation of $\frac{1}{p(u_{n}|b_{n})}$.
For the debiasing purpose, we can train $p_{\theta}(y|x)$ via either
loss weighting (LW) or weighted sampling (WS) with the weight $w_{n}$.
In the case of LW, the debiasing loss becomes:
\begin{align}
\Loss_{\theta}^{\text{LW}} & =\Expect_{p(b_{n})p(u_{n}|b_{n})}\left[\frac{\mathcal{L}_{\theta}^{\text{xent}}(x_{n},y_{n})}{p(u_{n}|b_{n})}\right]\label{eq:LW_1}\\
 & \approx\Expect_{p_{\Data}(x_{n})}\left[w_{n}\cdot\mathcal{L}_{\theta}^{\text{xent}}(x_{n},y_{n})\right]\label{eq:LW_2}
\end{align}

Although LW and WS are statistically equivalent, they perform differently
in practice. LW preserves the diversity of training data but introduces
different scales to the loss. To make training with LW stable, we
rescale $w_{n}$ so that its maximum value is not $\gamma$ (which
could be thousands) but a small constant value, which is 10 in this
work. This means $w_{n}$ lies in the range $\left[\frac{10}{\gamma},10\right]$.
WS, by contrast, maintains a constant scale of the loss but fails
to ensure the diversity of training data due to over/under-sampling.
During our experiment, we observed that LW often yields better performance
than WS, which highlights the importance of data diversity.

However, if we simply fix the sample weight to be $w_{n}$ throughout
training, a classifier trained via LW will take long time to achieve
good results, and the results will be not optimal in some cases. It
is because bias-aligned (BA) samples, which dominates the training
data, have very small weights. The classifier will spend most of the
training time performing very small updates on these BA samples, and
thus, struggles to capture useful information in the training data.
To deal with this problem, we propose a simple yet effective annealing
strategy for LW. We initially set the weights of all training samples
to the same value $\beta$ and linearly transform $\beta$ to $w_{n}$
for $T_{\text{anneal}}$ steps. Mathematically, the weight for sample
$n$ at step $t$ is $w_{n}(t)=\begin{cases}
\beta+\frac{t(w_{n}-\beta)}{T_{\text{anneal}}} & \text{if }0<t<T_{\text{anneal}}\\
w_{n} & \text{otherwise}
\end{cases}$ and the loss for annealed loss weighting (ALW) is given below:
\begin{equation}
\Loss_{\theta}^{\text{ALW}}(t)=\Expect_{p_{\Data}(x_{n})}\left[w_{n}(t)\cdot\Loss_{\theta}^{\text{xent}}(x_{n},y_{n})\right]\label{eq:ALW_1}
\end{equation}

\subsection{Interpretation of $\protect\Loss_{\theta}^{\text{LW}}$ from a causal
perspective}

\begin{figure}
\begin{centering}
\includegraphics[width=0.4\textwidth]{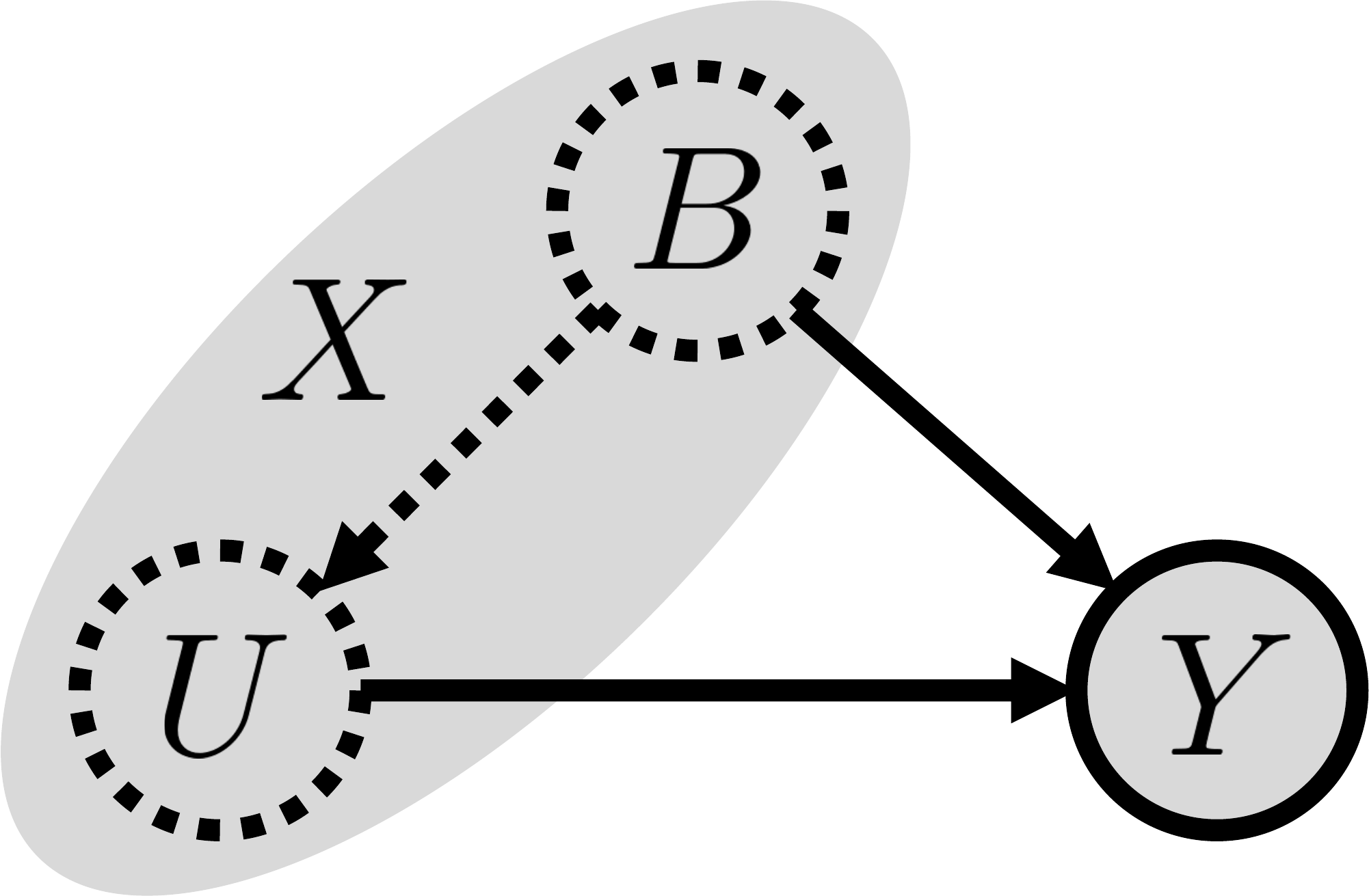}
\par\end{centering}
\caption{A causal graphical model representing the feature-correlation bias
problem. $X$, $Y$ are the input image and class label, respectively.
Both are \emph{observed} (marked with shaded background) during training.
$U$, $B$ are the \emph{hidden} class and non-class attributes of
$X$, respectively.\label{fig:causal-graph}}
\end{figure}

Interestingly, we can interpret the debiasing loss $\Loss_{\theta}^{\text{LW}}$
in the language of causal reasoning by utilizing the Potential Outcomes
framework \cite{rubin1974estimating} illustrated in Fig.~\ref{fig:causal-graph},
where the class label $Y$, class attribute $U$, and non-class attribute
$B$ play the roles of the outcome, treatment, and confounder, respectively.
In our setting, both $U$ and $B$ are hidden but can be accessed
through the observed input $X$. When the unconfoundedness and positivity
assumptions (i.e. the backdoor assumptions) \cite{yao2021survey}
are met, we can estimate the causal quantity $p(y|\do(u))$ from observational
data via backdoor adjustment \cite{pearl2000models} as follows:
\begin{align}
p(y|\do(u)) & =\sum_{b}\frac{p(y,u,b)}{p(u|b)}\\
 & =\Expect_{p(u,b)}\left[\frac{p(y|u,b)}{p(u|b)}\right]\label{eq:p_y_do_u_3}\\
 & =\Expect_{p(b)}\left[p(y|u,b)\right]\label{eq:p_y_do_u_4}
\end{align}

Eq.~\ref{eq:p_y_do_u_3} is typically known as Inverse Probability
Weighting (IPW), where $p(u|b)$ is called the propensity score \cite{hirano2003efficient,hirano2001estimation}.
In Eq.~\ref{eq:p_y_do_u_3}, the class prediction $p(y|u,b)$ is
weighted by the inverse propensity score $\frac{1}{p(u|b)}$, exhibiting
a degree of resemblance to our loss $\Loss_{\theta}^{\text{LW}}$
in Eq.~\ref{eq:LW_1}. It suggests that we can interpret $\Loss_{\theta}^{\text{LW}}$
from a causal standpoint. In fact, $\Loss_{\theta}^{\text{LW}}$ acts
as an upper bound of the expected \emph{negative interventional log-likelihood}
(NILL) $\Loss_{\theta}^{\text{NILL}}=\Expect_{n}\left[-\log p_{\theta}(y_{n}|\do(u_{n}))\right]$.
The relationship between these two losses is provided below:
\begin{align}
\Loss_{\theta}^{\text{NILL}} & =\Expect_{n}\left[-\log p_{\theta}(y_{n}|\do(u_{n}))\right]\label{eq:NILL_LW_1}\\
 & \leq\Expect_{p(u_{n},b_{n},y_{n})}\left[\frac{-\log p_{\theta}(y_{n}|u_{n},b_{n})}{p(u_{n}|b_{n})}\right]\label{eq:NILL_LW_2}\\
 & =\Expect_{p(u_{n},b_{n})}\left[\frac{\Loss_{\theta}^{\text{xent}}(x_{n},y_{n})}{p(u_{n}|b_{n})}\right]=\Loss_{\theta}^{\text{LW}}\label{eq:NILL_LW_3}
\end{align}
where the inequality in Eq.~\ref{eq:NILL_LW_2} is derived from:
\begin{align}
-\log p_{\theta}(y|\do(u)) & =-\log\Expect_{p(b)}\left[p_{\theta}(y|u,b)\right]\\
 & \leq-\Expect_{p(b)}\left[\log p_{\theta}(y|u,b)\right]\label{eq:log_p_y_do_u_2}\\
 & =\Expect_{p(u,b)}\left[\frac{-\log p_{\theta}(y|u,b)}{p(u|b)}\right]\label{eq:log_p_y_do_u_3}
\end{align}

Eq.~\ref{eq:log_p_y_do_u_2} is the Jensen inequality with equality
attained when $p_{\theta}(y|u,b)=p_{\theta}(y|u,b')$ $\forall$ $b,b'$,
i.e. $y$ is independent of $b$ given $u$. This condition matches
our target of learning an unbiased classifier $p_{\theta}(y|u,b)$.
In Eq.~\ref{eq:log_p_y_do_u_3}, $p(u|b)$ is introduced to allow
$(u,b)$ to be sampled jointly from observational data. Eq.~\ref{eq:NILL_LW_2}
can be viewed as a Monte Carlo estimation of Eq.~\ref{eq:log_p_y_do_u_3}
(or Eq.~\ref{eq:log_p_y_do_u_2}) using a single sample of $b$,
i.e. $b_{n}$. From Eqs.~\ref{eq:NILL_LW_1} - \ref{eq:log_p_y_do_u_3},
we see that minimizing $\Loss_{\theta}^{\text{LW}}$ also minimizes
$\Loss_{\theta}^{\text{NILL}}$ and encourages $p_{\theta}(y|u,b)$
to be close to $p_{\theta}(y|u)$. Minimizing $\Loss_{\theta}^{\text{NILL}}$
causes the model to focus more (less) on uncommon (common) samples
$n$ which has small (big) $p(u_{n}|b_{n})$.

\subsection{Bias mitigation based on $p(b|u)$}

Eq.~\ref{eq:ERM_3} suggests an alternative approach to mitigating
the feature-correlation bias, which involves weighting each individual
sample $n$ by $\frac{1}{p(b_{n}|u_{n})}$ rather than $\frac{1}{p(u_{n}|b_{n})}$.
However, accurately estimating $p(b|u)$ poses challenges in practical
implementations. In Appdx.~\ref{subsec:Bias-mitigation-based-on-p(u|b)},
we present an idea about using conditional generative models to approximate
$p(b|u)$ and discuss its limitations.

\section{Related Work}

A plethora of techniques for mitigating bias are present in the literature.
However, in this paper's context, we focus on the most relevant and
recent methods, leaving the discussion of other approaches in Appdx.~\ref{sec:More-discussion-about-related-work}.
These methods can be broadly categorized into two groups: i) those
that utilize the bias label or prior knowledge about bias, and ii)
those that do not. The two groups are discussed in detail below.

\paragraph{Debiasing given the bias label or certain types of bias}

When the bias label is available, a straightforward approach is to
train a bias prediction network or a \emph{``biased''} network in
a supervised manner. The bias knowledge acquired from the biased network
can then be utilized as a form of regularization to train another
\emph{``debiased''} network. One commonly used regularization strategy
involves minimizing the mutual information between the biased and
debiased networks through adversarial training \cite{Bahng2020,Kim2019,Zhu2021}.
This compels the debiased network to learn features independent of
the bias information, which are considered unbiased. In the model
proposed by \cite{Kim2019}, a ``biased'' head is positioned on
top of a ``debiased'' backbone with a gradient reversal layer \cite{ganin2016domain}
in between to facilitate adversarial learning. The backbone is trained
to trick the biased head into predicting incorrect bias labels while
the biased head attempts to make correct bias predictions. Other works,
such as \cite{Bahng2020,wang2019learning}, do not make use of the
bias label; rather, they assume that image texture is the main source
of bias. This comes from the observation that outputs of deep neural
networks depend heavily on superficial statistics of the input image
such as texture \cite{geirhos2018imagenet}. The framework in \cite{wang2019learning}
consists of two branches: a conventional CNN for encoding visual features
from the input image, and a set of learnable gray-level co-occurrence
matrices (GLCMs) for extracting the textural bias information. Besides
adversarial regularization, the authors of \cite{wang2019learning}
introduce another regularization technique known as HEX, which projects
the CNN features into a hidden space so that the projected vectors
contain minimal information about the texture bias captured by the
GLCM branch. \cite{Bahng2020}, on the other hand, use a CNN with
small receptive fields as the biased network, and the Hilbert-Schmidt
Independence Criterion (HSIC) as a measure of mutual information between
the biased and debiased classifiers' features. \cite{Hong2021} draw
a probabilistic connection between data generated by $p(y)p(b)$ and
by $p(y|b)p(b)$, and utilize the assumption that $p(x|y,b)$ remains
unchanged regardless of the change in $p(y,b)$ to derive an effective
bias correction method called BiasBal. They also propose BiasCon -
a debiasing method based on contrastive learning. In the case the
bias label is not provided, they assume the bias is texture and make
use of the biased network proposed in \cite{Bahng2020}. EnD \cite{Tartaglione2021}
employs a regularization loss for debiasing that comprises two terms:
a ``disentangling'' term that promotes the decorrelation of samples
with similar bias labels, and an ``entangling'' term which forces
samples belonging to the same class but having different bias labels
to be correlated. EnD exhibits certain similarities to BiasCon \cite{Hong2021},
as the ``entangling'' and ``disentangling'' terms can be viewed
as the positive and negative components of a contrastive loss, respectively.

In visual question answering (VQA), bias can arise from the co-occurrence
of words in the question and answer, causing the model to overlook
visual cues when making predictions \cite{agrawal2016analyzing,agrawal2018don}.
To overcome this bias, common approaches involve training a biased
network that takes only questions as input to predict answers. The
prediction from this biased network is then used to modulate the prediction
of a debiased network trained on both questions and images \cite{Cadene2019,clark2019don,ramakrishnan2018overcoming}.

\begin{table*}
\begin{centering}
\resizebox{\textwidth}{!}{
\begin{tabular}{>{\raggedright}p{0.08\textwidth}cccccccc}
\hline 
Dataset & BC (\%) & Vanilla & ReBias & LfF & DFA & SelecMix & PGD & LW (Ours)\tabularnewline
\hline 
\hline 
\multirow{3}{0.08\textwidth}{Colored MNIST} & 0.5 & 80.18$\pm$1.38 & 74.85$\pm$1.97 & 93.38$\pm$0.52 & 91.85$\pm$0.92 & 83.41$\pm$1.26 & \textbf{96.15$\pm$0.28} & \textcolor{gray}{95.57$\pm$0.41}\tabularnewline
 & 1.0 & 87.48$\pm$1.75 & 84.23$\pm$1.56 & 94.09$\pm$0.78 & 94.32$\pm$0.89 & 91.59$\pm$0.99 & \textbf{97.93$\pm$0.19} & \textcolor{gray}{97.18$\pm$0.34}\tabularnewline
 & 5.0 & 97.04$\pm$0.21 & 95.76$\pm$0.50 & 97.40$\pm$0.25 & 96.74$\pm$0.43 & 97.37$\pm$0.15 & \textbf{98.74$\pm$0.12} & \textcolor{gray}{98.61$\pm$0.09}\tabularnewline
\hline 
\multirow{3}{0.08\textwidth}{Corrupted CIFAR10} & 0.5 & 28.00$\pm$1.15 & - & 41.95$\pm$1.56 & 40.54$\pm$1.98 & 31.67$\pm$0.90 & \textcolor{gray}{44.89$\pm$1.36} & \textbf{45.76$\pm$1.49}\tabularnewline
 & 1.0 & 34.56$\pm$0.87 & - & \textbf{53.36$\pm$1.87} & 50.27$\pm$0.94 & 36.28$\pm$1.22 & 47.38$\pm$1.01 & \textcolor{gray}{51.64$\pm$1.12}\tabularnewline
 & 5.0 & 59.33$\pm$1.26 & - & \textcolor{gray}{70.04$\pm$1.05} & 67.05$\pm$1.82 & 63.15$\pm$1.17 & 63.60$\pm$0.58 & \textbf{70.45$\pm$1.24}\tabularnewline
\hline 
\end{tabular}}
\par\end{centering}
\caption{Results of different debiasing methods on Colored MNIST and Corrupted
CIFAR10 when the bias label is unavailable. The best and second best
results are highlighted in bold and gray, respectively.\label{tab:Main-Results-MNIST-Cifar10}}
\end{table*}

\paragraph{Debiasing without prior knowledge about bias}

Due to the challenges associated with identifying and annotating bias
in real-world scenarios, recent attention has shifted towards methods
that do not rely on bias labels or make assumptions about specific
types of bias. LfF \cite{Nam2020} is a pioneering method in this
regard. It utilizes the GCE loss \cite{zhang2018generalized}, which
is capable of amplifying the bias in the input, to train the biased
classifier, thereby eliminating the need for bias labels. This strategy
has been inherited and extended in numerous subsequent works \cite{Ahn2023,Hwang2022,Kim2021,Lee2021,Lee2023}.
\cite{Lee2021} emphasize the importance of diversity in bias mitigation,
and propose a method that augments the training data by swapping the
bias features of two samples, as extracted by the biased classifier.
BiaSwap \cite{Kim2021}, on the other hand, leverages SwapAE \cite{Park2020}
and CAM \cite{zhou2016learning} to generate ``bias-swapped'' images.
SelecMix \cite{Hwang2022} applies mixup on ``contradicting'' pairs
of samples (i.e., those having the same label but far away in the
latent space, or different labels but close), and uses the mixed-up
samples for training the debiased classifier. PGD \cite{Ahn2023}
uses the biased network's gradient to compute the resampling weight.
\cite{Lee2023} aim to improve the biased classifier by training it
using bias-aligned samples only. LWBC \cite{Kim2022} trains a ``biased
committee'' - a group of multiple biased classifiers - using the
cross-entropy loss and knowledge distilled from the main classifier
trained in parallel. Outputs from the biased classifiers are used
to compute the sample weights for training the main classifier. \cite{Shrestha2022}
conduct an extensive empirical study about some existing bias mitigation
methods, and discover that many of them are sensitive to hyperparameter
tuning. Based on their findings, they suggest to adopt more rigorous
assessments.

\section{Experiments}

\subsection{Experimental Setup}

\begin{table}
\begin{centering}
\begin{tabular}{>{\raggedright}p{0.08\textwidth}ccccc}
\hline 
Dataset & BC (\%) & Vanilla & LfF & PGD & LW (Ours)\tabularnewline
\hline 
\hline 
\multirow{3}{0.08\textwidth}{Biased CelebA} & 0.5 & 77.43$\pm$0.42 & 77.81$\pm$1.01 & \textcolor{gray}{78.07$\pm$2.18} & \textbf{87.54$\pm$0.32}\tabularnewline
 & 1.0 & 80.58$\pm$0.41 & \textcolor{gray}{85.54$\pm$1.27} & 79.26$\pm$0.88 & \textbf{86.38$\pm$0.37}\tabularnewline
 & 5.0 & \textcolor{gray}{86.35$\pm$0.33} & 80.22$\pm$1.58 & 83.47$\pm$0.95 & \textbf{87.43$\pm$0.34}\tabularnewline
\hline 
BAR & - & 68.45$\pm$0.32 & 62.09$\pm$0.21 & \textcolor{gray}{70.49$\pm$0.65} & \textbf{71.24$\pm$0.53}\tabularnewline
\hline 
\end{tabular}
\par\end{centering}
\caption{Results of different debiasing methods on Biased CelebA and BAR when
the bias label is unavailable. The best and second best results are
highlighted in bold and gray, respectively.\label{tab:Results-on-CelebA-and-BAR}}
\end{table}

\subsubsection{Datasets}

We evaluate our proposed methods on 4 popular datasets for bias correction,
namely Colored MNIST \cite{Bahng2020,Nam2020}, Corrupted CIFAR10
\cite{Nam2020}, Biased CelebA \cite{Nam2020,sagawa2020distributionally},
and BAR \cite{Nam2020}. Details about these datasets are provided
below and in Appdx.~\ref{subsec:Datasets}.

In Colored MNIST, the target attribute is the digit, while the bias
attribute is the background color. In Corrupted CIFAR10, the target
attribute is the object, and the bias attribute is the corruption
noise. We created Colored MNIST and Corrupted CIFAR10 from the standard
MNIST \cite{lecun2010mnist} and CIFAR10 \cite{krizhevsky2009learning}
datasets respectively using the official code provided by the authors
of \cite{Nam2020} with some slight modifications. Specifically, we
used distinctive background colors for Colored MNIST, and set the
severity of the corruption noise to 2 for Corrupted CIFAR10 to retain
enough semantic information for the main task. Following \cite{Nam2020},
we created 3 versions of Colored MNIST and Corrupted CIFAR10 with
3 different bias-conflicting ratios (BC ratios) which are 0.5\%, 1\%,
and 5\%. 

In Biased CelebA, the hair color (\emph{blond(e)} or \emph{not blond(e)})
serves as the target attribute, while the gender (\emph{male} or \emph{female})
is considered the bias attribute. Individuals with blond(e) hair exhibit
a bias toward being female, whereas those without blond(e) hair are
biased toward being male. We created Biased CelebA ourselves by selecting
a random subset of the original training samples from the CelebA dataset
\cite{liu2015faceattributes} to ensure a certain BC ratio is achieved.
We consider 3 BC ratios of 0.5\%, 1\%, and 5\%. Each BC ratio is associated
with a specific number of BC samples per target class, which is 100,
200, and 500, respectively. As a result, the training set for Biased
CelebA comprises a total of 39998, 39998, and 19998 samples for the
BC ratios of 0.5\%, 1\%, and 5\%, respectively.

BAR is a dataset for action recognition which consists of 6 action
classes, namely \emph{climbing}, \emph{diving}, \emph{fishing}, \emph{racing},
\emph{throwing}, and \emph{vaulting}. The bias in this dataset is
the place where the action is performed. For example, climbing is
usually performed on rocky mountains, or diving is typically practiced
under water. This dataset does not have bias labels. We use the default
train/valid/test splits provided by the authors \cite{Nam2020}. 

\subsubsection{Baselines}

We conduct a comprehensive comparison of our method with popular and
up-to-date baselines for bias correction \cite{Ahn2023,Bahng2020,Hong2021,Hwang2022,Lee2021,Nam2020}.
The selected baselines encompass a diverse range of approaches including
information-theoretic-based methods \cite{Bahng2020}, loss weighting
techniques \cite{Nam2020}, weighted sampling strategies \cite{Ahn2023},
mix-up approaches \cite{Hwang2022}, and BC samples synthesis methods
\cite{Lee2021}. Some of them \cite{Nam2020,Ahn2023} are closely
related to our methods, and will receive in-depth analysis. To establish
a fair playing ground, we employ \emph{identical} classifier architectures,
data augmentations, optimizers, and learning rate schedules for both
our methods and the baselines. We also search for the learning rates
that lead to the best performances of the baselines. For other hyperparameters
of the baselines, we primarily adhere to the default settings outlined
in the original papers. Details about these settings are provided
below and in Appdx\@.~\ref{sec:More-Details-about-Training-Settings}.

\subsubsection{Implementation details\label{subsec:Implementation-details}}

We implement the classifier using a simple convolutional neural network
(CNN) for Colored MNIST, a small ResNet18 \footnote{https://github.com/kuangliu/pytorch-cifar}
for Corrupted CIFAR10, and the standard ResNet18 \cite{he2016deep}
for Biased CelebA and BAR. The CNN used for Colored MNIST is adapted
from the code provided in \cite{Ahn2023}. The standard ResNet18 architecture
is sourced from the torchvision library. Given that the input size
for Biased CelebA is 128$\times$128, we simply replace the first
convolution layer of the standard ResNet18, which originally has a
stride of 2, with another convolution layer having a stride of 1.

Following \cite{Hong2021,Hwang2022,Lee2021,Nam2020}, we augment the
input image with random horizontal flip, random crop, and random resized
crop, depending on the dataset (details in Appdx.~\ref{subsec:Data-augmentation}).
Unlike \cite{Ahn2023}, we choose \emph{not} to employ color jitter
as a data augmentation technique. This deliberate decision is based
on the understanding that such augmentation has the potential to eliminate
specific types of bias present in the input image, thereby bolstering
the classifier's robustness without necessitating any additional bias
mitigation techniques. Consequently, it becomes challenging to ascertain
whether the observed performance improvements of a bias mitigation
method genuinely stem from its inherent capabilities or simply result
from the applied augmentation, especially on datasets having color
bias like Colored MNIST.

\begin{figure*}
\begin{centering}
\resizebox{\textwidth}{!}{%
\par\end{centering}
\begin{centering}
\begin{tabular}{ccc}
\multicolumn{3}{c}{\includegraphics[height=0.06\textwidth]{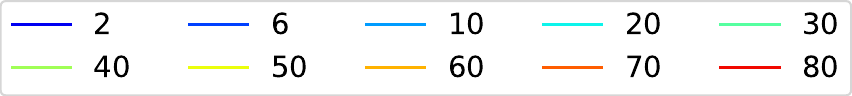}}\tabularnewline
\includegraphics[width=0.4\textwidth]{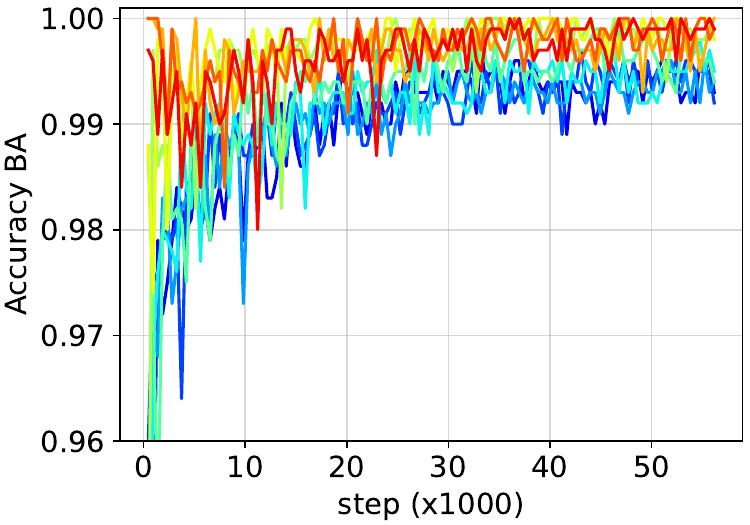} & \includegraphics[width=0.4\textwidth]{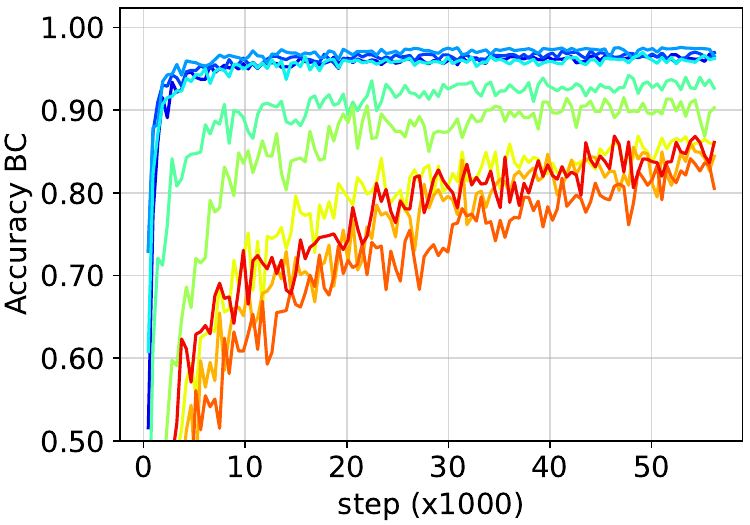} & \includegraphics[width=0.4\textwidth]{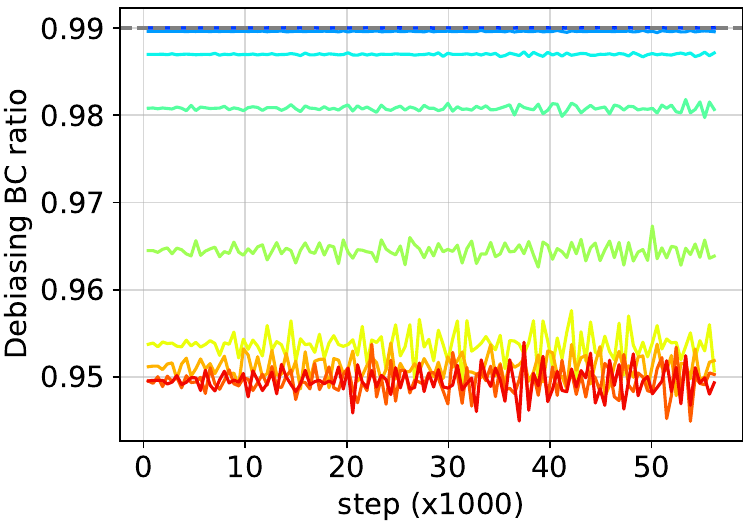}\tabularnewline
(a) Test accuracy (BA) & (b) Test accuracy (BC) & (c) Debiasing BC ratio\tabularnewline
\end{tabular}}
\par\end{centering}
\caption{Learning curves of LW w.r.t. different training epochs of the biased
classifier ($T_{\text{bias}}$), ranging from 2 to 80. The biased
dataset is Colored MNIST with the BC ratio of 1\%. The maximum sample
weight is set to 100. Since BC samples account for 90\% of the total
test samples, the test accuracies for all samples are very similar
to those for BC sample in (b).\label{fig:biased-epoch-main-results}}
\end{figure*}

We provide details for the optimizer, training epochs, learning rate,
learning rate schedule, etc. corresponding to each dataset in Appdx.~\ref{subsec:Optimization-hyperparameters}.

\subsection{Results when the bias label is available}

Due to space constraint, we provide a detailed examination of our
method's performance in cases where the bias label is accessible in
Appdx.~\ref{subsec:Results-when-the-bias-is-available}. To summarize,
both WS and LW substantially outperform the vanilla classifier across
Colored MNIST, Corrupted CIFAR10, and Biased CelebA, affirming the
validity of our theoretical analysis in Section~\ref{sec:Mitigating-Dataset-Bias}.
Furthermore, our results indicate that LW consistently surpasses WS
on the three datasets. Therefore, in the subsequent experiments, we
will exclusively focus on LW.

\subsection{Results when the bias label is unavailable}

\subsubsection{Results on Colored MNIST and Corrupted CIFAR10\label{subsec:Results-on-MNIST-and-CIFAR10}}

As shown in Table~\ref{tab:Main-Results-MNIST-Cifar10}, our proposed
method LW significantly outperforms the vanilla classifier trained
with the standard cross-entropy loss, as well as several other debiasing
baselines such as ReBias \cite{Bahng2020}, DFA \cite{Lee2021}, and
SelecMix \cite{Hwang2022}. Furthermore, LW achieves higher test accuracies
than LfF in most settings of Colored MNIST and Corrupted CIFAR10.
Compared to the current state-of-the-art debiasing method PGD, LW
performs slightly worse on Colored MNIST but demonstrates superior
performance on Corrupted CIFAR10. Specifically, LW achieves about
1\%, 4\%, and 7\% higher accuracy than PGD on Corrupted CIFAR10 with
BC ratios of 0.5\%, 1\%, and 5\%, respectively. Detailed comparisons
between our method and LfF and PGD are provided in Appds.~\ref{subsec:Additional-results-of-PGD}
and \ref{subsec:Additional-results-of-LfF}, respectively. These outcomes
substantiate the efficacy of using the GCE loss to train a model of
$p(y|x)$ that approximates $p(y|b)$, and endorse the practice of
weighting each sample with $\frac{1}{p(y|b)}$ in our method.

\subsubsection{Results on Biased CelebA and BAR\label{subsec:Results-on-CelebA-and-BAR}}

In this experiment, we only choose LfF and PGD as our baselines since
the two methods have demonstrated superior performances compared to
other approaches in our previous experiment and also in \cite{Ahn2023}.
From Table~\ref{tab:Results-on-CelebA-and-BAR}, it is clear that
LW outperforms LfF and PGD significantly on both Biased CelebA and
BAR. Surprisingly, our experiments have revealed that in certain settings,
LfF and PGD may perform even worse than the vanilla classifier, particularly
on Biased CelebA with BC ratios of 1\% and 5\%. Additionally, we have
noticed that LfF and PGD exhibit sensitivity to hyperparameters on
Biased CelebA, as evidenced by the large standard deviations in their
results. These findings raise concerns about the effectiveness of
the debiasing formulas employed by LfF and PGD, which rely heavily
on heuristics. 

It is worth noting that on BAR, our implementation of the vanilla
classifier achieves a higher test accuracy than what has been reported
in \cite{Nam2020} and \cite{Ahn2023}. This implies that the improvement
of LW over the vanilla classifier on BAR can be attributed to its
debiasing capability rather than the under-performance of the vanilla
classifier. As far as we know, the result of LW on BAR presented in
Table~\ref{tab:Results-on-CelebA-and-BAR} currently represents the
state-of-the-art performance in this domain.

\subsection{Ablation Study}

In this section, we closely examine two key hyperparameters that mainly
influence the performance of our proposed LW. They are the number
of training epochs for the biased classifier ($T_{\text{bias}}$),
and the maximum sample weight ($\gamma$).

\begin{figure*}
\begin{centering}
\resizebox{\textwidth}{!}{%
\par\end{centering}
\begin{centering}
\begin{tabular}{ccc}
\multicolumn{3}{c}{\includegraphics[height=0.06\textwidth]{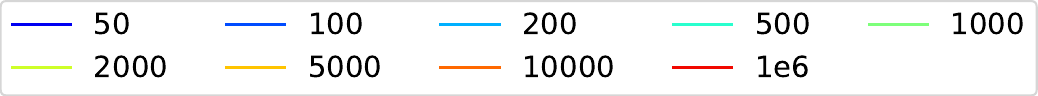}}\tabularnewline
\includegraphics[width=0.4\textwidth]{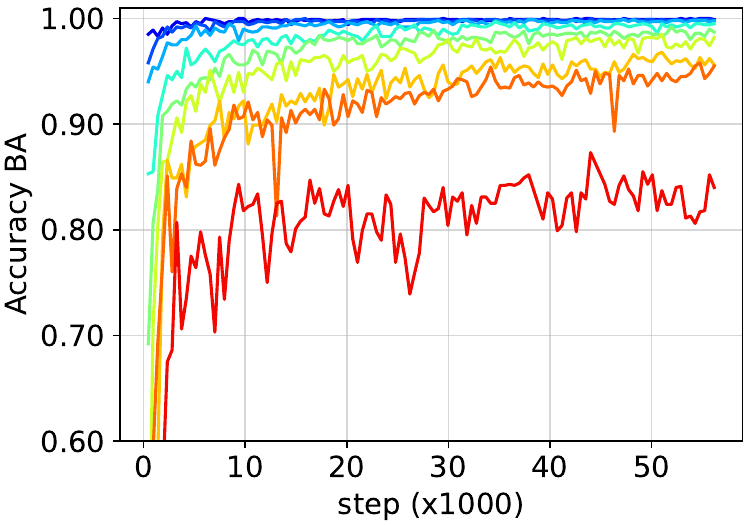} & \includegraphics[width=0.4\textwidth]{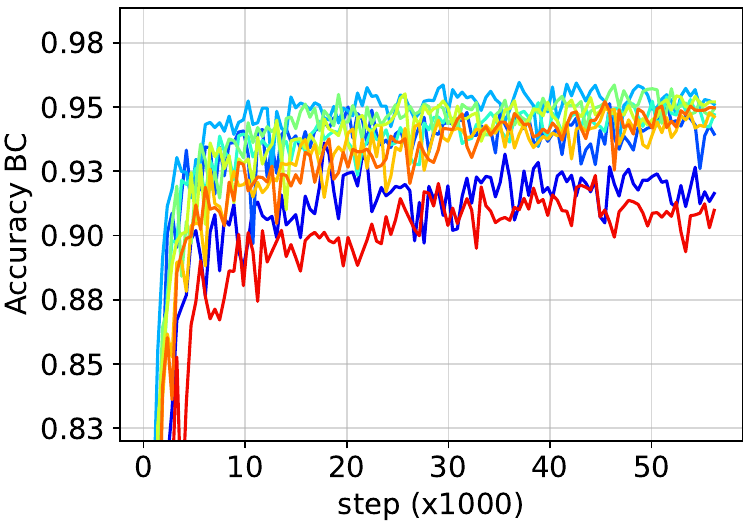} & \includegraphics[width=0.4\textwidth]{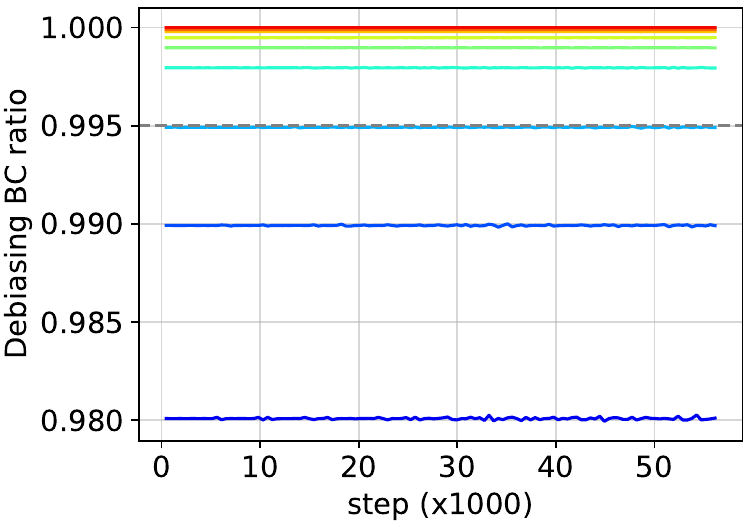}\tabularnewline
(a) Test accuracy (BA) & (b) Test accuracy (BC) & (c) Debiasing BC ratio\tabularnewline
\end{tabular}}
\par\end{centering}
\caption{Learning curves of LW w.r.t. different values of the maximum sample
weight ($\gamma$), ranging from 50 to $10^{6}$. The biased dataset
is Colored MNIST with the BC ratio of 0.5\%. The biased classifier
is trained for 10 epochs. Since BC samples account for 90\% of the
total test samples, the test accuracies for all samples are very similar
to those for BC sample in (b).\label{fig:max-sample-weights-main-results}}
\end{figure*}

\subsubsection{Effect of the number of training epochs for the biased classifier\label{subsec:Effect-of-the-biased-epoch}}

Since the primary role of the biased classifier in LW is to capture
solely the bias attribute $b$ in the input $x$, it is critical that
the biased classifier should not be trained for too long. Otherwise,
this classifier may capture the class attribute $u$, leading to an
inaccurate approximation of $p(y|b)$ by $p_{\psi}(y|x)$. To validate
this intuition, we compare the performances of various versions of
LW in Fig.~\ref{fig:biased-epoch-main-results}, each utilizing a
biased classifier trained for different numbers of epochs $T_{\text{bias}}$.
Apparently, as $T_{\text{bias}}$ decreases, LW achieves higher test
accuracies on BC samples (Fig.~\ref{fig:biased-epoch-main-results}b),
indicating that $p_{\psi}(y|x)$ becomes a more reliable approximation
of $p(y|b)$. However, if $T_{\text{bias}}$ is too small (e.g., $T_{\text{bias}}$
< 10), the biased classifier may not have undergone sufficient training
to produce accurate class predictions, potentially hurting LW's performance.
In fact, the choice of $T_{\text{bias}}$ hinges on the learning capability
of the biased classifier, which is examined in the appendix. Further
empirical results of $T_{\text{bias}}$ can be found in the appendix. 

It is worth noting that there is a strong correlation between the
high test accuracy of LW and the proximity of its \emph{``debiasing
BC ratio''} $\beta$ to the ground-truth BA ratio (Fig.~\ref{fig:biased-epoch-main-results}c).
The debiasing BC ratio is computed by dividing the average loss weight
over BC samples by the sum of the average loss weight over BC samples
and the average loss weight over BA samples. Mathematically, $\beta=\frac{\Expect_{n\in\text{BC}}[w_{n}]}{\Expect_{n\in\text{BC}}[w_{n}]+\Expect_{n\in\text{BA}}[w_{n}]}$.
Interestingly, in this particular setting, LW attains the highest
test accuracy when \emph{its debiasing BC ratio coincides with the
ground-truth BA ratio}, which is 0.99. This remarkable alignment
further corroborates our theoretical analysis, and demonstrates that
the hidden true bias term $p(y|b)$ can be effectively approximated
via a biased classifier trained with a limited number of epochs.

\subsubsection{Effect of the maximum sample weight\label{subsec:Effect-of-the-max-sample-weight}}

The optimal value of the maximum sample weight $\gamma$ plays a crucial
role in ensuring the proper normalization of sample weights $w_{n}$
and achieving a closer match with the true debiasing terms $\frac{1}{p(y_{n}|b_{n})}$.
Generally, the choice of $\gamma$ depends primarily on the bias ratio
present in the training data. To illustrate this, let's consider the
Colored MNIST dataset with a BA ratio of 0.995 and a BC ratio of 0.005.
In this scenario, we would expect that BC samples are weighted approximately
200 times more than BA samples to achieve full debiasing (since $\frac{0.995}{0.005}$
$\approx$ 200). This implies that if BA samples have a weight of
1, BC samples should have a weight of 200. Assuming we have a perfect
biased classifier which assigns a class confidence of 1 to BA samples
and very low class confidences to BC samples (i.e., $p_{\psi}(y|x)$
$\approx$ 1 for BA samples and small for BC samples)\footnote{Such a perfect biased classifier can be attainable in practice if
is trained with an appropriate number of epochs.}, we can compute the weights for BA samples as 1 and for BC samples
as very large weights, which are then clamped at $\gamma$. By setting
$\gamma$ to 200, we obtain the appropriate weights for BC samples,
allowing for full debiasing as discussed earlier. Empirical validation
of this reasoning is presented in Fig.~\ref{fig:max-sample-weights-main-results}b,
where LW achieves the highest test accuracy on BC samples when $\gamma$
is equal to 200. Moreover, at this specific value of $\gamma$, the
debiasing BC ratio of LW aligns with the ground-truth BA ratio of
0.995 (Fig.~\ref{fig:max-sample-weights-main-results}c), indicating
that LW likely achieves full debiasing. However, it should be noted
that when the bias classifier is not perfect, the optimal value of
$\gamma$ would be different from the ground-truth $\frac{\text{BA ratio}}{\text{BC ratio}}$.
Typically, a larger value of $\gamma$ would be used to assign more
weight to BC samples predicted with high class confidences by the
imperfect biased classifier. Regarding the test accuracy on BA samples,
it monotonically decreases as $\gamma$ increases due to the widening
gap between the weights of BC samples and BA samples (Fig.~\ref{fig:max-sample-weights-main-results}a).
For more results of $\gamma$ on other datasets, please refer to the
appendix.

\section{Conclusion}

We have proposed a novel method for mitigating dataset bias and analyzed
our method from statistical and causal perspectives. While our approach
yields promising results, certain limitations persist: i) our method
depends on the training epochs of the biased classifier ($T_{\text{bias}}$)
and the clamp threshold ($\gamma$) to produce good approximations
of $p(y|b)$, which are hard to control in practice due to the variability
introduced by the unknown bias rate; and ii) our method mitigates
bias via balancing the sampling distribution during learning rather
than directly adjusting the target. Nevertheless, these limitations
are not unique to our method and are shared by other debiasing techniques.
For instance, PGD's dependence on $T_{\text{bias}}$ is also evident
(in the appendix). Additionally, both PGD and LfF employ indirect
bias mitigation like ours. In forthcoming work, we aim to develop
a novel method to address the aforementioned limitations.

\bibliographystyle{plain}

\addtocontents{toc}{\protect\setcounter{tocdepth}{2}}

\clearpage{}

\appendix
\tableofcontents{}

\section{More discussion about related work\label{sec:More-discussion-about-related-work}}

\paragraph{Early works on dataset bias mitigation}

The presence of bias in the training data has long been recognized
and acknowledged due to its negative impacts on the generalizability
of practical machine learning systems. Earlier works, such as \cite{Khosla2012,Torralba2011}
mention \emph{``dataset bias''} and view it as the phenomenon where
a model performs well on one dataset but poorly on others. They provide
some explanations for this phenomenon, such as, bias may come from
the process of data selection, capturing, or labeling \cite{Torralba2011}.
They also propose a simple solution to reduce bias by learning an
SVM on multiple datasets, with additional regularization on the weights
that capture various forms of bias across those datasets \cite{Khosla2012}.
Later, Alvi et al. \cite{Alvi2018} introduce a debiased method that
also leverages multiple datasets. Their model has a deep convolutional
neural network backbone for feature extraction, a primary head for
the main task, and multiple secondary heads for capturing spurious
variations or biases present in the datasets. By minimizing the classification
loss w.r.t. the primary head and the (bias) confusion loss w.r.t.
the secondary heads, their model could learn features that are useful
for the main task but invariant to the spurious variations. In another
study, Li et al. \cite{Li2018} examine a different type of bias referred
to as \emph{``representation bias''}, and contrast it with the dataset
bias in \cite{Khosla2012}. To put it simply, representation bias
arises when a model is trained on datasets that favor certain representations
over others. Consequently, the model tends to rely on these biased
representations to solve the task, limiting its ability to generalize
beyond the training datasets. This bias is closely related to the
feature-correlation bias considered in our paper. Li et al. \cite{Li2018}
also introduce a method to quantify representation bias  by analyzing
the classifier's performance. In their subsequent work \cite{Li2019},
they define representation bias as the normalized mutual information
between the feature vector and the class label, and proposed a debiasing
method that optimizes the classifier's parameters and the sample weights
in an adversarial manner.

\paragraph{Group-DRO-based Methods}

Group DRO \cite{Hu2018,sagawa2020distributionally} is a variant of
DRO \cite{BenTal2013,Duchi2016}, which aims to train a robust (against
group shifts) classifier by minimizing the worst-case expected loss
over a set of distributions $Q$ defined as a mixture of sample groups.
There are various ways to characterize a group. In the context of
bias mitigation, a group is typically associated with a pair of class
and bias attributes $(u,b)$ \cite{liu2021just,sagawa2020distributionally}.
When $u$, and $b$ are known and discrete, a group $(u,b)$ can be
explicitly identified. Sagawa et al. \cite{sagawa2020distributionally}
discovered that naively applying group DRO to train overparameterized
neural networks does not enhance performance on the test worst group
compared to the standard empirical risk minimization (ERM). To address
this, they propose imposing stronger regularizations, such as L2 penalty
or early stopping, on the group DRO model to improve generalization
and achieve better worst-group accuracies in testing. Liu et al. \cite{liu2021just}
introduce Just Train Twice (JTT), a two-stage approach, that does
not require group annotation during training. In the first stage,
this method trains a classifier with ERM for some epochs on the original
training data, and constructs an error set comprising samples misclassified
by the classifier. In the second stage, it trains another \emph{``robust''}
classifier with samples in the error set upweighted by a constant
value $\lambda>1$. The rationale is that samples in the error set
likely originate from challenging groups (i.e., groups corresponding
to bias-conflicting samples), and assigning them higher weights encourages
the classifier to focus more on these groups. Correct-N-Contrast (CnC)
\cite{zhang2022correct} is an extension of JTT, also involving two
training stages. The first stage of CnC resembles that of JTT, entailing
training an ERM classifier. The second stage incorporates contrastive
learning with the anchor, positives, and negatives selected based
on the class label and ERM prediction.

\paragraph{Class Imbalance Bias Mitigation}

A closely related type of bias that also arises in our formula in
Eq.~\ref{eq:ERM_3} is \emph{class imbalance} \cite{buda2018systematic,cao2019learning,cui2019class,chawla2002smote,he2009learning,khan2019striking,kukar1998cost,lawrence2012neural,lin2017focal}.
When the classes are not evenly distributed, the model is prone to
predicting the more common classes. Compared to feature-correlation
bias, this bias is more apparent and has been studied for a longer
period of time. Popular methods for mitigating class imbalance bias
include resampling \cite{buda2018systematic,chawla2002smote,estabrooks2004multiple},
and cost-sensitive learning \cite{cao2019learning,cui2019class,elkan2001foundations,khan2017cost,kukar1998cost}.
Resampling aims to balance the class prior by either oversampling
a minor class $y$ at a rate of $\frac{p(y_{\text{most}})}{p(y)}$
or undersampling a major class $y$ at a rate of $\frac{p(y_{\text{least}})}{p(y)}$,
depending on whether the reference is the most class $y_{\text{most}}$
or the least class $y_{\text{least}}$. In mini-batch learning via
random sampling with replacement, the two resampling strategies are
the same. An inherent limitation of resampling is that samples of
the minor class are duplicated, which can lead to overfitting \cite{he2009learning}.
To address this problem, different variants of resampling have been
proposed \cite{he2009learning}. One notable technique is Synthetic
Minority Oversampling TEchnique (SMOTE) \cite{chawla2002smote}, which
generates synthetic samples of the minor class for oversampling by
combining the feature vector of a sample of interest with its nearest
neighbor feature vector through interpolation, similar to latent mixup
\cite{verma2019manifold} and SelecMix \cite{Hwang2022}. In cost-sensitive
learning, higher losses are assigned to samples belonging to the minor
class to account for their lower frequency. This can be achieved by
either reweighting the loss function by a scalar (e.g., the inverse
class frequency) or by designing new losses that inherently incorporate
the sample weight \cite{cao2019learning,cui2019class,khan2017cost,lin2017focal}.
In general, addressing the class imbalance bias is easier than addressing
the feature-correlation bias, as the former does not involve the bias
attribute.

\paragraph{Causal Inference}

In the realm of causal inference, bias typically arises as a non-causal
(spurious) association between the treatment and target variables
mediated by a confounder. This bias is commonly known as \emph{``confounding
bias''}. Causal inference offers useful techniques and methodologies
for mitigating such bias, with inverse probability weighting (IPW)
\cite{hirano2001estimation,hirano2003efficient} being one of them.
Other approaches include causal invariant representation learning
\cite{lu2021invariant,mitrovic2020representation,wang2022out}, proxy
variables \cite{yue2021transporting}, front-door adjustment \cite{nguyen2022front},
which have been studied recently for various generalization tasks.

\section{Technical comparisons between LfF, PGD and our method\label{sec:Comparison-with-LfF-PGD}}

Below, we provide a technical comparisons between LfF, PGD and our
method. 

LfF \cite{Nam2020} is a \emph{reweighting} method like our LW, which
trains a biased classifier $p_{\psi}(y|x)$ and a debiased classifier
$p_{\theta}(y|x)$ in parallel. The weight for each sample $n$ is
computed as follows:
\begin{align}
w_{n} & =\frac{\Loss_{\psi}^{\text{xent}}(x_{n},y_{n})}{\Loss_{\psi}^{\text{xent}}(x_{n},y_{n})+\Loss_{\theta}^{\text{xent}}(x_{n},y_{n})}\\
 & =\frac{1}{1+\frac{\Loss_{\theta}^{\text{xent}}(x_{n},y_{n})}{\Loss_{\psi}^{\text{xent}}(x_{n},y_{n})}}=\frac{1}{1+\frac{-\log p_{\theta}(y_{n}|x_{n})}{-\log p_{\psi}(y_{n}|x_{n})}}\label{eq:LfF_weight_2}
\end{align}
where $\Loss_{\psi}^{\text{xent}}(x_{n},y_{n})$ and $\Loss_{\theta}^{\text{xent}}(x_{n},y_{n})$
are the losses of the biased and debiased classifiers, respectively.
Since $\Loss_{\psi}^{\text{xent}}(x_{n},y_{n})$ and $\Loss_{\theta}^{\text{xent}}(x_{n},y_{n})$
are non-negative, $w_{n}\in[0,1]$, which ensures stability during
training of the debiased classifier. Moreover, the weight in Eq.~\ref{eq:LfF_weight_2}
exhibits a time-varying behavior similar to ALW. At the initial stages
of training, when both the biased and debiased classifiers have limited
learning, $\Loss_{\psi}^{\text{xent}}(x_{n},y_{n})$ and $\Loss_{\theta}^{\text{xent}}(x_{n},y_{n})$
have large values and are roughly the same. Consequently, the weights
for all training samples are approximately equal, around 0.5. This
enables the debiased classifier to learn from all samples effectively
in the early stages. As the training progresses, $\Loss_{\psi}^{\text{xent}}(x_{n},y_{n})$
tends to decrease for BA samples and increase (or decrease at a much
slower rate) for BC samples (Fig.~\ref{fig:biased-classifier-curves}).
As a result, the weights for BA and BC samples gradually become smaller
and larger, respectively, encouraging the debiased classifier to focus
less on BA samples and more on BC samples.

PGD \cite{Ahn2023} is a two-stage \emph{resampling} method like our
WS. In the first stage, it trains a biased classifier $p_{\psi}(y|x)$
using the GCE loss to distinguish between BC and BA samples. In the
second stage, it calculates the resampling weight $w_{n}$ for each
training sample $n$ as the normalized L2 norm of the gradient of
$\Loss_{\psi}^{\text{xent}}(x_{n},y_{n})$ w.r.t. the parameters of
the final fully-connected (FC) layer of the biased classifier. The
computation of $w_{n}$ is given by:
\begin{equation}
w_{n}=\frac{\left\Vert \nabla_{\psi_{\text{top}}}\Loss_{\psi}^{\text{xent}}(x_{n},y_{n})\right\Vert _{2}}{\sum_{(x_{m},y_{m})\in\Data}\left\Vert \nabla_{\psi_{\text{top}}}\Loss_{\psi}^{\text{xent}}(x_{m},y_{m})\right\Vert _{2}}\label{eq:PGD_weight_1}
\end{equation}
Since $w_{n}$ is used for resampling instead of reweighting, we can
safely ignore the denominator in the above formula of $w_{n}$ and
consider $w_{n}=\left\Vert \nabla_{\psi_{\text{top}}}\Loss_{\psi}^{\text{xent}}(x_{n},y_{n})\right\Vert _{2}$.
It is worth noting that the final layer's output is the class probability
vector $p_{n}$ computed as $p_{n}$ = $p_{\psi}(Y|x_{n})$ = $\softmax\left(\mathrm{W}^{\top}h_{n}+\mathrm{b}\right)$
where $h_{n}$, $\mathrm{W}$, and $\mathrm{b}$ represent the input,
weight and bias of the final layer, respectively. Therefore, we can
compute $w_{n}$ as follows $w_{n}$ = $\left\Vert \nabla_{\mathrm{W}}\left(-\log p_{\psi}(y_{n}|x_{n})\right)\right\Vert _{2}$
= $\left\Vert \left(p_{n}-1_{y_{n}}\right)h_{n}^{\top}\right\Vert _{2}$
with $1_{y_{n}}$ representing the one-hot vector at $y=y_{n}$. For
BA samples, $w_{n}$ tends to be 0 since $p_{n}\approx1_{y_{n}}$
and $\nabla_{\mathrm{W}}\Loss_{\psi}^{\text{xent}}(x_{n},y_{n})\approx0$.
This implies that BA samples may hardly contribute to training the
debiased classifier, which can results in the inferior performance
of PGD in some cases.

\section{Further details about datasets and training settings\label{sec:More-Details-about-Training-Settings}}

\subsection{Datasets\label{subsec:Datasets}}

We provide summarization of the datasets used in our experiment in
Table~\ref{tab:dataset-summary}.

\begin{table*}
\begin{centering}
\begin{tabular}{ccccc}
\hline 
Dataset & Image size & Target & Bias & \#classes\tabularnewline
\hline 
\hline 
Colored MNIST & 3$\times$32$\times$32 & Digit & BG Color & 10\tabularnewline
Corrupted CIFAR10 & 3$\times$32$\times$32 & Object & Noise & 10\tabularnewline
Biased CelebA & 3$\times$128$\times$128 & Hair color & Gender & 2\tabularnewline
BAR & 3$\times$256$\times$256 & Action & Context & 6\tabularnewline
\hline 
\end{tabular}
\par\end{centering}
\caption{Summarization of the datasets used in our experiments\label{tab:dataset-summary}}
\end{table*}

\subsection{Optimization hyperparameters\label{subsec:Optimization-hyperparameters}}

We provide the optimization hyperparameters for our method in Table~\ref{tab:optimization-hyperparams}.

\begin{table*}
\begin{centering}
\begin{tabular}{ccccccccc}
\hline 
Dataset & $opt$ & $lr$ & $wd$ & $mo$ & $ld$ & $ldep$ & $ep$ & $bs$\tabularnewline
\hline 
\hline 
Colored MNIST & Adam & 1e-3 & - & - & - & - & 120 & 128\tabularnewline
Corrupted CIFAR10 & Adam & 1e-3 & - & - & - & - & 160 & 128\tabularnewline
Biased CelebA & SGD & 1e-3 & 1e-4 & 0.9 & 0.1 & 80, 120 & 160 & 256\tabularnewline
BAR & SGD & 1e-3 & 1e-5 & 0.9 & 0.1 & 80, 120 & 160 & 128\tabularnewline
\hline 
\end{tabular}
\par\end{centering}
\caption{Optimization hyperparameters for our method across different datasets.
Abbreviation meanings: $opt$ (optimizer), $lr$ (learning rate),
$wd$ (weight decay), $mo$ (momentum), $ld$ (learning rate decay),
$ldep$ (epochs at which learning rate are decayed), $ep$ (total
number of epochs), $bs$ (batch size).\label{tab:optimization-hyperparams}}
\end{table*}

\subsection{Data augmentation\label{subsec:Data-augmentation}}

For Colored MNIST, we do not use any data augmentation. For Corrupted
CIFAR10, we use RandomCrop(32, padding=4) followed by RandomHorizontalFlip().
For Biased CelebA, we resize the input image to the specified size,
and use RandomHorizonalFlip(). For BAR, we use RandomResizedCrop(256,
scale=(0.3, 1.0)) followed by RandomHorizontalFlip().

\subsection{Optimal settings of our method}

In Table~\ref{tab:Optimal-settings}, we provide the optimal settings
of our proposed method LW.

\begin{table*}
\begin{centering}
\begin{tabular}{|c|c|c|c|c|}
\hline 
\multirow{2}{*}{Dataset} & \multirow{2}{*}{BC (\%)} & \multicolumn{1}{c|}{Biased Classifier} & \multicolumn{2}{c|}{LW}\tabularnewline
\cline{3-5} 
 &  & $T_{\text{bias}}$ & $\gamma$ & $T_{\text{anneal}}$\tabularnewline
\hline 
\hline 
\multirow{3}{*}{Colored MNIST} & 0.5 & \multirow{1}{*}{10} & 200 & \multirow{3}{*}{0}\tabularnewline
\cline{2-4} 
 & 1.0 & 10 & 100 & \tabularnewline
\cline{2-4} 
 & 5.0 & 1$\rightarrow$2 & 20 & \tabularnewline
\hline 
\multirow{3}{*}{Corrupted CIFAR10} & 0.5 & \multirow{1}{*}{100$\rightarrow$160} & 2k$\rightarrow$5k & \multirow{3}{*}{40}\tabularnewline
\cline{2-4} 
 & 1.0 & 60$\rightarrow$160 & 1k$\rightarrow$5k & \tabularnewline
\cline{2-4} 
 & 5.0 & 20 & 500$\rightarrow$5k & \tabularnewline
\hline 
\multirow{3}{*}{Biased CelebA} & 0.5 & \multirow{1}{*}{160} & 500 & \multirow{3}{*}{0}\tabularnewline
\cline{2-4} 
 & 1.0 & 160 & 500 & \tabularnewline
\cline{2-4} 
 & 5.0 & 160 & 500 & \tabularnewline
\hline 
BAR & - & 10 & 50$\rightarrow$100 & 0\tabularnewline
\hline 
\end{tabular}
\par\end{centering}
\caption{Optimal settings of our method.\label{tab:Optimal-settings}}
\end{table*}

\section{Results when the bias label is available\label{subsec:Results-when-the-bias-is-available}}

\begin{table*}
\begin{centering}
\resizebox{0.9\textwidth}{!}{%
\par\end{centering}
\begin{centering}
\begin{tabular}{>{\raggedright}p{0.08\textwidth}cccccc}
\hline 
Dataset & BC (\%) & Vanilla & LNL & BiasBal & \multicolumn{1}{c}{WS} & \multicolumn{1}{c}{LW}\tabularnewline
\hline 
\hline 
\multirow{3}{0.08\textwidth}{Colored MNIST} & 0.5 & 80.18$\pm$1.38 & 80.05$\pm$1.02 & \textbf{97.44$\pm$0.12} & 96.15$\pm$0.33 & 96.16$\pm$0.35\tabularnewline
 & 1.0 & 87.48$\pm$1.75 & 87.52$\pm$1.53 & \textbf{97.89$\pm$0.20} & \textcolor{gray}{97.58$\pm$0.09} & 97.11$\pm$0.12\tabularnewline
 & 5.0 & 97.04$\pm$0.21 & 98.77$\pm$0.18 & \textcolor{gray}{98.92$\pm$0.13} & \textbf{98.92$\pm$0.07} & 98.55$\pm$0.07\tabularnewline
\hline 
\multirow{3}{0.08\textwidth}{Corrupted CIFAR10} & 0.5 & 28.00$\pm$1.15 & 28.13$\pm$0.98 & \textbf{45.52$\pm$1.37} & 31.02$\pm$0.72 & \textcolor{gray}{36.50$\pm$0.51}\tabularnewline
 & 1.0 & 34.56$\pm$0.87 & 34.43$\pm$1.03 & \textbf{53.18$\pm$1.87} & 40.06$\pm$1.24 & \textcolor{gray}{46.41$\pm$1.46}\tabularnewline
 & 5.0 & 59.33$\pm$1.26 & 59.52$\pm$0.85 & \textbf{72.60$\pm$0.69} & 69.91$\pm$0.54 & \textcolor{gray}{70.29$\pm$0.75}\tabularnewline
\hline 
\multirow{3}{0.08\textwidth}{Biased CelebA} & 0.5 & 77.43$\pm$0.42 & 76.64$\pm$0.73 & \textbf{87.74$\pm$0.19} & 78.48$\pm$0.62 & \textcolor{gray}{86.49$\pm$0.33}\tabularnewline
 & 1.0 & 80.58$\pm$0.41 & 79.72$\pm$0.54 & \textbf{87.82$\pm$0.36} & 82.41$\pm$0.33 & \textcolor{gray}{85.32$\pm$0.40}\tabularnewline
 & 5.0 & 86.35$\pm$0.33 & 86.41$\pm$0.35 & \textbf{88.78$\pm$0.22} & 86.90$\pm$0.23 & \textcolor{gray}{87.83$\pm$0.21}\tabularnewline
\hline 
\end{tabular}}
\par\end{centering}
\caption{Results of different debiasing methods when the bias label is available.
The best and second best results are highlighted in bold and gray,
respectively.\label{tab:results-with-bias-labels}}
\end{table*}

\begin{figure*}
\begin{centering}
\resizebox{\textwidth}{!}{%
\par\end{centering}
\begin{centering}
\begin{tabular}{>{\raggedright}m{0.06\textwidth}ccc}
 & Colored MNIST & Corrupted CIFAR10 & Biased CelebA\tabularnewline
\multirow{1}{0.06\textwidth}[0.175\textwidth]{Test Acc.} & \includegraphics[width=0.4\textwidth]{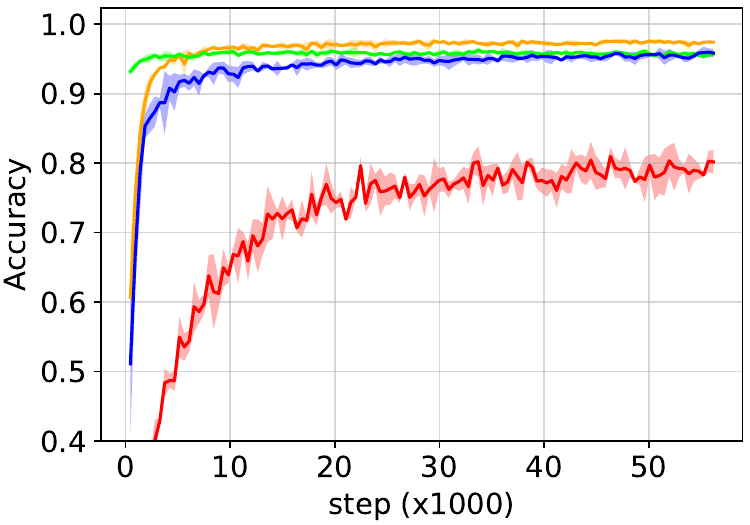} & \includegraphics[width=0.4\textwidth]{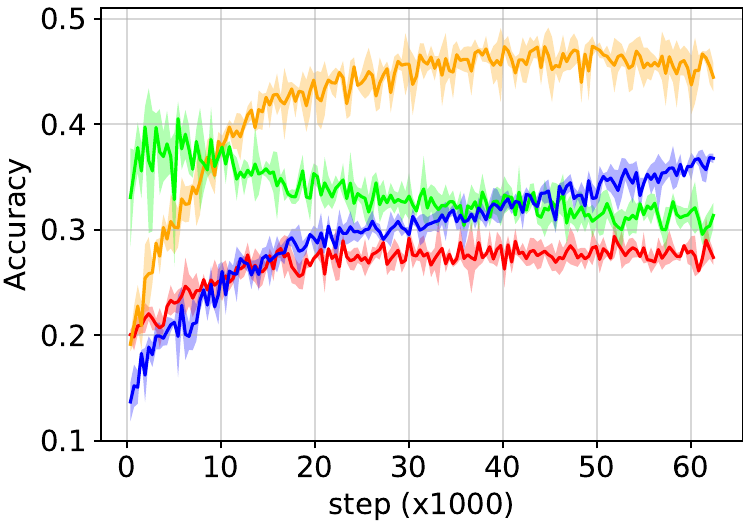} & \includegraphics[width=0.4\textwidth]{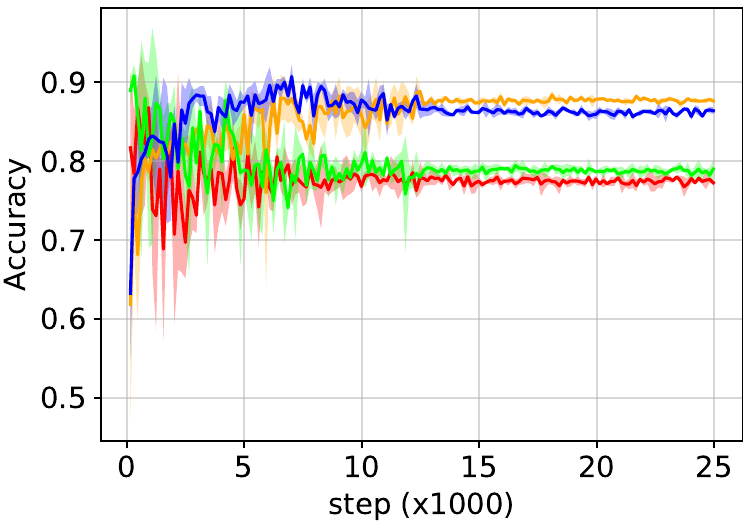}\tabularnewline
\multirow{1}{0.06\textwidth}[0.175\textwidth]{Test Xent} & \includegraphics[width=0.4\textwidth]{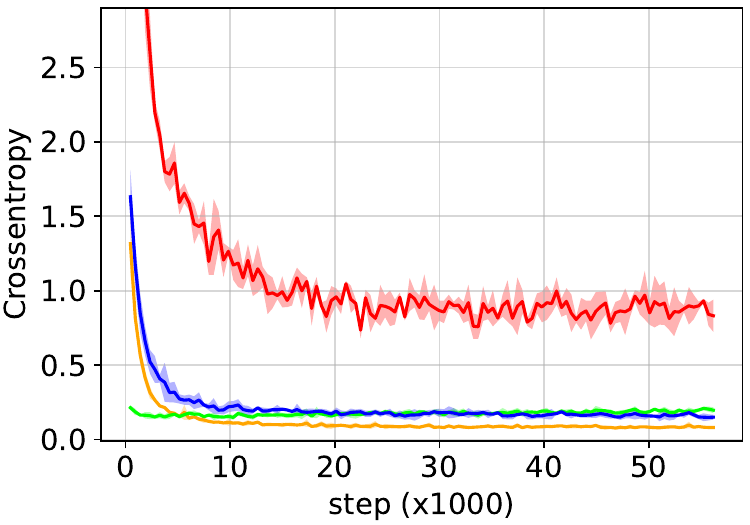} & \includegraphics[width=0.4\textwidth]{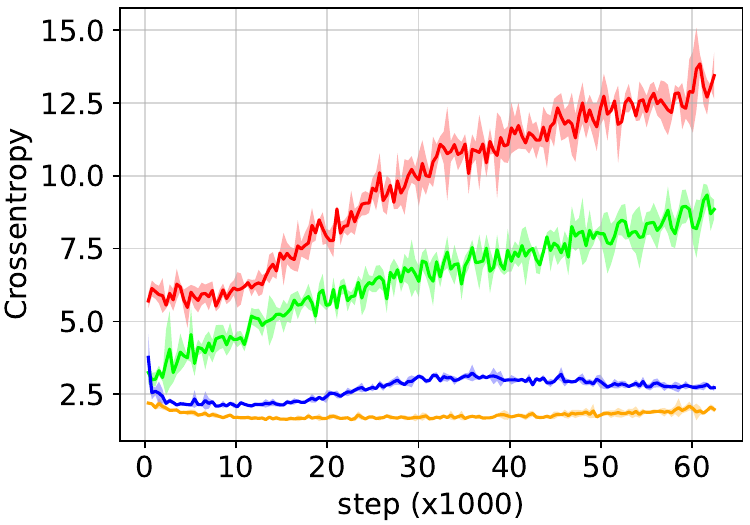} & \includegraphics[width=0.4\textwidth]{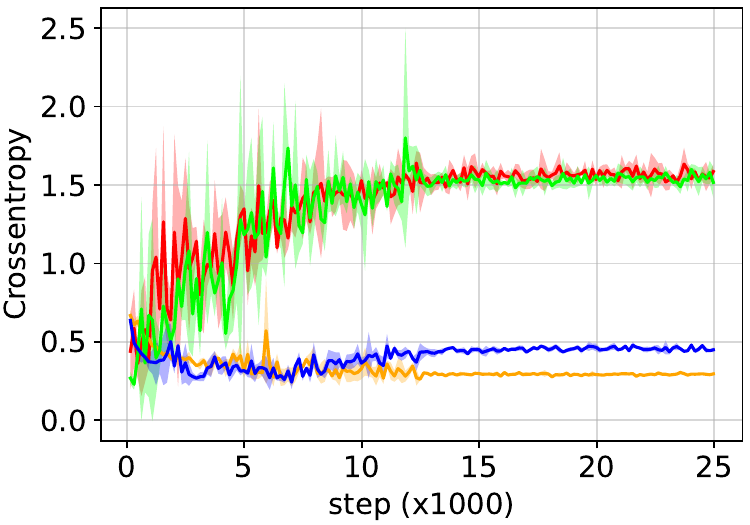}\tabularnewline
 & \multicolumn{3}{c}{\includegraphics[width=0.5\textwidth]{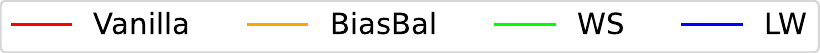}}\tabularnewline
\end{tabular}}
\par\end{centering}
\caption{Test curves of the vanilla classifier, BiasBal, and our proposed methods
WS and LW on Colored MNIST, Corrupted CIFAR10, and Biased CelebA with
the BC ratio of 0.5\%. The bias label is assumed to be given, and
$p(y|b)$ used in BiasBal, WS, and LW is estimated directly from training
data.\label{fig:Test-curves-with-bias-labels}}
\end{figure*}

In this section, we aim to empirically validate our theoretical analysis
on feature-correlation bias in Sections~\ref{sec:A-Statistical-View}
and \ref{sec:Mitigating-Dataset-Bias} by examining the effectiveness
of our proposed weighted sampling (WS) and loss weighting\emph{ }(LW)
techniques in mitigating bias in datasets with known discrete bias
labels. We consider Learning Not to Learn (LNL) \cite{Kim2019} and
Bias Balancing (BiasBal) \cite{Hong2021} as baselines since these
approaches explicitly utilize the bias label. From Table~\ref{tab:results-with-bias-labels},
it is evident that classifiers trained with either WS or LW achieve
significantly higher test accuracies than the vanilla classifier.
This observation suggests that WS and LW effectively reduce bias in
training data, validating our theoretical analysis. However, WS and
LW exhibit distinct training dynamics, resulting in varying levels
of effectiveness in bias mitigation. As depicted in Fig.~\ref{fig:Test-curves-with-bias-labels},
on Corrupted CIFAR10, WS quickly achieves good performance in the
early stages of training but gradually degrades over time. We hypothesize
that this behavior is attributed to the lack of data diversity resulting
from over/under-sampling of BC/BA samples in WS. On the other hand,
LW takes more time than WS to achieve similar performance due to its
small updates on most training samples (i.e., BA samples). Nonetheless,
LW demonstrates a steady improvement in performance.

Besides, WS and LW outperform LNL, a method that aims to learn features
independent of bias information. Interestingly, during our experiments,
we observed minimal performance difference between LNL and the vanilla
classifier, as shown in Table~\ref{tab:results-with-bias-labels}.
This suggests that the features learned by LNL still contain bias
information. We hypothesize that when class and bias labels are highly
correlated, it becomes challenging from a statistical perspective
to learn features that are truly independent of bias while accurately
predicting the class.

However, our methods perform worse and are less robust than BiasBal,
especially on Corrupted CIFAR10. Fig.~\ref{fig:Test-curves-with-bias-labels}
illustrates that BiasBal converges more rapidly than our methods with
higher test accuracies and smaller test crossentropy losses in most
settings. The exceptional performance of BiasBal motivates us to delve
deeper into the theoretical mechanisms and design differences between
BiasBal and our methods that account for BiasBal's superior performance.

We discovered that like LW, BiasBal mitigates bias via reweighting
the model trained on biased data. It also arrives at the same reweighting
term of $\frac{1}{p(y|b)}$ as in LW despite employing a different
reasoning technique that leverages Bayes' rule and two assumptions:
i) $p(y|x)=p(y|x,b)$ and ii) $p(x|y,b)$ are the same across different
joint distributions $p(y,b)$. The bias correction formula of BiasBal
(Eq.~A.8 in the supplementary material of \cite{Hong2021}) is given
as follows:
\begin{equation}
p_{train}(y|x)=p_{u}(y|x)p_{train}(y|b)\frac{K}{C}\label{eq:BiasBal-bias-correction}
\end{equation}
where $p_{train}$ and $p_{u}$ denotes the distribution over the
biased training data and the unbiased distribution in which $y$ and
b are independent, respectively; $C$ is the number of classes, and
$K=\frac{p_{u}(x|b)}{p_{train}(x|b)}$ is a constant w.r.t. $y$.
Eq.~\ref{eq:BiasBal-bias-correction} implies that $p_{u}(y|x)$
$\propto$ $p_{train}(y|x)\frac{1}{p_{train}(y|b)}$, allowing us
to reweight the biased distribution $p_{train}(y|x)$ by $\frac{1}{p_{train}(y|b)}$
to achieve a theoretically unbiased distribution.

The key distinction between BiasBal and WS/LW lies in their respective
debiasing mechanisms. While WS/LW \emph{corrects the learning process}
of $p_{\theta}(y|x)$ (via resampling/reweighting the training data
distribution) to achieve an unbiased target $p_{\theta}(y|x)$ \emph{indirectly},
BiasBal \emph{directly corrects the target} $p_{\theta}(y|x)$ and
learns with the (unnormalized) bias-adjusted target $\hat{p}_{\theta}(y|x)=p_{\theta}(y|x)p_{train}(y|b)$.
This debiasing mechanism of BiasBal is generally more robust than
that of WS/LW, as it neither compromises the diversity of training
data nor introduces training instability. Moreover, while WS and LW
focus on correcting bias in $p(y_{n}|x_{n})$ for the class $y_{n}$
associated with a particular input $x_{n},$ BiasBal considers bias
correction of $p(y|x_{n})$ for every class $y$. These advancements
in the debiasing mechanism of BiasBal likely contribute to its superior
performance.

It is worth noting that the original paper \cite{Hong2021} on BiasBal
exclusively applies the method when the bias label is discrete and
known, as $p_{train}(y|b)$ can be estimated nonparametrically from
the training data. However, with our proposed method for approximating
$p_{train}(y|b)$ using the biased classifier, as discussed in Section~\ref{subsec:Bias-mitigation-based-on-p(u|b)},
we can readily extend BiasBal to scenarios where \emph{the bias label
is either unavailable or continuous}. In Appdx.~\ref{subsec:Target-Bias-Adjustment},
we provide a detailed analysis of this extended version of BiasBal,
which we refer to as \emph{``Target Bias Adjustment''} (TBA) to
better describe its characteristic of adjusting the target distribution.

\section{Additional ablation study results for LW\label{sec:Additional-ablation-study}}

\begin{figure*}
\begin{centering}
\resizebox{0.7\textwidth}{!}{%
\par\end{centering}
\begin{centering}
\begin{tabular}{>{\raggedright}p{0.06\textwidth}cc}
 & Colored MNIST (5\%) & Corrupted CIFAR10 (5\%)\tabularnewline
\multirow{1}{0.06\textwidth}[0.175\textwidth]{Test Acc.} & \includegraphics[width=0.4\textwidth]{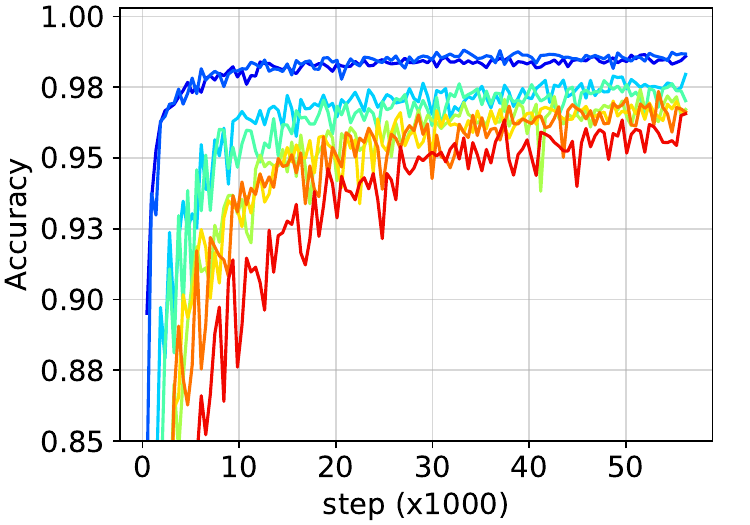} & \includegraphics[width=0.4\textwidth]{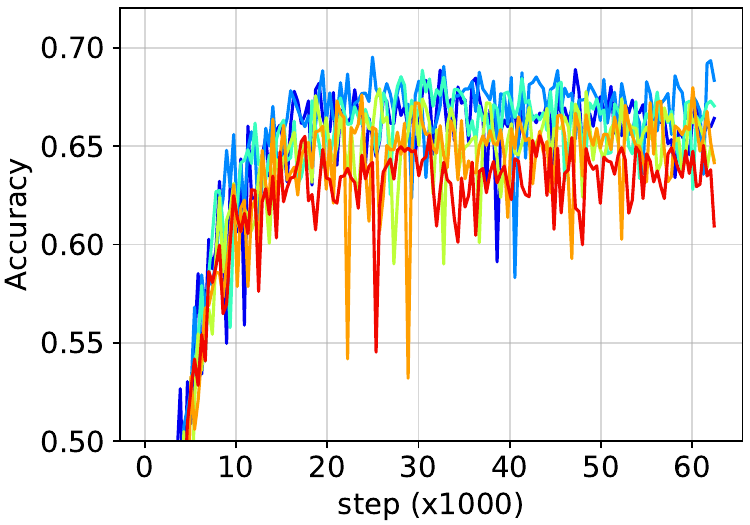}\tabularnewline
\multirow{1}{0.06\textwidth}[0.185\textwidth]{Test Acc. BA} & \includegraphics[width=0.4\textwidth]{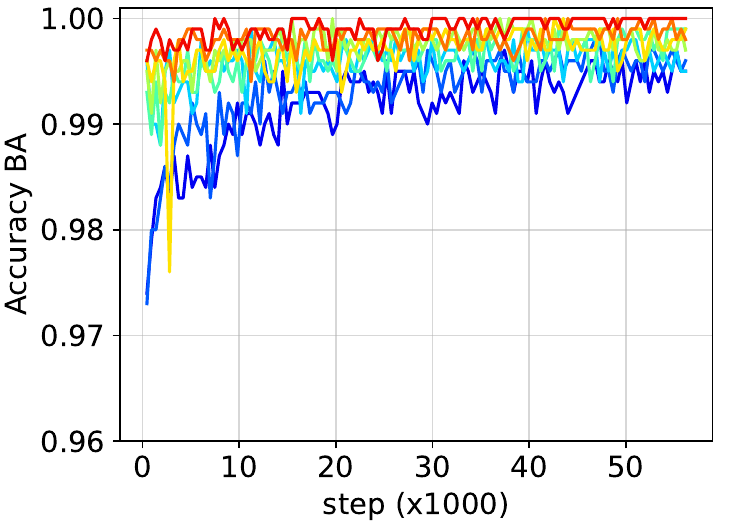} & \includegraphics[width=0.4\textwidth]{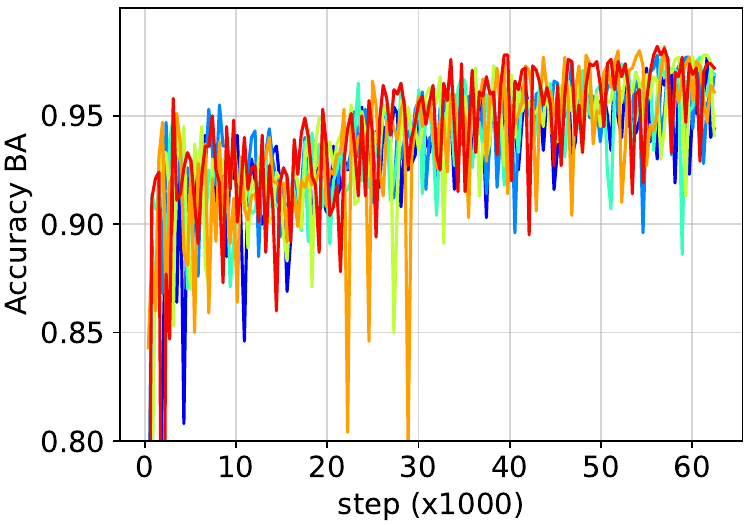}\tabularnewline
\multirow{1}{0.06\textwidth}[0.185\textwidth]{Test Acc. BC} & \includegraphics[width=0.4\textwidth]{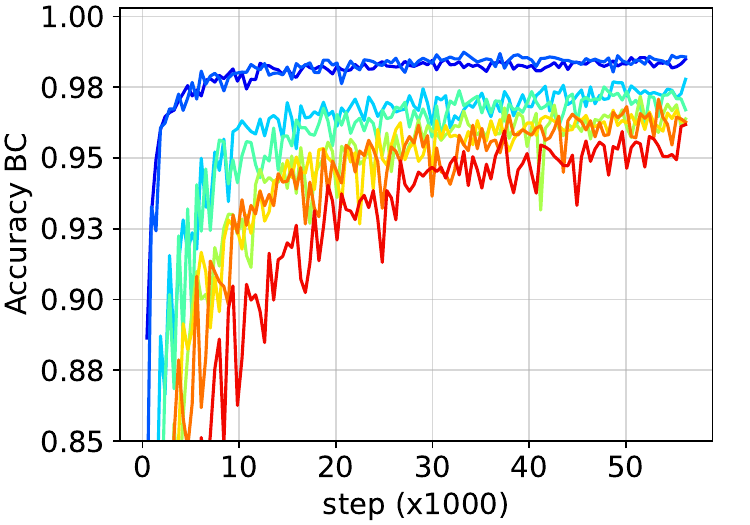} & \includegraphics[width=0.4\textwidth]{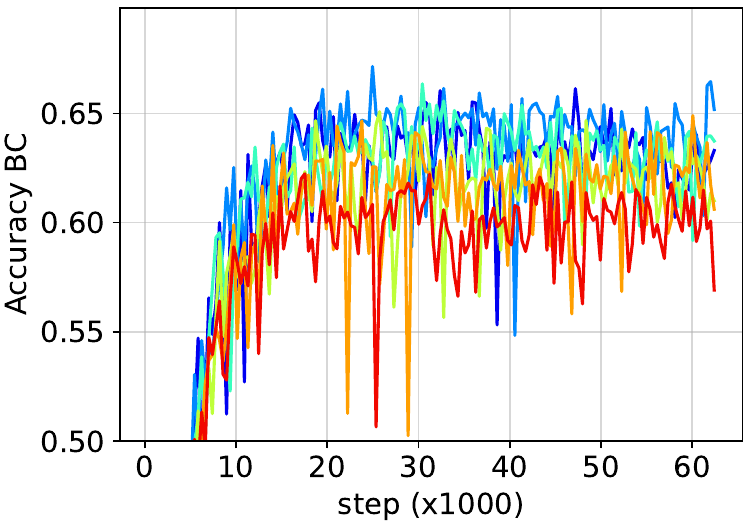}\tabularnewline
\multirow{1}{0.06\textwidth}[0.185\textwidth]{Debias. BC Ratio} & \includegraphics[width=0.4\textwidth]{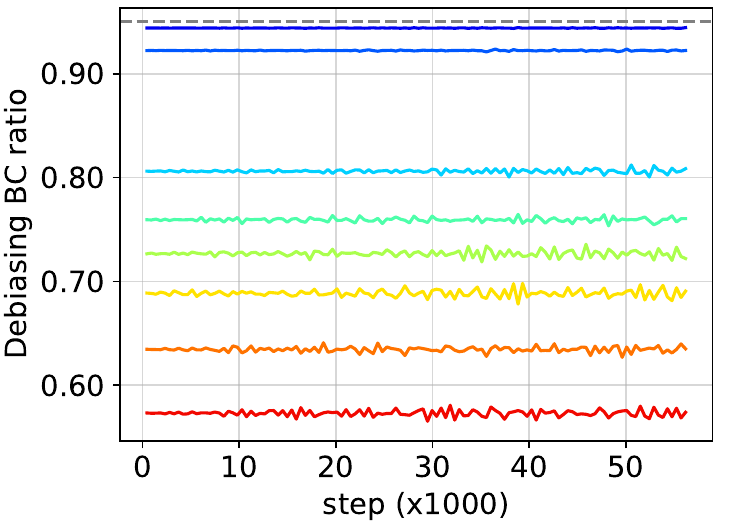} & \includegraphics[width=0.4\textwidth]{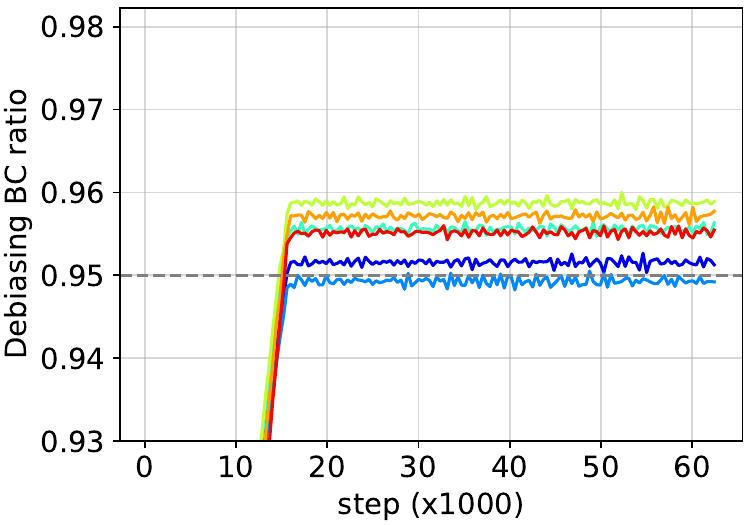}\tabularnewline
 & \includegraphics[width=0.4\textwidth]{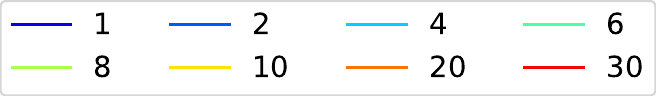} & \includegraphics[width=0.3\textwidth]{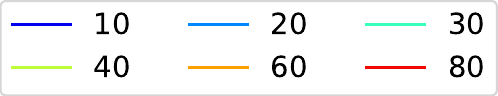}\tabularnewline
\end{tabular}}
\par\end{centering}
\caption{Learning curves of LW w.r.t. different training epochs of the biased
classifier ($T_{\text{bias}}$) on Colored MNIST and Corrupted CIFAR10.
Both datasets have the BC ratio of 5\%. Maximum sample weights ($\gamma$)
for the two datasets are set to 20 and 50, respectively.\label{fig:Diff-epochs-LW}}
\end{figure*}

\begin{figure*}
\begin{centering}
\resizebox{\textwidth}{!}{%
\par\end{centering}
\begin{centering}
\begin{tabular}{>{\raggedright}p{0.06\textwidth}ccc}
 & Colored MNIST (5\%) & Corrupted CIFAR10 (0.5\%) & Biased CelebA (0.5\%)\tabularnewline
\multirow{1}{0.06\textwidth}[0.175\textwidth]{Test Acc.} & \includegraphics[width=0.4\textwidth]{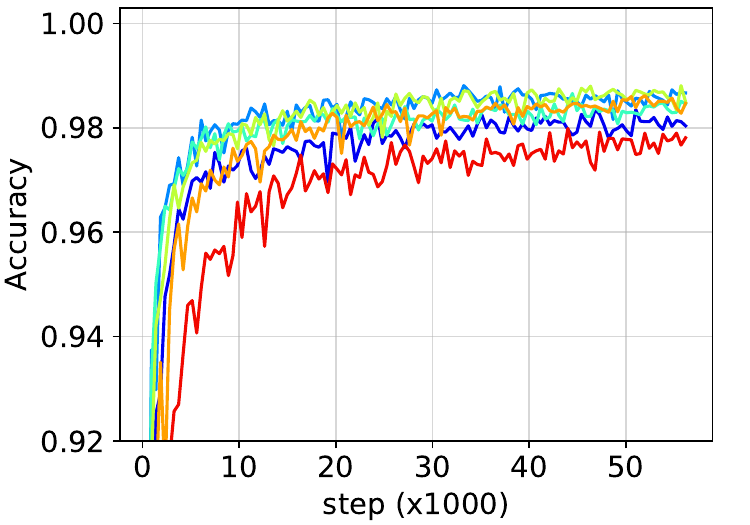} & \includegraphics[width=0.4\textwidth]{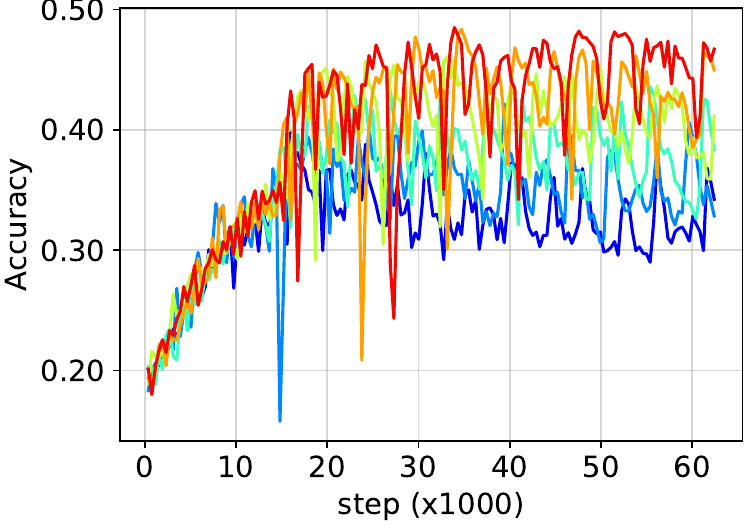} & \includegraphics[width=0.4\textwidth]{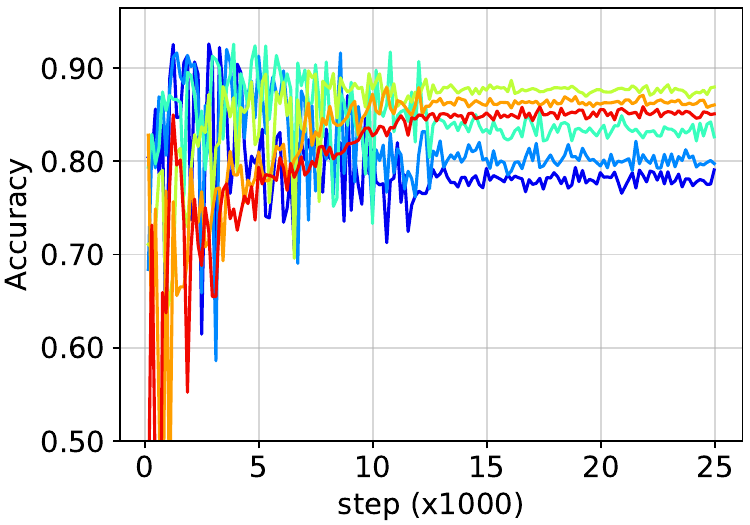}\tabularnewline
\multirow{1}{0.06\textwidth}[0.185\textwidth]{Test Acc. BA} & \includegraphics[width=0.4\textwidth]{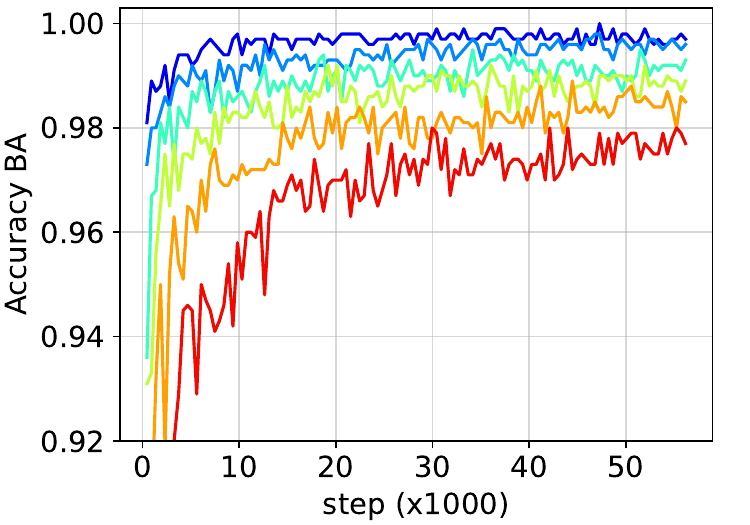} & \includegraphics[width=0.4\textwidth]{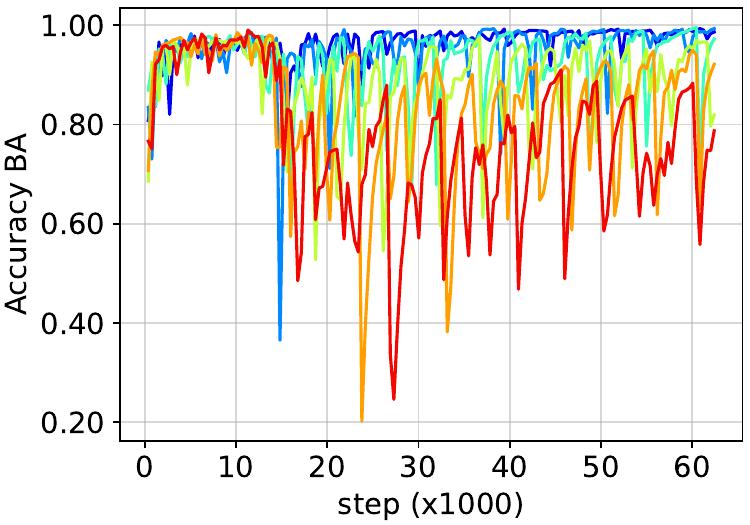} & \includegraphics[width=0.4\textwidth]{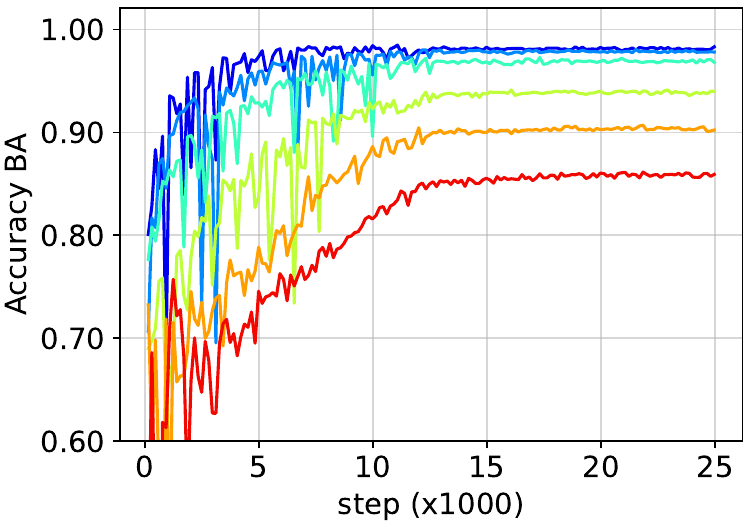}\tabularnewline
\multirow{1}{0.06\textwidth}[0.185\textwidth]{Test Acc. BC} & \includegraphics[width=0.4\textwidth]{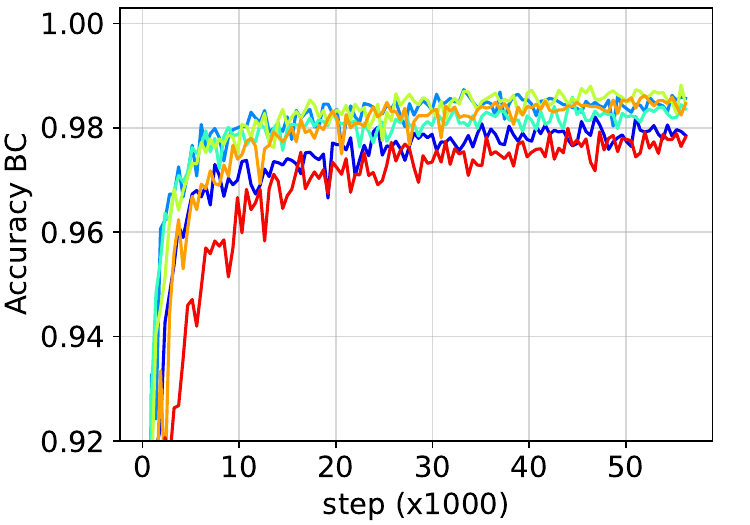} & \includegraphics[width=0.4\textwidth]{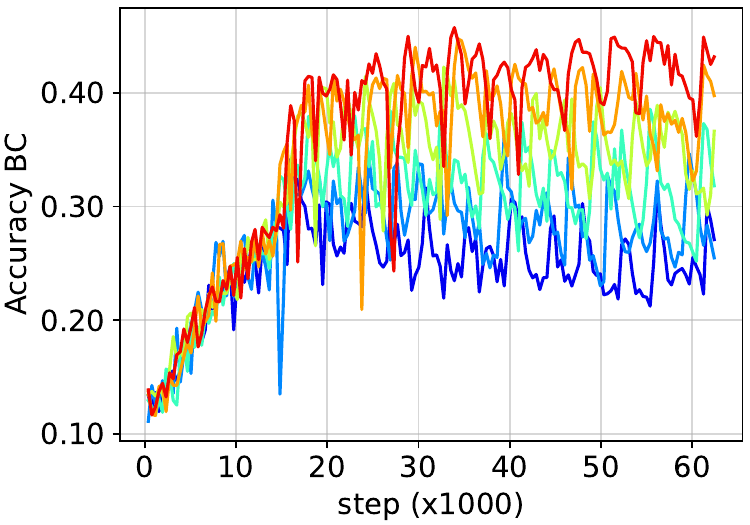} & \includegraphics[width=0.4\textwidth]{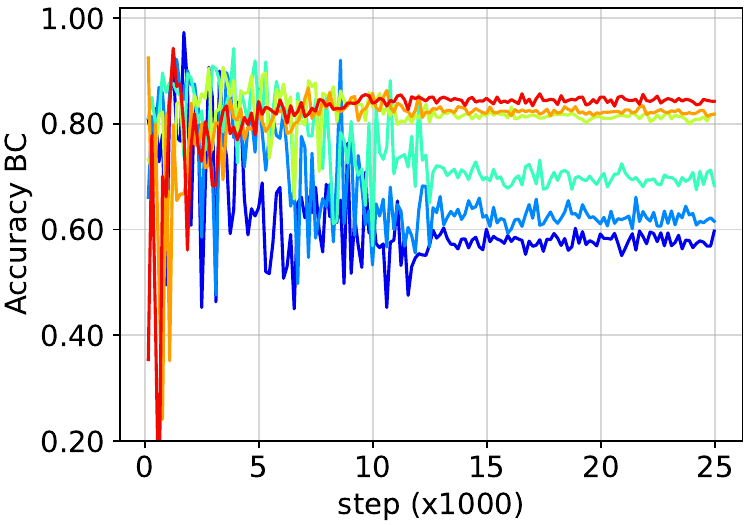}\tabularnewline
\multirow{1}{0.06\textwidth}[0.185\textwidth]{Debias. BC Ratio} & \includegraphics[width=0.4\textwidth]{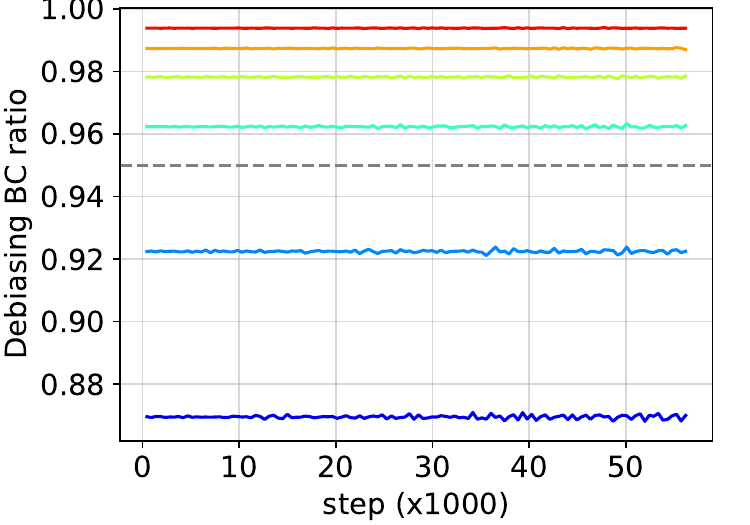} & \includegraphics[width=0.4\textwidth]{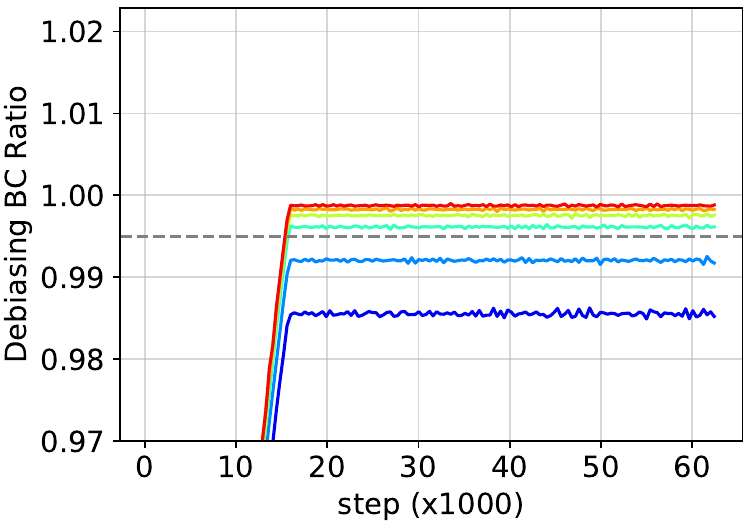} & \tabularnewline
 & \includegraphics[height=0.06\textwidth]{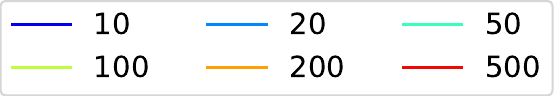} & \includegraphics[height=0.06\textwidth]{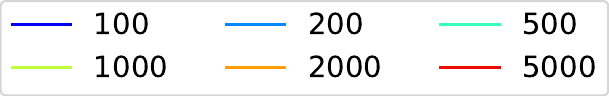} & \includegraphics[height=0.06\textwidth]{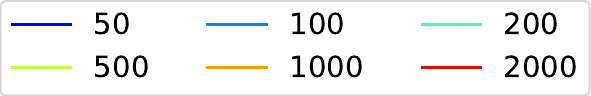}\tabularnewline
\end{tabular}}
\par\end{centering}
\caption{Learning curves of LW w.r.t. different values of $\gamma$ on Colored
MNIST, Corrupted CIFAR10, and Biased CelebA. These datasets have respective
BC ratios of 5\%, 0.5\%, and 0.5\%. Biased classifiers for these datasets
are trained for 2, 120, and 160 epochs, respectively.\label{fig:Diff-sample-weights-LW}}
\end{figure*}

We provide additional experiments examining the impact of varying
the number of training epochs for the biased classifier and adjusting
the maximum cutoff for sample weights on the performance of LW in
Figs.~\ref{fig:Diff-epochs-LW} and \ref{fig:Diff-sample-weights-LW}.

\section{Additional experimental results}

\subsection{Additional results of LfF in comparison with LW\label{subsec:Additional-results-of-LfF}}

\begin{figure*}
\begin{centering}
\resizebox{0.95\textwidth}{!}{%
\par\end{centering}
\begin{centering}
\begin{tabular}{ccc}
\includegraphics[width=0.32\textwidth]{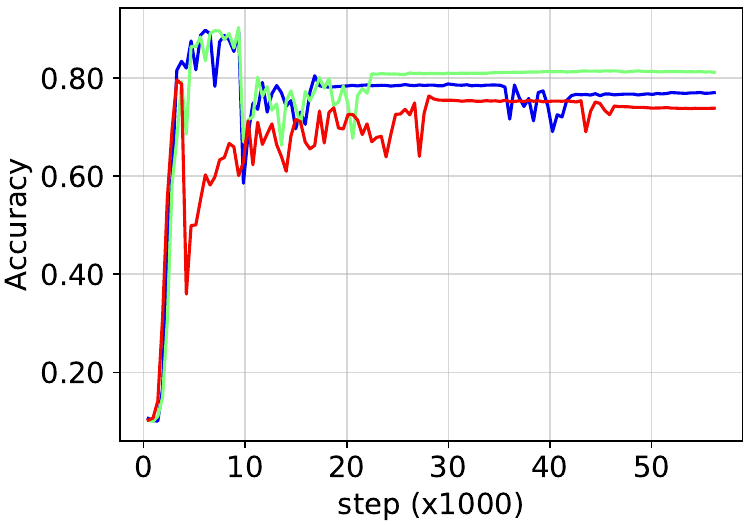} & \includegraphics[width=0.32\textwidth]{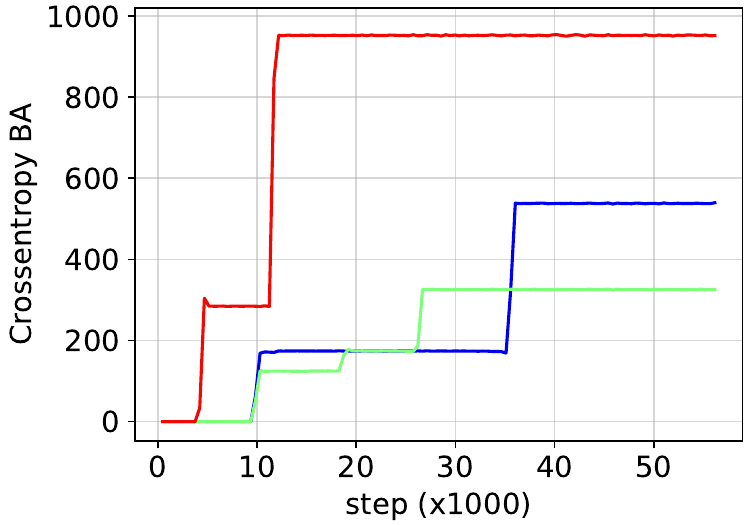} & \includegraphics[width=0.32\textwidth]{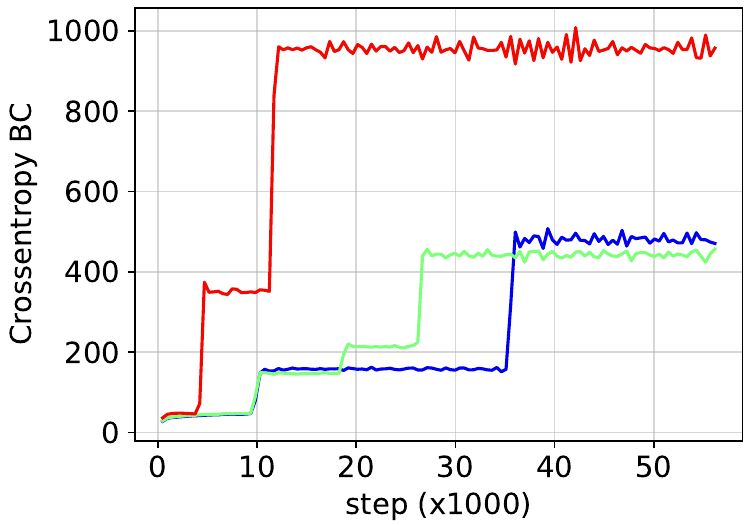}\tabularnewline
(a) Accuracy & (b) Cross-entropy BA & (c) Cross-entropy BC\tabularnewline
\end{tabular}}
\par\end{centering}
\centering{}\caption{Test results of 3 different runs of LfF \cite{Nam2020} on Colored
MNIST with a BC ratio of 0.5\% where the classifier's architecture
is LeNet5 .\label{fig:LfF-LeNet5}}
\end{figure*}

It can be seen from Table~\ref{tab:Main-Results-MNIST-Cifar10} that
LfF, despite requiring minimal hyperparameter tuning, performs well
on both Colored MNIST and Corrupted CIFAR10. Surprisingly, it even
outperforms later proposed methods such as DFA, SelecMix on both Colored
MNIST and Corrupted CIFAR10, as well as PGD on Corrupted CIFAR10.
We hypothesize that there could be several possible reasons for this: 
\begin{enumerate}
\item Some methods like PGD rely heavily on specific data augmentation techniques
(e.g., color jitter) for optimal performance, which are not utilized
in this work (Section~\ref{subsec:Implementation-details}).
\item The selected hyperparameters for DFA and SelecMix in this study may
not be optimal since we reimplemented these methods ourselves and
were unable to thoroughly explore every configuration due to time
constraints.
\item It \emph{might} be that in previous works, LfF unintentionally yields
poor results, particularly on Colored MNIST, despite being implemented
correctly.
\end{enumerate}
Regarding the third point, we have found that the biased classifier
often experiences sudden collapses during training \emph{if its architecture
does not include Batch Norm} \cite{ioffe2015batch} while the vanilla
classifier does not exhibit this issue. Our hypothesis is that the
GCE loss of the biased classifier is primarily optimized for frequently
occurring BA samples. As a result, the loss surface surrounding sporadic
BC samples becomes highly volatile, leading to substantial gradient
updates (in the absence of Batch Norm) whenever these BC samples are
encountered during training. This volatility causes the biased classifier's
parameters to collapse to irrecoverable states. We visualize this
phenomenon of LfF in Fig.~\ref{fig:LfF-LeNet5}, utilizing LeNet5
as the biased classifier architecture without any Batch Norm layer.
It is evident that the cross-entropy loss of this classifier abruptly
spikes to hundreds for both BC and BA samples at certain steps during
training and fails to recover. This makes the relative difficulty
scores of BA samples close to 1 (instead of being close to 0), resulting
in a degradation of LfF's test accuracy.

The reason why we raise the third point and carefully analyze is due
to the varying results of LfF reported in different studies \cite{Ahn2023,Hong2021,Hwang2022,Lee2021}.
Some works utilizing multilayer perceptrons (MLPs) as the architecture
for the biased classifier only achieve test accuracies of 52.50\%
\cite{Lee2021} and 63.86\% \cite{Hwang2022} for LfF on Colored MNIST
with a BC ratio of 0.5\%, while others adopting convolutional networks
with Batch Norm can achieve significantly higher test accuracies of
90.3\% \cite{Hong2021} and 91.35\% \cite{Ahn2023}.

\begin{figure*}
\begin{centering}
\resizebox{\textwidth}{!}{%
\par\end{centering}
\begin{centering}
\begin{tabular}{>{\raggedright}m{0.06\textwidth}ccc}
 & 0.5\% & 1\% & 5\%\tabularnewline
\multirow{1}{0.06\textwidth}[0.185\textwidth]{Test Acc. BA} & \includegraphics[width=0.4\textwidth]{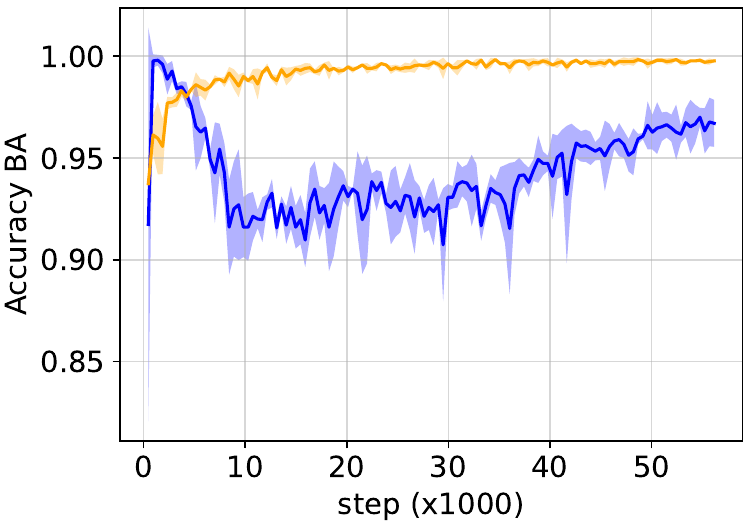} & \includegraphics[width=0.4\textwidth]{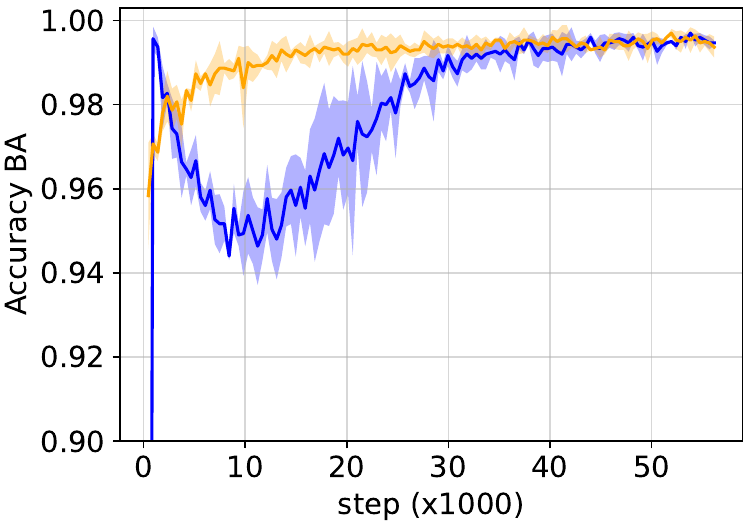} & \includegraphics[width=0.4\textwidth]{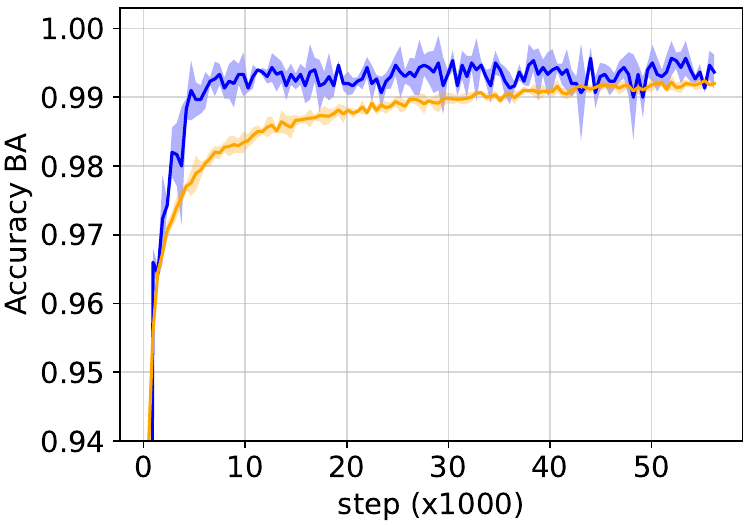}\tabularnewline
\multirow{1}{0.06\textwidth}[0.185\textwidth]{Test Acc. BC} & \includegraphics[width=0.4\textwidth]{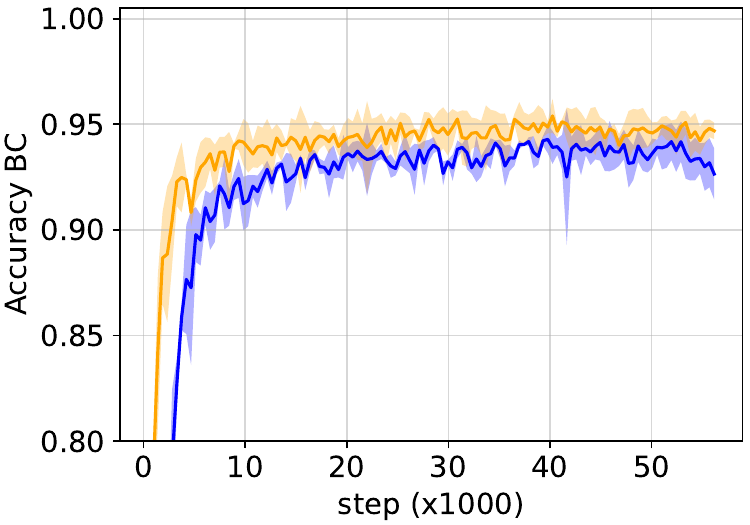} & \includegraphics[width=0.4\textwidth]{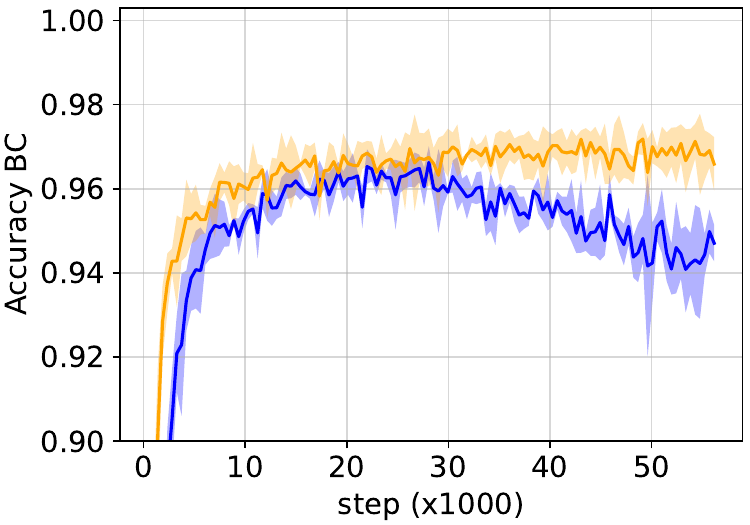} & \includegraphics[width=0.4\textwidth]{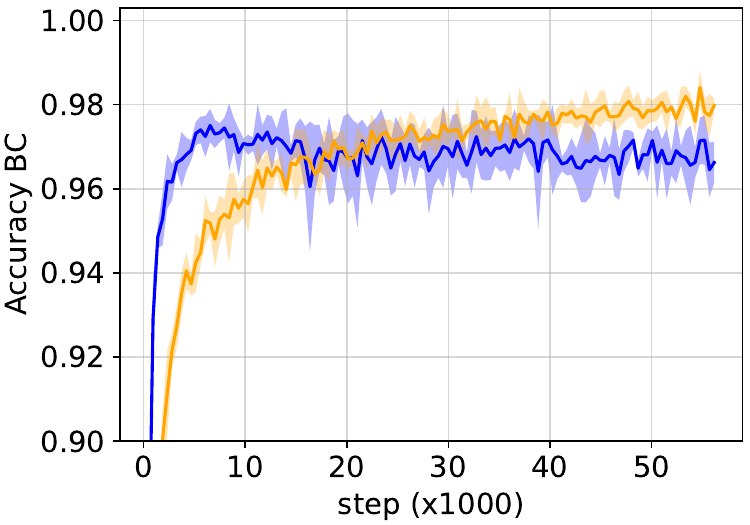}\tabularnewline
\multirow{1}{0.06\textwidth}[0.185\textwidth]{Debias. BC Ratio} & \includegraphics[width=0.4\textwidth]{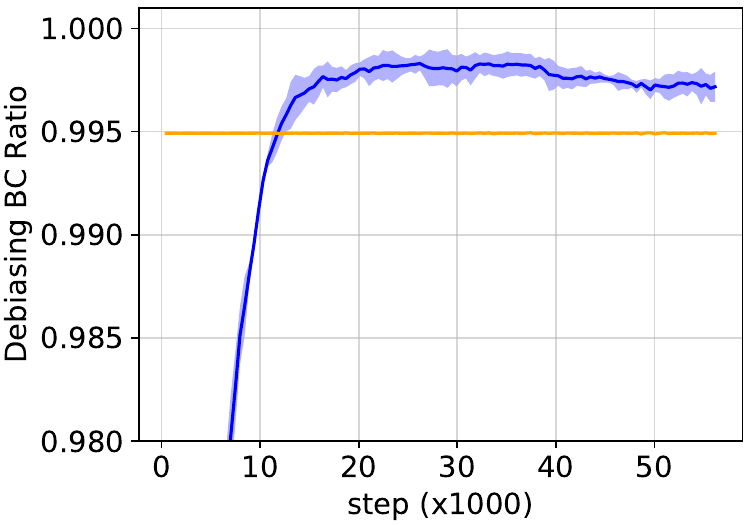} & \includegraphics[width=0.4\textwidth]{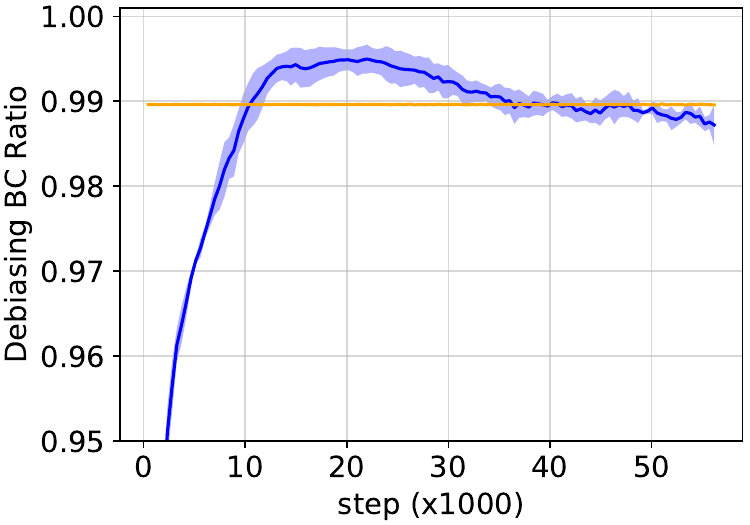} & \includegraphics[width=0.4\textwidth]{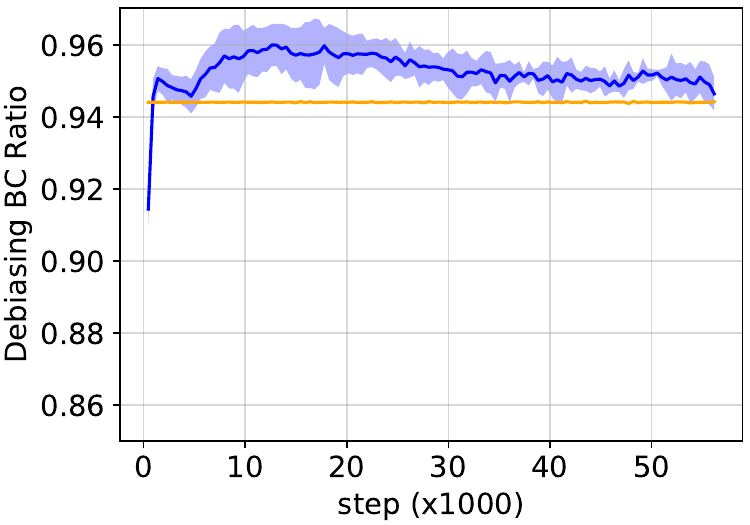}\tabularnewline
\multirow{1}{0.06\textwidth}[0.185\textwidth]{Sample Weight BA} & \includegraphics[width=0.4\textwidth]{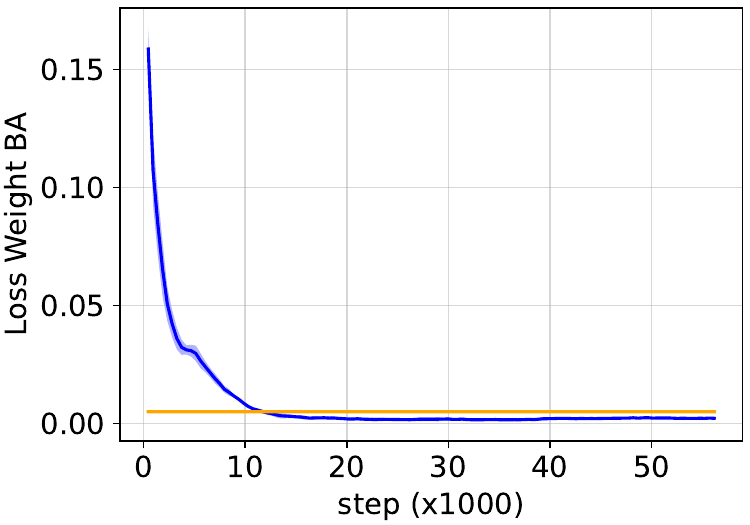} & \includegraphics[width=0.4\textwidth]{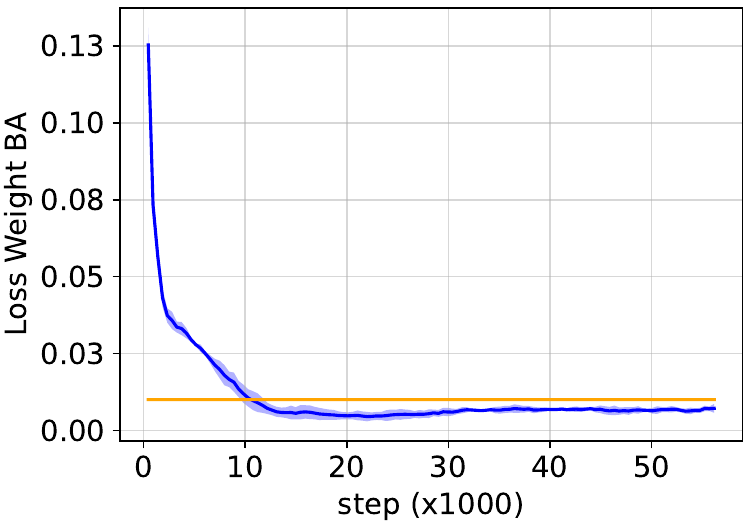} & \includegraphics[width=0.4\textwidth]{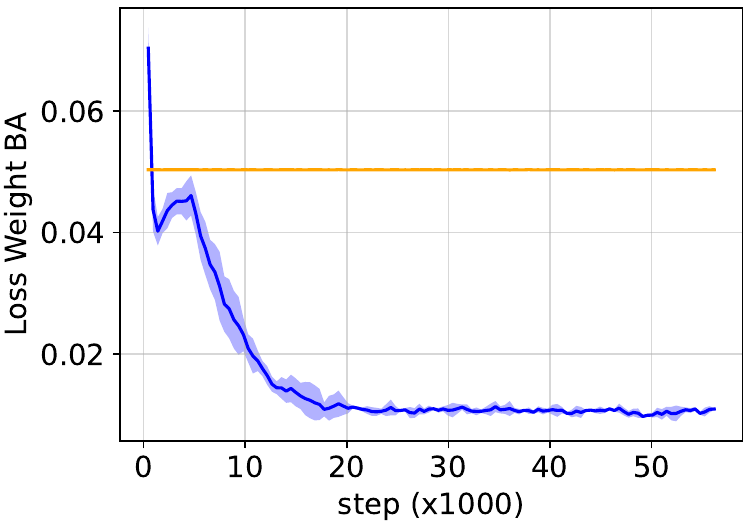}\tabularnewline
\multirow{1}{0.06\textwidth}[0.185\textwidth]{Sample Weight BC} & \includegraphics[width=0.4\textwidth]{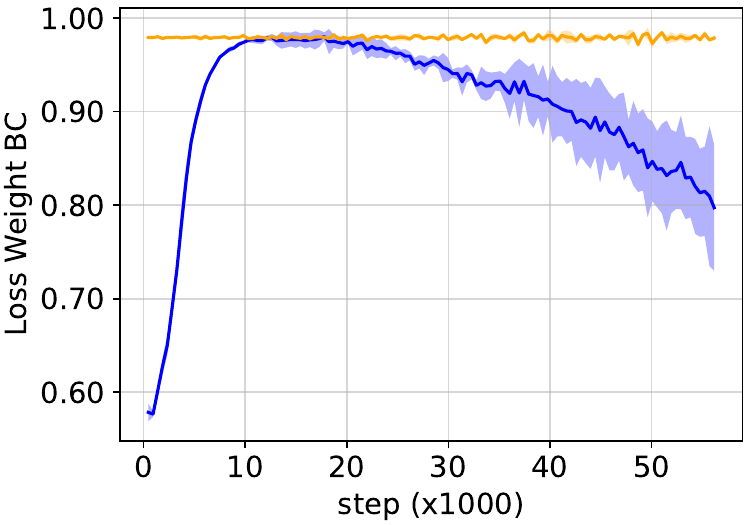} & \includegraphics[width=0.4\textwidth]{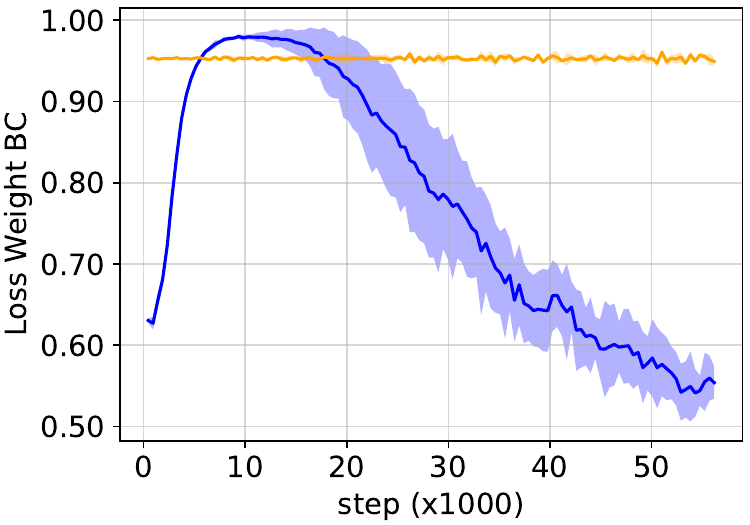} & \includegraphics[width=0.4\textwidth]{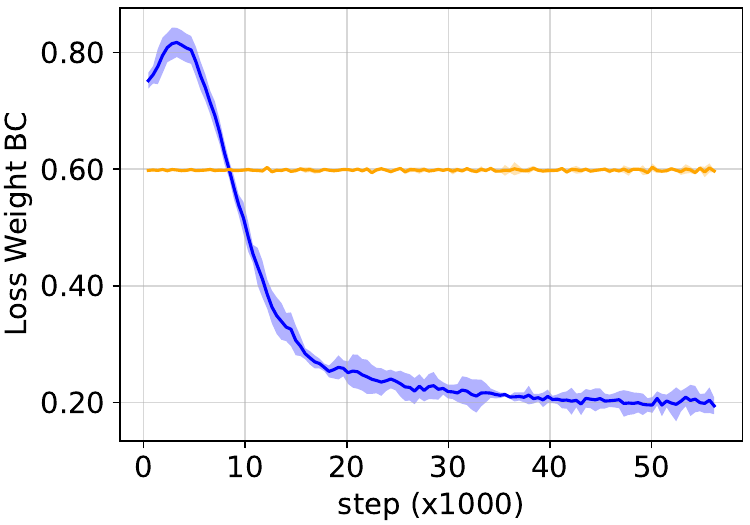}\tabularnewline
 & \multicolumn{3}{c}{\includegraphics[width=0.3\textwidth]{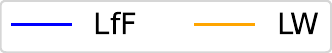}}\tabularnewline
\end{tabular}}
\par\end{centering}
\caption{Learning curves of LfF and LW on Colored MNIST. The sample weights
of LW, originally in the rage $\left[\frac{10}{\gamma},10\right]$,
are scaled to the range $\left[\frac{1}{\gamma},1\right]$ to match
the range {[}0, 1{]} of the sample weights of LfF. \label{fig:LfF-results-MNIST}}
\end{figure*}

\begin{figure*}
\begin{centering}
\resizebox{\textwidth}{!}{%
\par\end{centering}
\begin{centering}
\begin{tabular}{>{\raggedright}p{0.06\textwidth}ccc}
 & 0.5\% & 1\% & 5\%\tabularnewline
\multirow{1}{0.06\textwidth}[0.185\textwidth]{Test Acc. BA} & \includegraphics[width=0.4\textwidth]{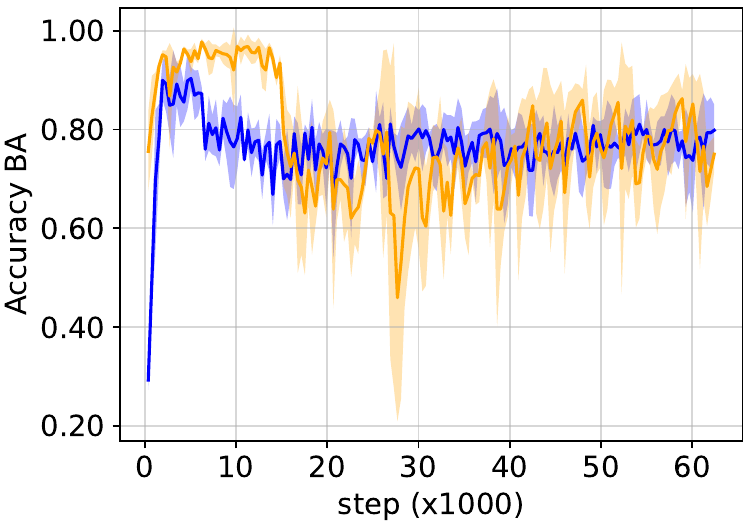} & \includegraphics[width=0.4\textwidth]{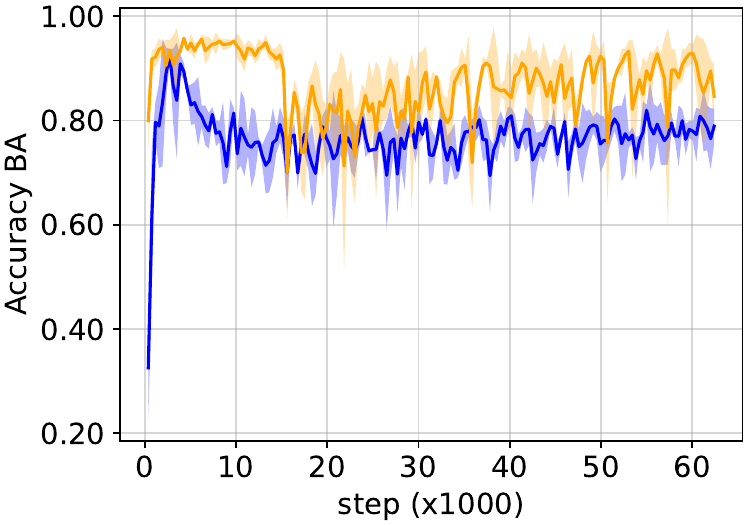} & \includegraphics[width=0.4\textwidth]{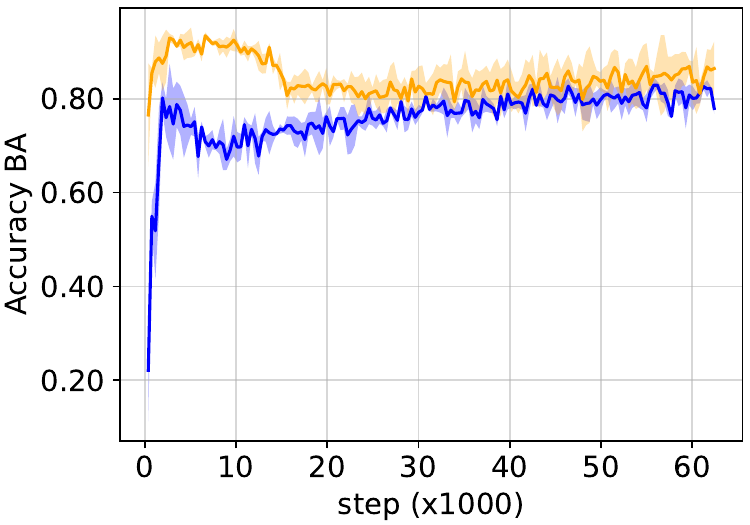}\tabularnewline
\multirow{1}{0.06\textwidth}[0.185\textwidth]{Test Acc. BC} & \includegraphics[width=0.4\textwidth]{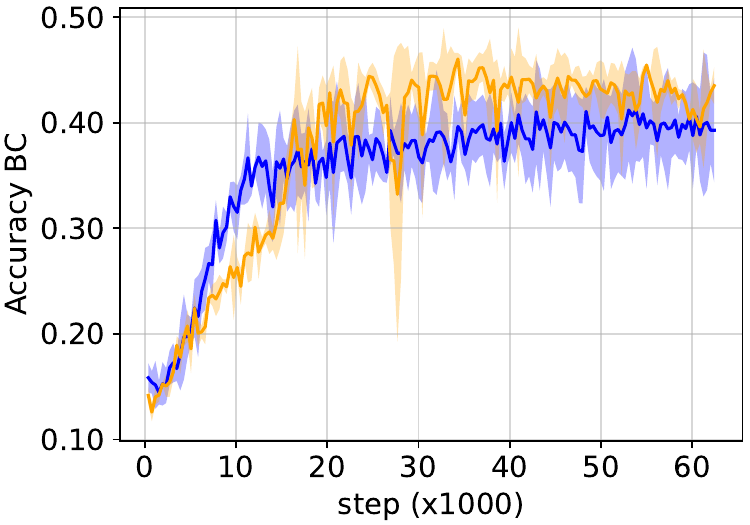} & \includegraphics[width=0.4\textwidth]{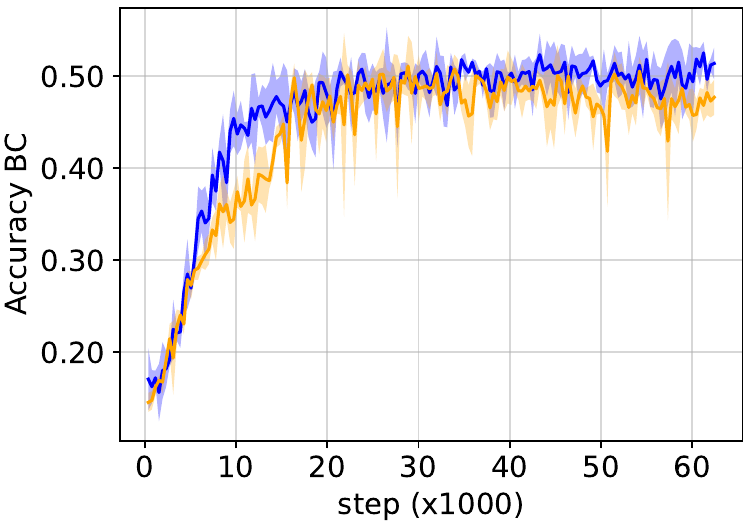} & \includegraphics[width=0.4\textwidth]{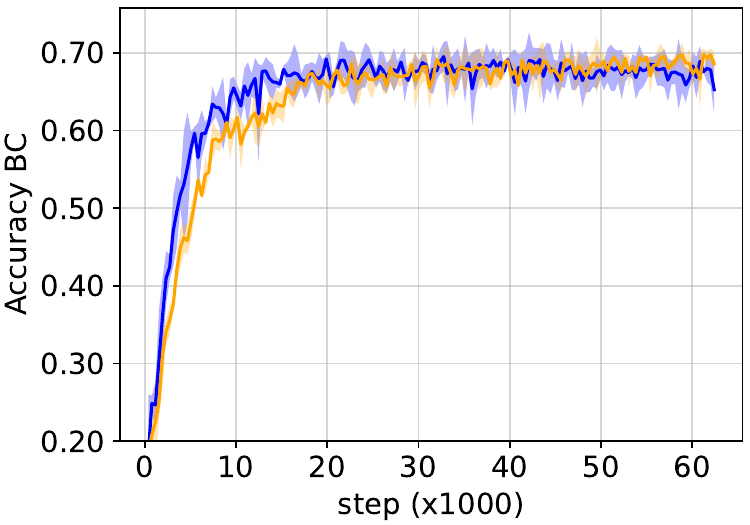}\tabularnewline
\multirow{1}{0.06\textwidth}[0.185\textwidth]{Debias. BC Ratio} & \includegraphics[width=0.4\textwidth]{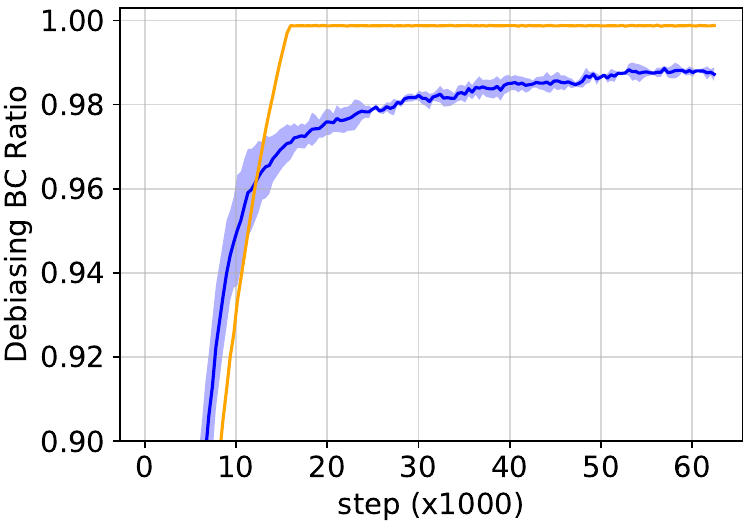} & \includegraphics[width=0.4\textwidth]{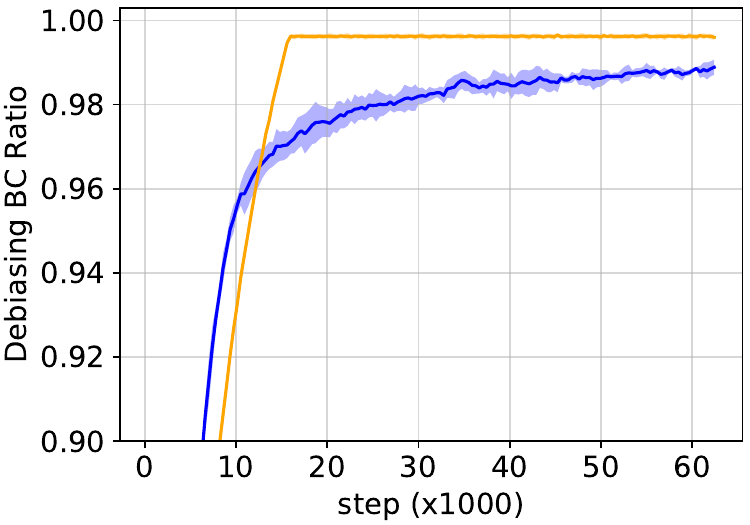} & \includegraphics[width=0.4\textwidth]{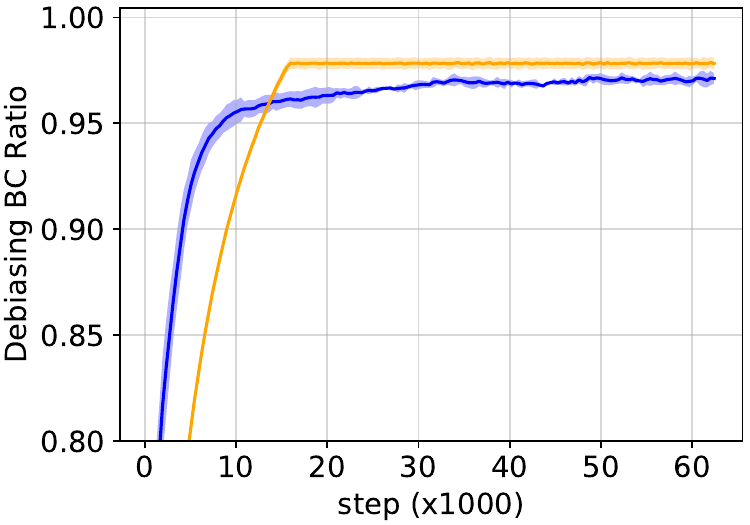}\tabularnewline
\multirow{1}{0.06\textwidth}[0.185\textwidth]{Sample Weight BA} & \includegraphics[width=0.4\textwidth]{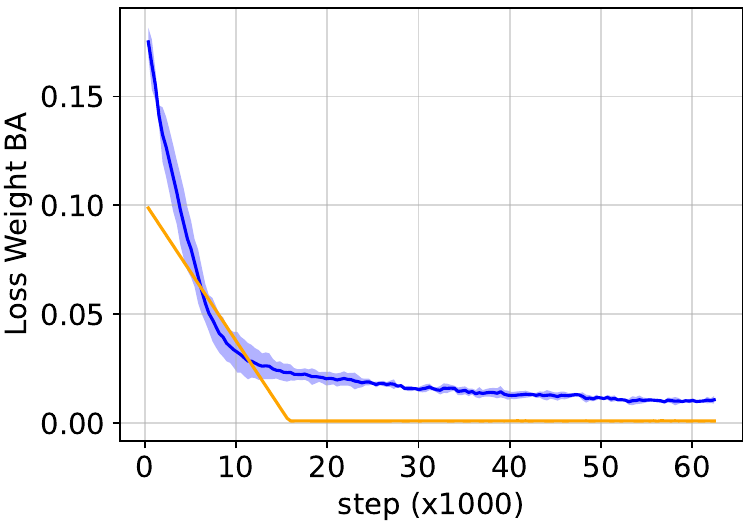} & \includegraphics[width=0.4\textwidth]{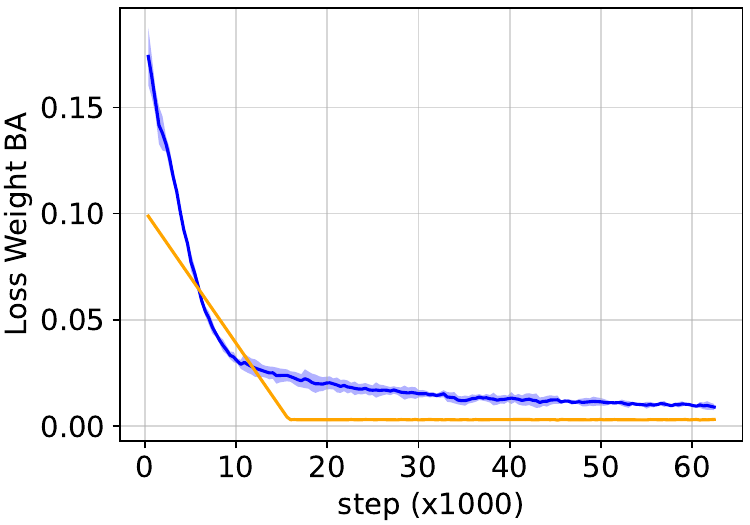} & \includegraphics[width=0.4\textwidth]{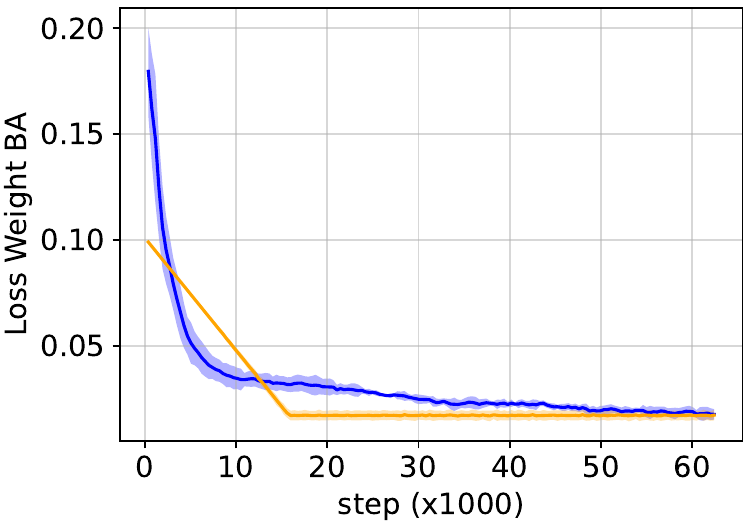}\tabularnewline
\multirow{1}{0.06\textwidth}[0.185\textwidth]{Sample Weight BC} & \includegraphics[width=0.4\textwidth]{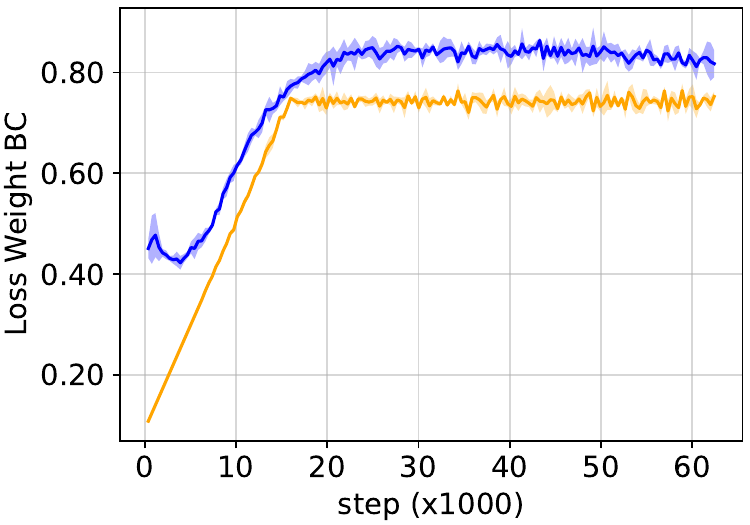} & \includegraphics[width=0.4\textwidth]{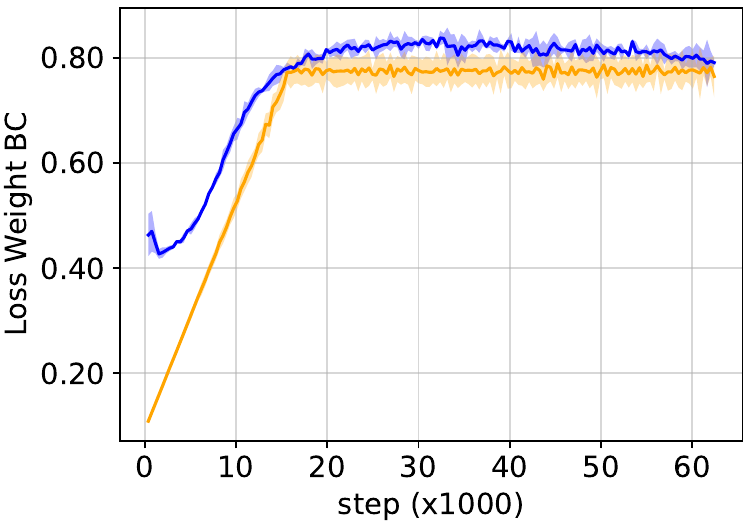} & \includegraphics[width=0.4\textwidth]{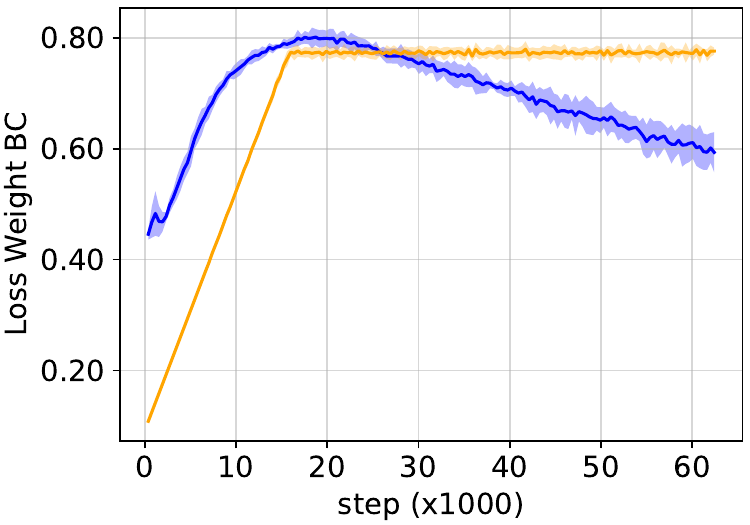}\tabularnewline
 & \multicolumn{3}{c}{\includegraphics[width=0.3\textwidth]{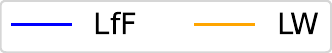}}\tabularnewline
\end{tabular}}
\par\end{centering}
\caption{Learning curves of LfF and LW on Corrupted CIFAR10. The sample weights
of LW, originally in the rage $\left[\frac{10}{\gamma},10\right]$,
are scaled to the range $\left[\frac{1}{\gamma},1\right]$ to match
the range {[}0, 1{]} of the sample weights of LfF.\label{fig:LfF-results-Cifar10}}
\end{figure*}

From Fig.~\ref{fig:LfF-results-MNIST}, we can clearly see that LfF
achieves lower test accuracies than LW for both BA and BC samples
on Colored MNIST. Furthermore, the test accuracy curves of LW exhibit
a steady upward trend over time for both BA and BC samples. In contrast,
the test accuracy curve for BA samples of LfF exhibits a decline followed
by an increase, while the test accuracy curve for BC samples of LfF
shows an increase followed by a decrease. This behavior aligns with
the dynamics of the debiasing BC ratio of LfF (row 3 in Fig.~\ref{fig:LfF-results-MNIST}),
which initially increases during the early stages of training and
then decreases. This can be attributed to the fact that the biased
classifier achieves a faster reduction in cross-entropy loss for BC
samples compared to BA samples as the training progresses (rows 4,
5 in \ref{fig:LfF-results-MNIST}), resulting in smaller sample weights
for BC samples at a faster rate than those for BA samples. However,
on Corrupted CIFAR10, the sample weights for BC samples decrease at
a slower rate than those for BA samples, making the debiasing BC ratios
of LfF increase over time (rows 4, 5 in \ref{fig:LfF-results-Cifar10}).
A larger debiasing BC ratio often leads to higher test accuracy for
BC samples since this kind of samples are assigned with larger weights
during training.

\subsection{Additional results of PGD in comparison with LW\label{subsec:Additional-results-of-PGD}}

\begin{figure*}
\begin{centering}
\resizebox{\textwidth}{!}{%
\par\end{centering}
\begin{centering}
\begin{tabular}{>{\raggedright}p{0.06\textwidth}ccc}
 & 0.5\% & 1\% & 5\%\tabularnewline
\multirow{1}{0.06\textwidth}[0.185\textwidth]{Test Acc. BA} & \includegraphics[width=0.4\textwidth]{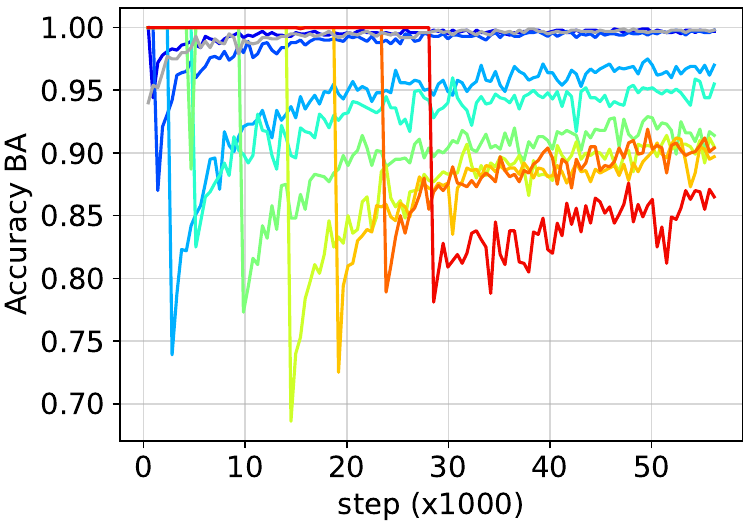} & \includegraphics[width=0.4\textwidth]{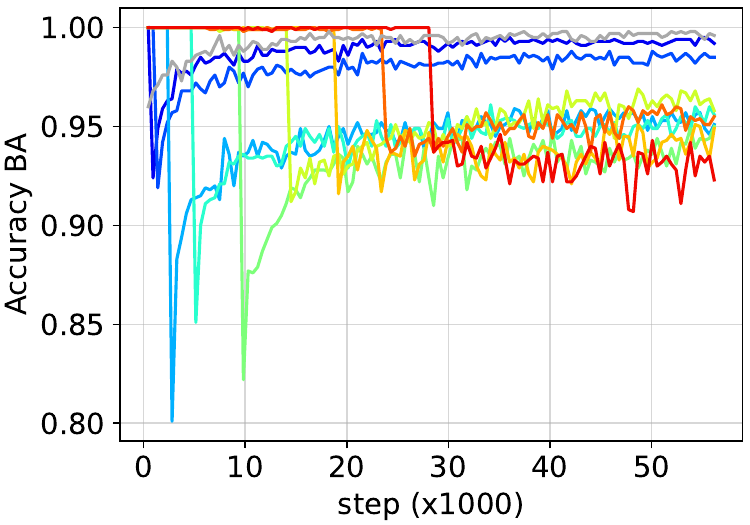} & \includegraphics[width=0.4\textwidth]{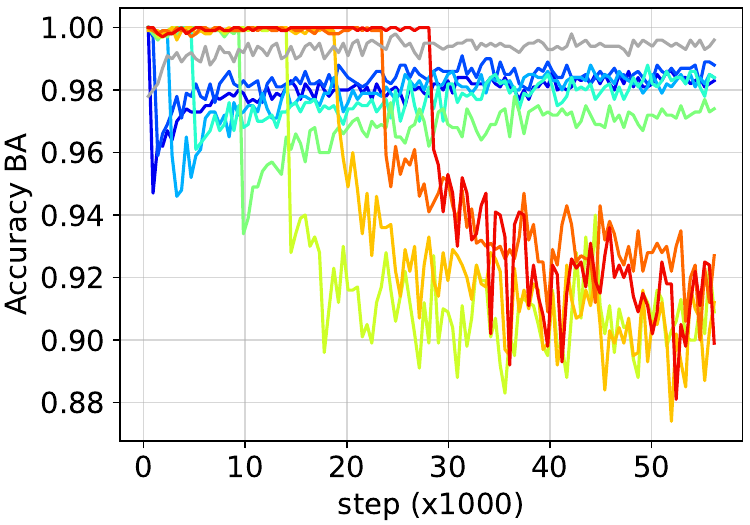}\tabularnewline
\multirow{1}{0.06\textwidth}[0.185\textwidth]{Test Acc. BC} & \includegraphics[width=0.4\textwidth]{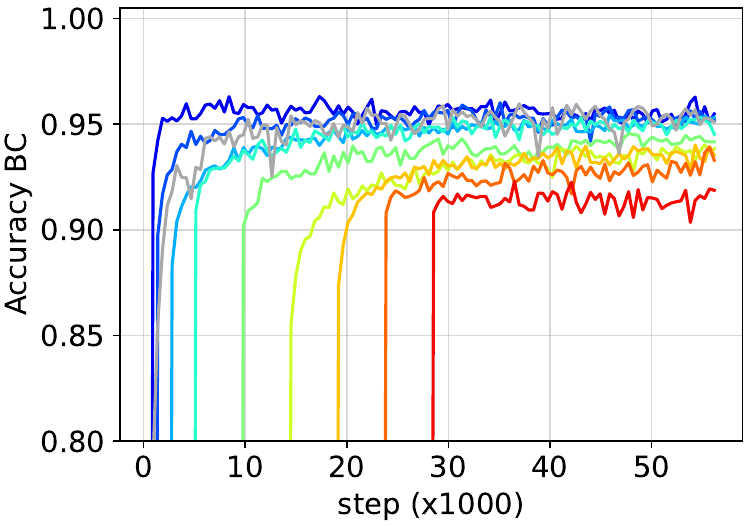} & \includegraphics[width=0.4\textwidth]{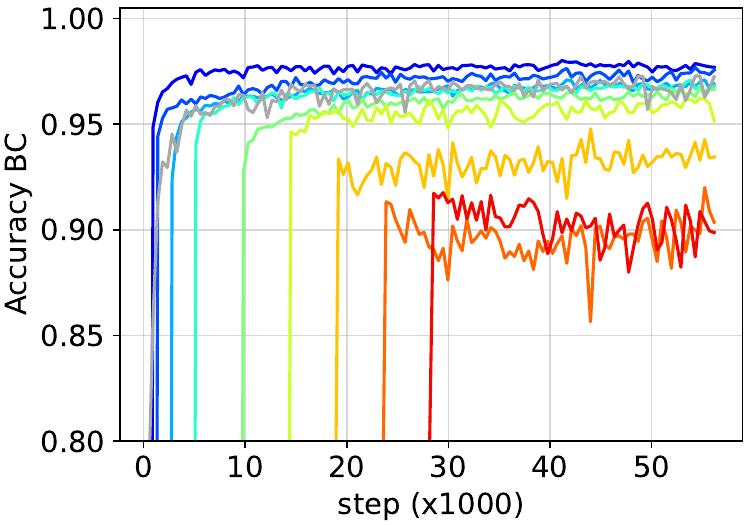} & \includegraphics[width=0.4\textwidth]{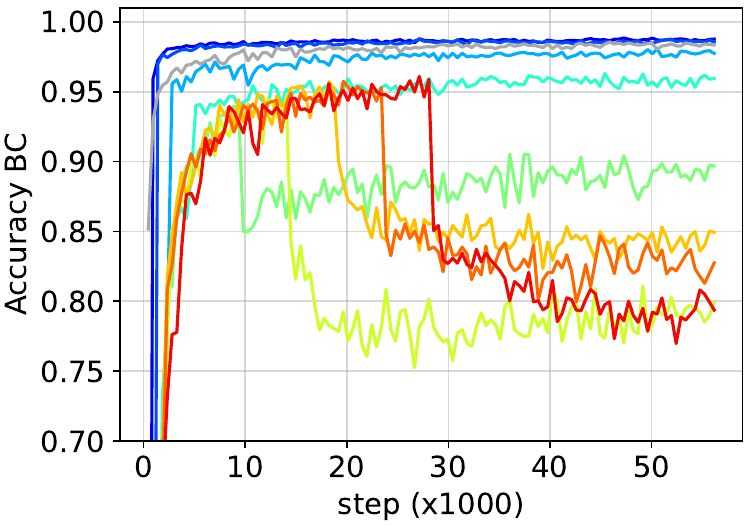}\tabularnewline
\multirow{1}{0.06\textwidth}[0.185\textwidth]{Test Xent BA} & \includegraphics[width=0.4\textwidth]{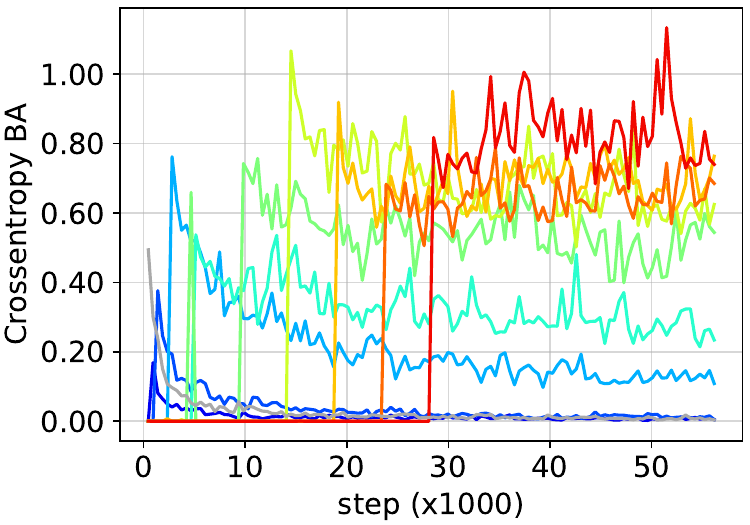} & \includegraphics[width=0.4\textwidth]{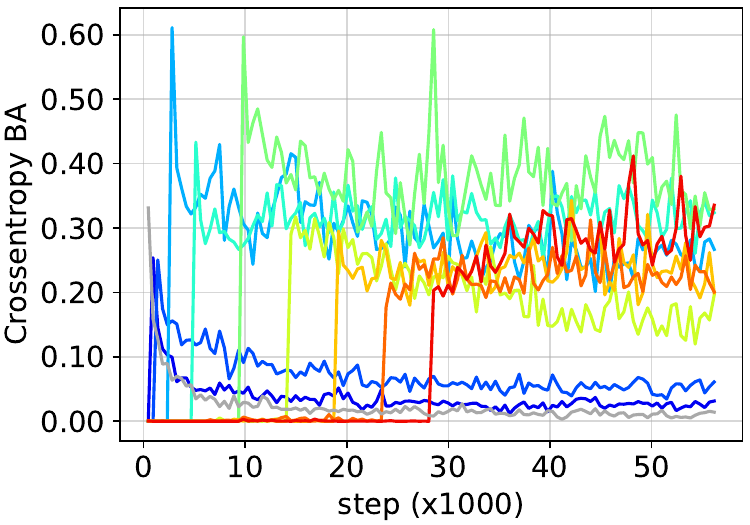} & \includegraphics[width=0.4\textwidth]{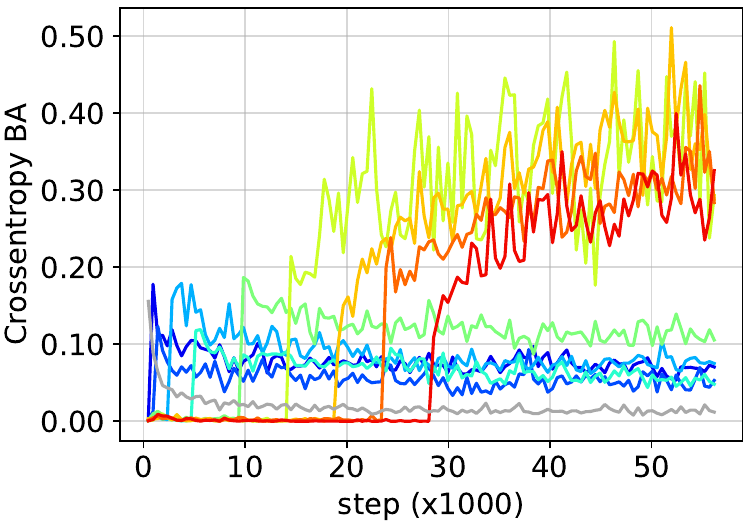}\tabularnewline
\multirow{1}{0.06\textwidth}[0.185\textwidth]{Test Xent BC} & \includegraphics[width=0.4\textwidth]{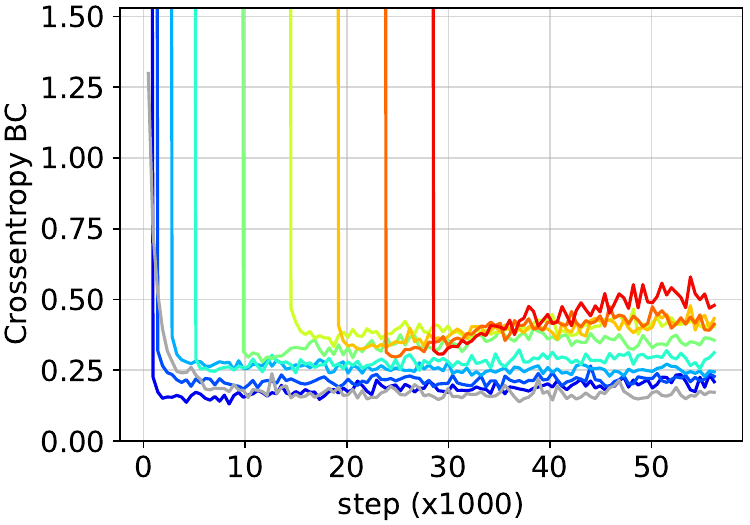} & \includegraphics[width=0.4\textwidth]{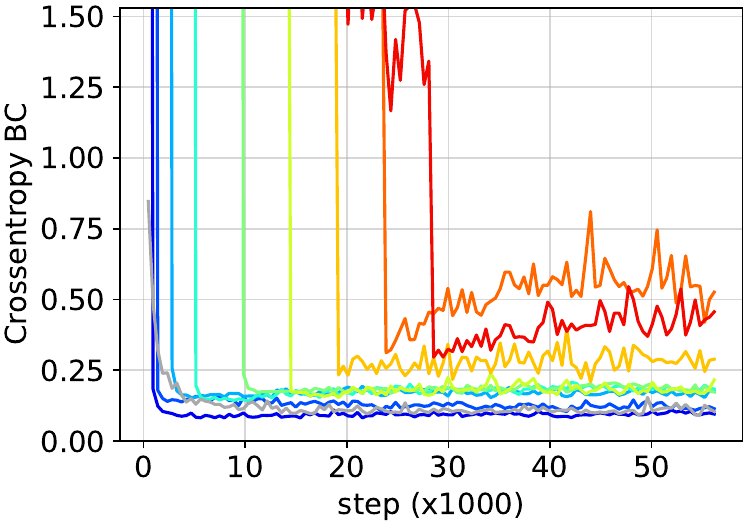} & \includegraphics[width=0.4\textwidth]{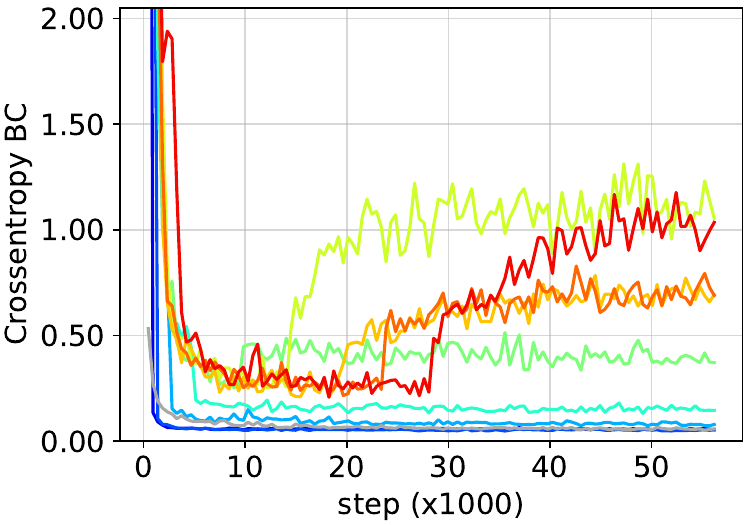}\tabularnewline
 & \multicolumn{3}{c}{\includegraphics[width=0.6\textwidth]{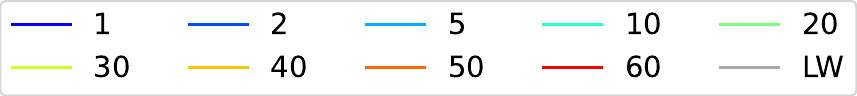}}\tabularnewline
\end{tabular}}
\par\end{centering}
\caption{Learning curves of PGD on Colored MNIST by varying the number of training
epochs for the biased classifier. The abrupt changes observed in the
above figures indicate the transition from biased classifier training
to debiased classifier training. We also show the learning curves
of our proposed LW in gray for convenient comparison.\label{fig:ablation-PGD-mnist}}
\end{figure*}

\begin{figure*}
\begin{centering}
\resizebox{\textwidth}{!}{%
\par\end{centering}
\begin{centering}
\begin{tabular}{>{\raggedright}p{0.06\textwidth}ccc}
 & 0.5\% & 1\% & 5\%\tabularnewline
\multirow{1}{0.06\textwidth}[0.185\textwidth]{Test Acc. BA} & \includegraphics[width=0.4\textwidth]{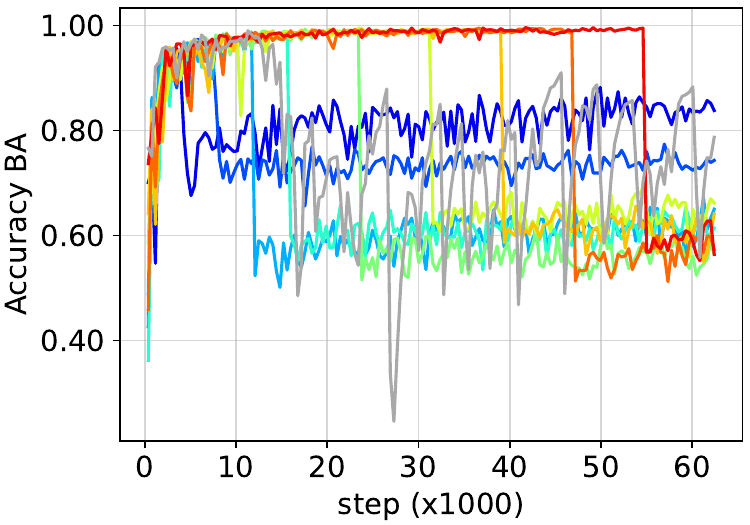} & \includegraphics[width=0.4\textwidth]{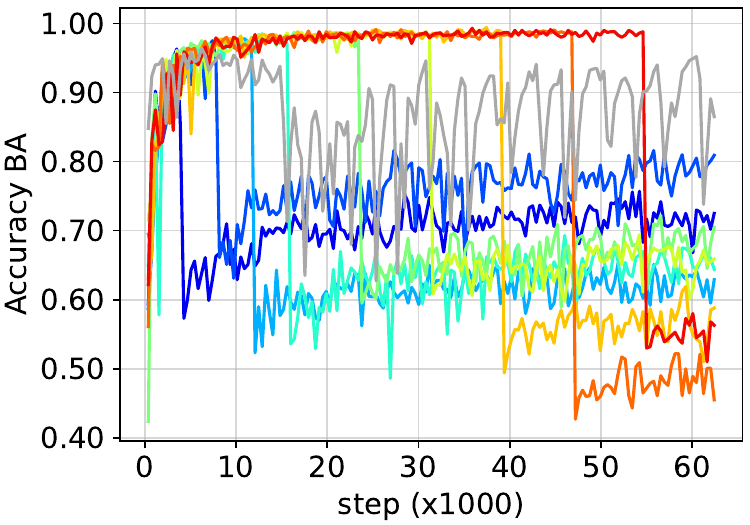} & \includegraphics[width=0.4\textwidth]{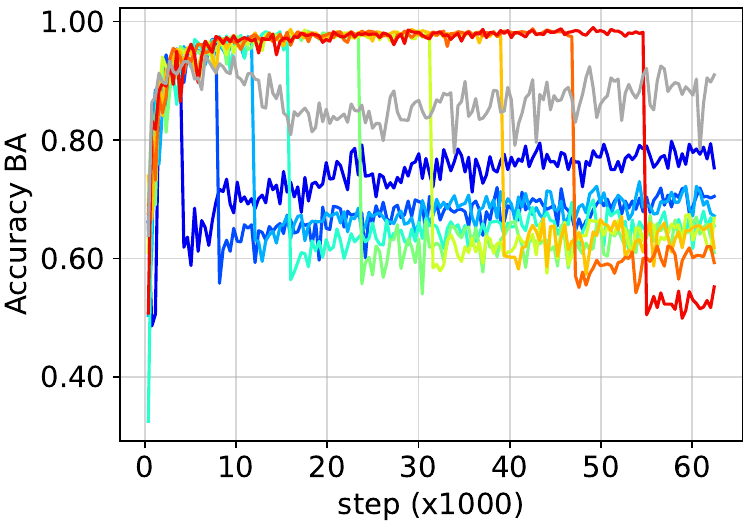}\tabularnewline
\multirow{1}{0.06\textwidth}[0.185\textwidth]{Test Acc. BC} & \includegraphics[width=0.4\textwidth]{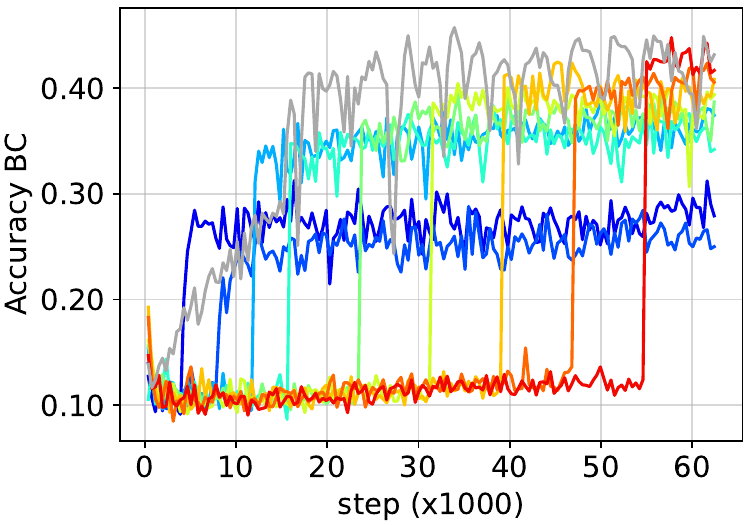} & \includegraphics[width=0.4\textwidth]{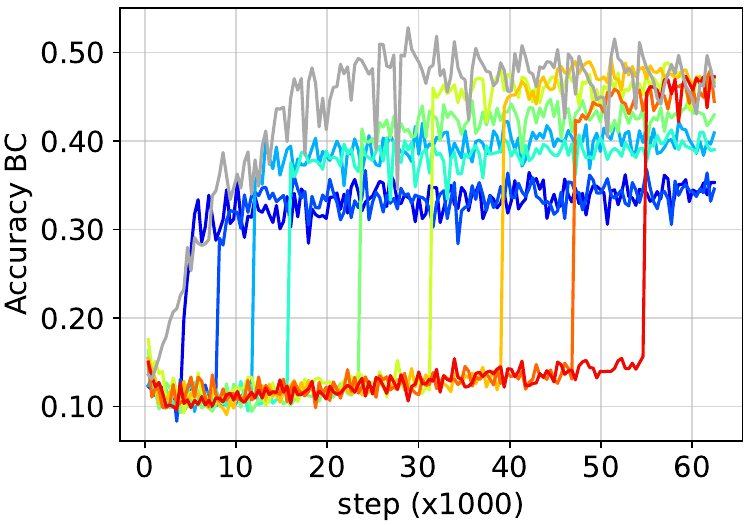} & \includegraphics[width=0.4\textwidth]{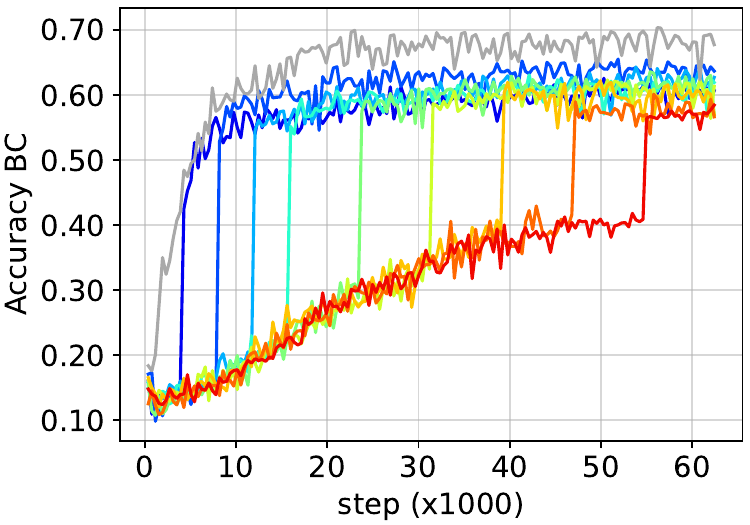}\tabularnewline
\multirow{1}{0.06\textwidth}[0.185\textwidth]{Test Xent BA} & \includegraphics[width=0.4\textwidth]{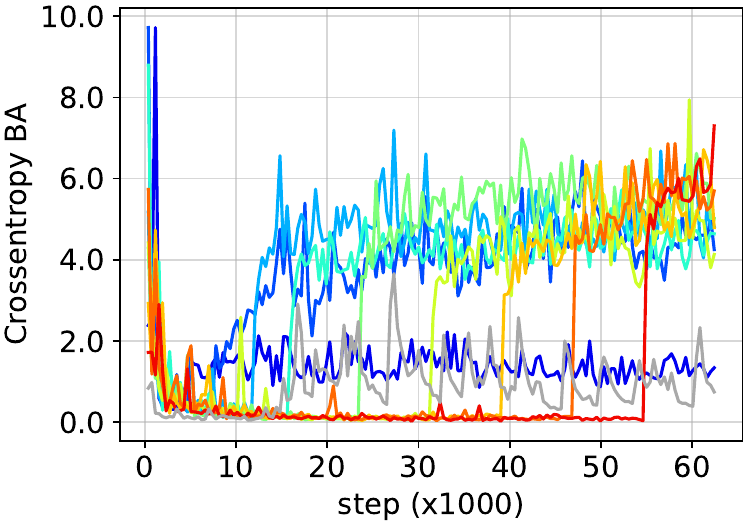} & \includegraphics[width=0.4\textwidth]{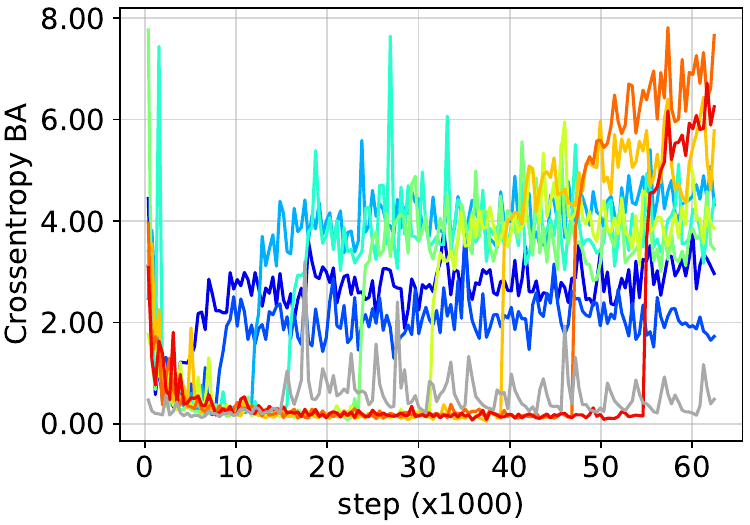} & \includegraphics[width=0.4\textwidth]{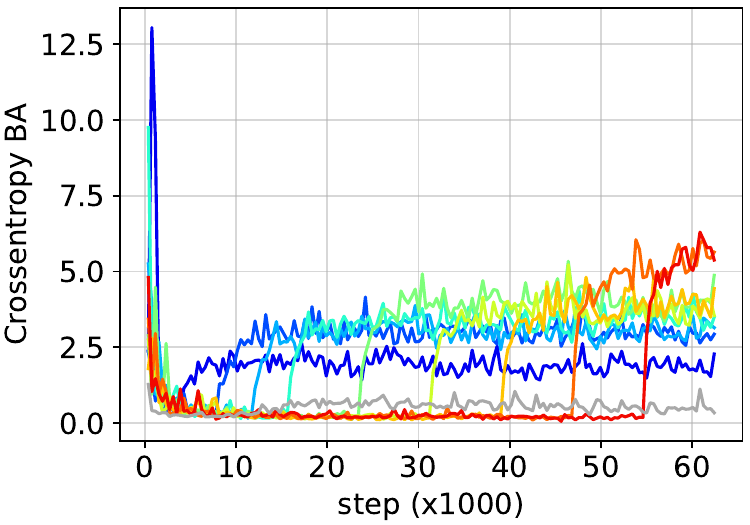}\tabularnewline
\multirow{1}{0.06\textwidth}[0.185\textwidth]{Test Xent BC} & \includegraphics[width=0.4\textwidth]{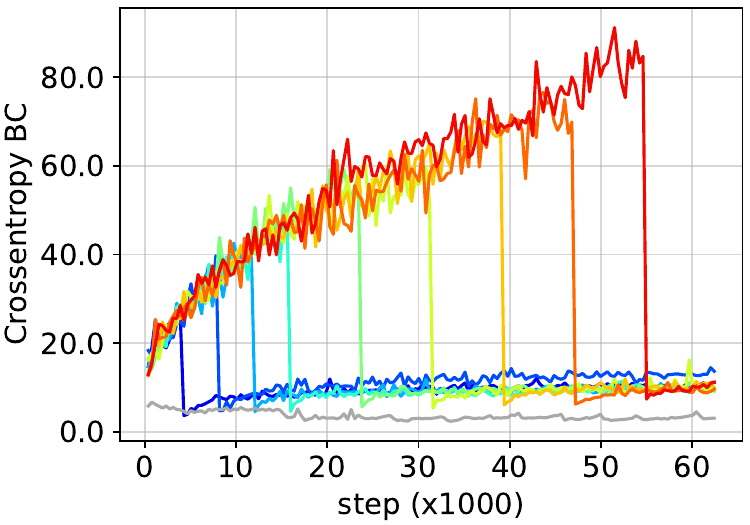} & \includegraphics[width=0.4\textwidth]{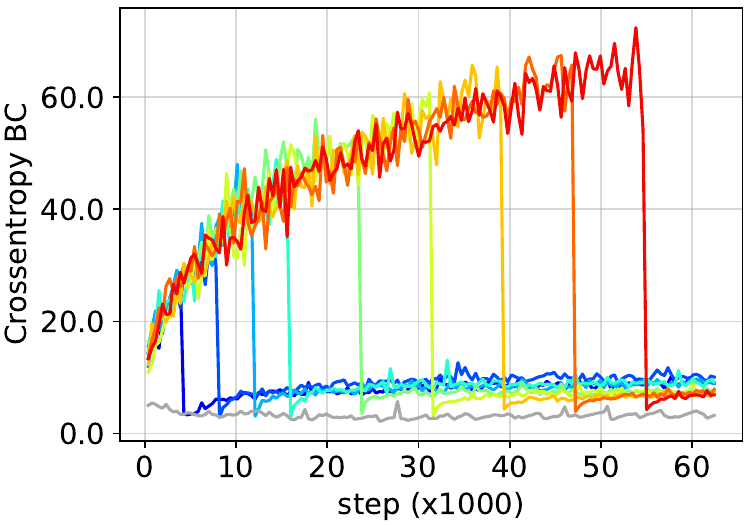} & \includegraphics[width=0.4\textwidth]{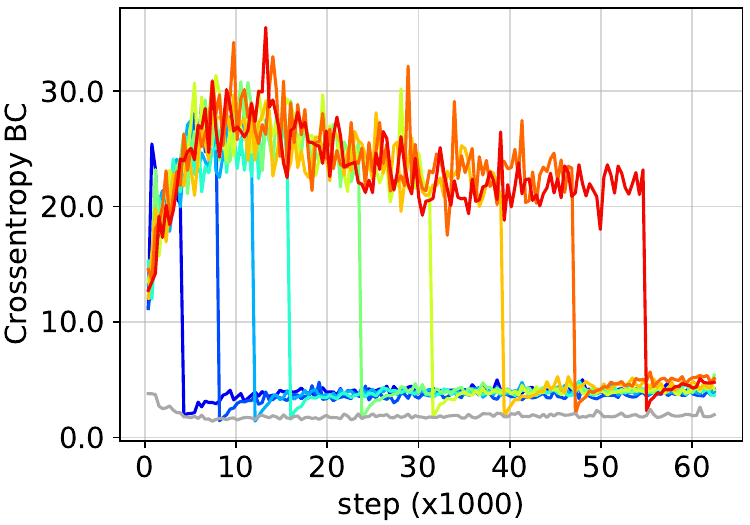}\tabularnewline
 & \multicolumn{3}{c}{\includegraphics[width=0.6\textwidth]{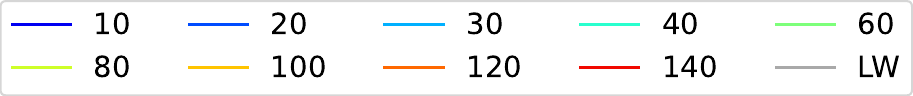}}\tabularnewline
\end{tabular}}
\par\end{centering}
\caption{Learning curves of PGD on Corrupted CIFAR10 by varying the number
of training epochs for the biased classifier. The abrupt changes observed
in the above figures indicate the transition from biased classifier
training to debiased classifier training. We also show the learning
curves of our proposed LW in gray for convenient comparison.\label{fig:ablation-PGD-cifar10}}
\end{figure*}

From Fig.~\ref{fig:ablation-PGD-mnist}, it is clear that PGD achieves
the best test accuracies for both BA and BC samples on Colored MNIST
when the biased classifier is trained for only one epoch. This holds
true for different values of the BC ratio. However, on Corrupted CIFAR10,
the best performance of PGD is obtained when the biased classifier
is trained with a large number of epochs (can be up to hundreds) as
shown in Fig.~\ref{fig:ablation-PGD-cifar10}. Such inconsistency
makes the analysis and real-world application of PGD difficult. In
almost all settings, we observe that PGD achieves higher cross-entropy
losses than our method for both BA and BC samples, suggesting that
the debiased classifier learned by PGD is less stable than the counterpart
learned by our method. We hypothesize that it is because the debiased
classifier of PGD is derived from the biased classifier, thus, may
inherit the instability of the biased classifier.

\subsection{Analysis of the Biased Classifier\label{subsec:Analysis-of-the-Biased-Classifier}}

\begin{figure*}
\begin{centering}
\resizebox{\textwidth}{!}{%
\par\end{centering}
\begin{centering}
\begin{tabular}{>{\raggedright}p{0.06\textwidth}ccc}
 & Colored MNIST & Corrupted CIFAR10 & Biased CelebA\tabularnewline
\multirow{1}{0.06\textwidth}[0.185\textwidth]{Test Acc. BA} & \includegraphics[width=0.4\textwidth]{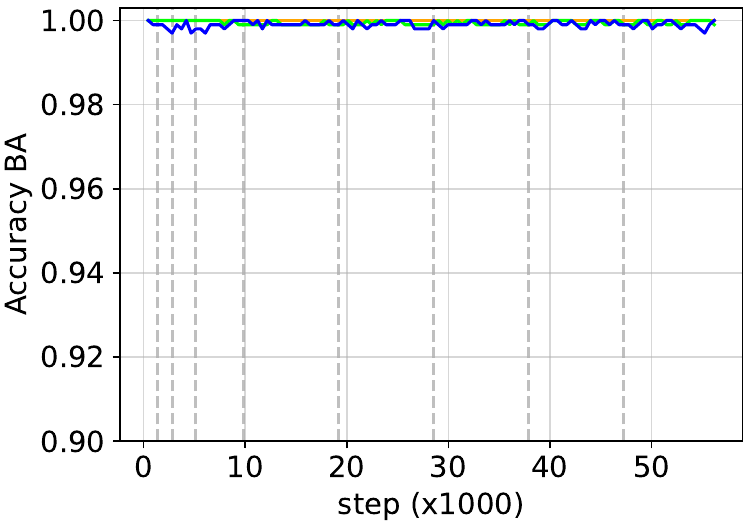} & \includegraphics[width=0.4\textwidth]{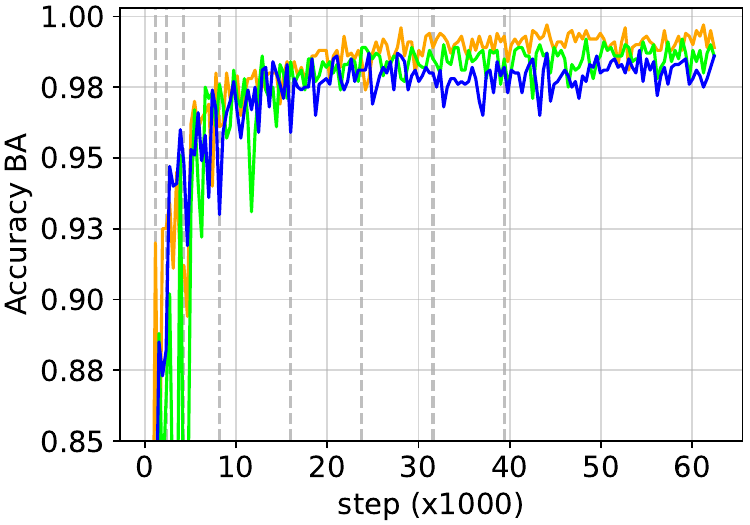} & \includegraphics[width=0.4\textwidth]{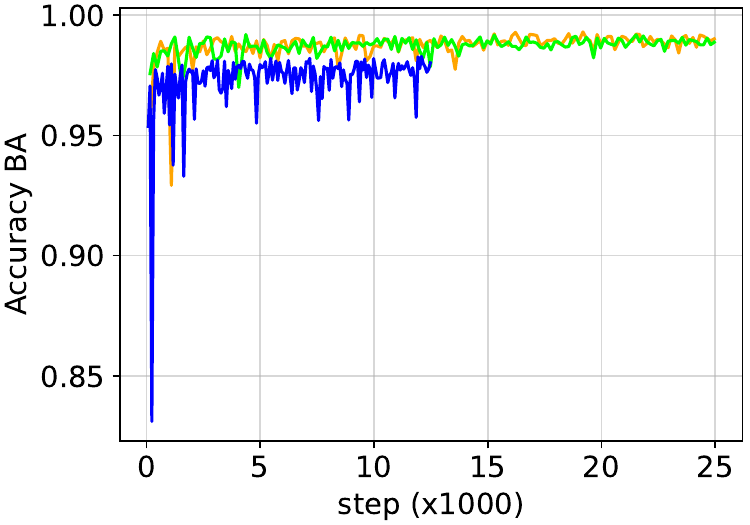}\tabularnewline
\multirow{1}{0.06\textwidth}[0.185\textwidth]{Test Acc. BC} & \includegraphics[width=0.4\textwidth]{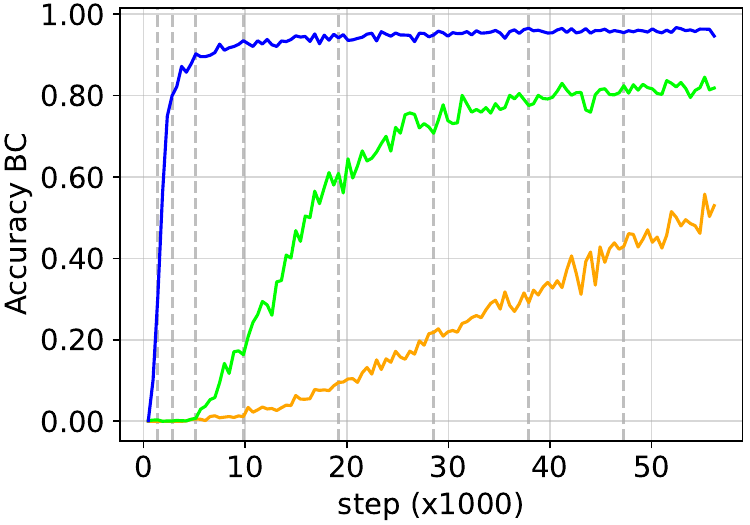} & \includegraphics[width=0.4\textwidth]{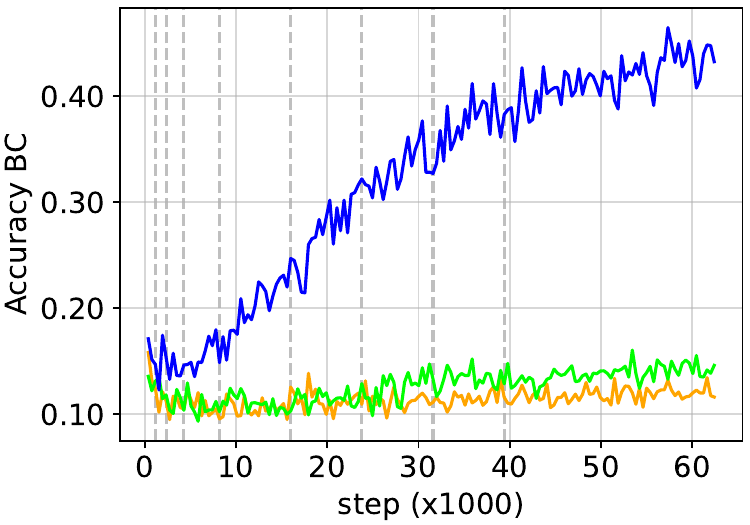} & \includegraphics[width=0.4\textwidth]{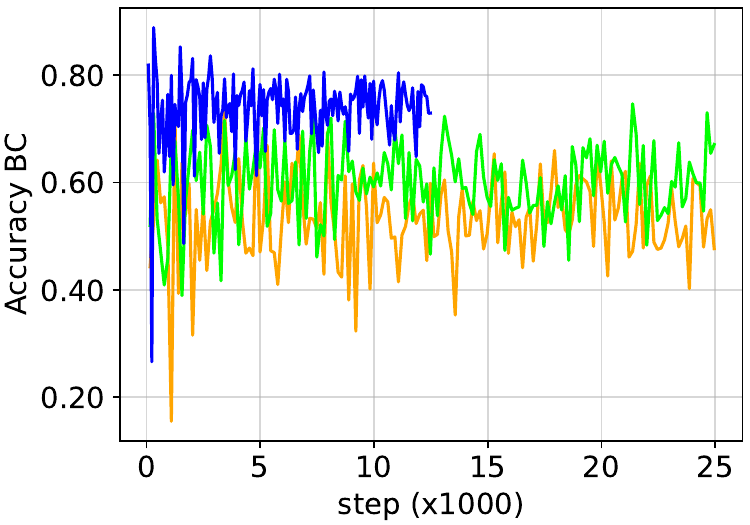}\tabularnewline
\multirow{1}{0.06\textwidth}[0.185\textwidth]{Train Acc. BA} & \includegraphics[width=0.4\textwidth]{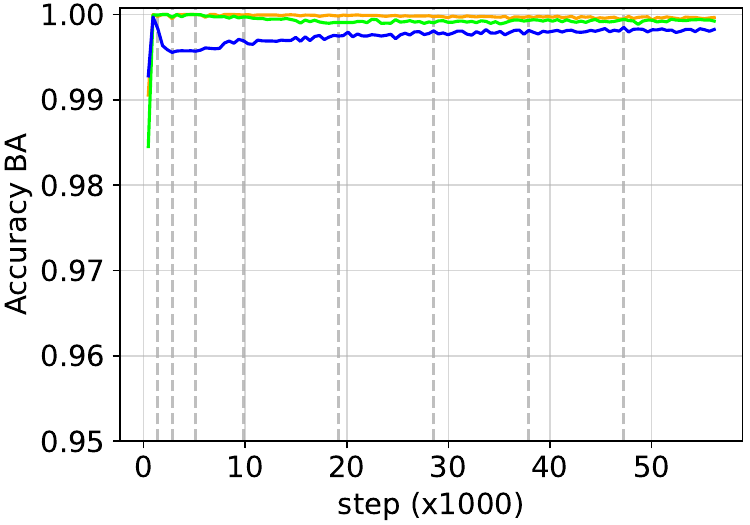} & \includegraphics[width=0.4\textwidth]{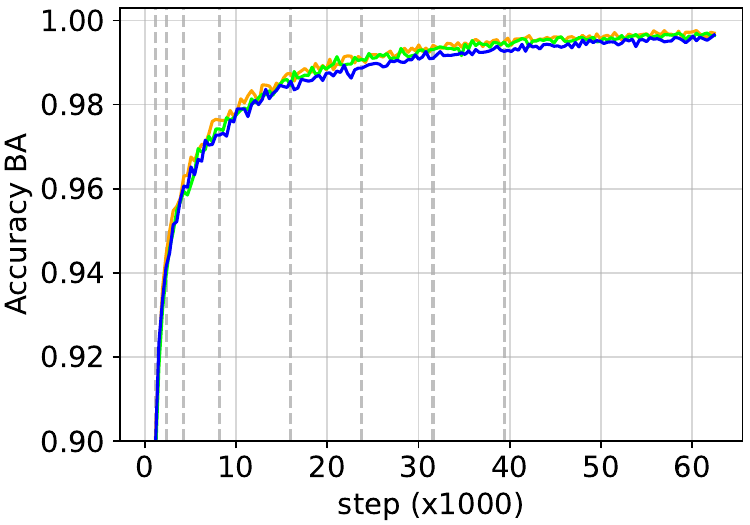} & \includegraphics[width=0.4\textwidth]{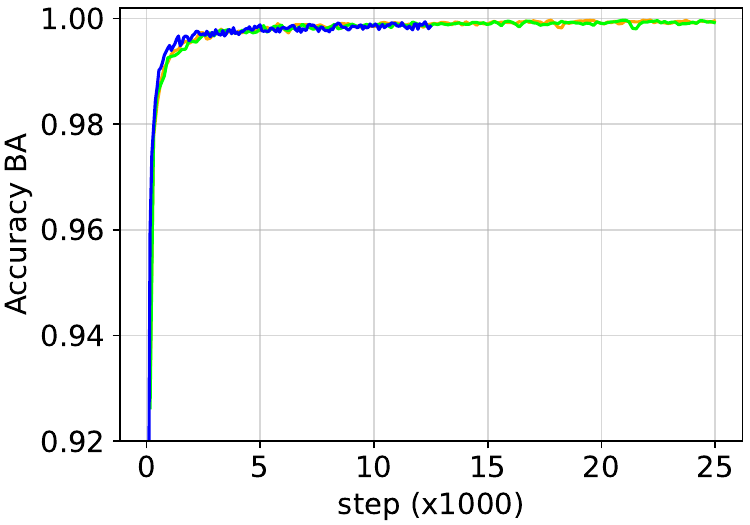}\tabularnewline
\multirow{1}{0.06\textwidth}[0.185\textwidth]{Train Acc. BC} & \includegraphics[width=0.4\textwidth]{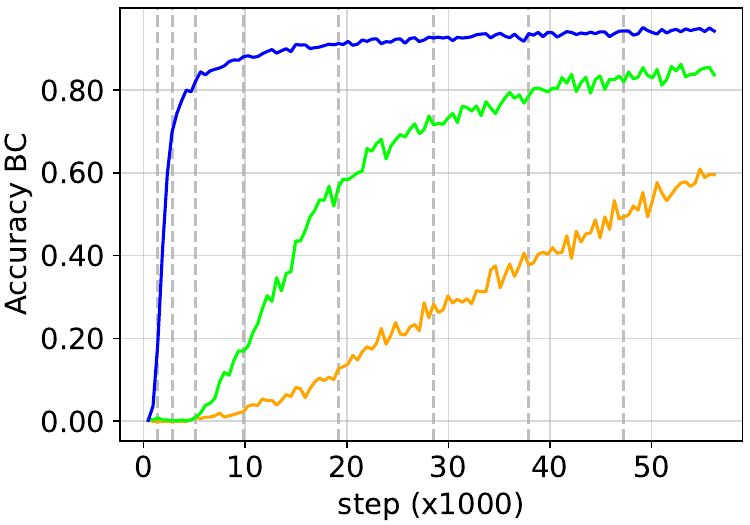} & \includegraphics[width=0.4\textwidth]{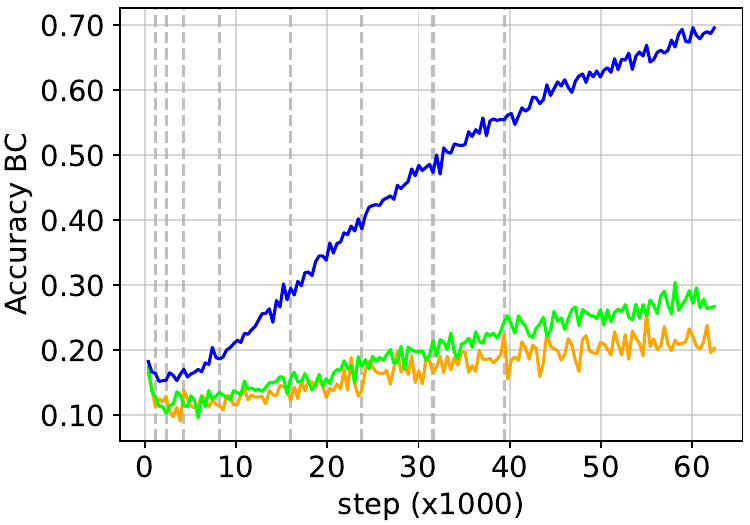} & \includegraphics[width=0.4\textwidth]{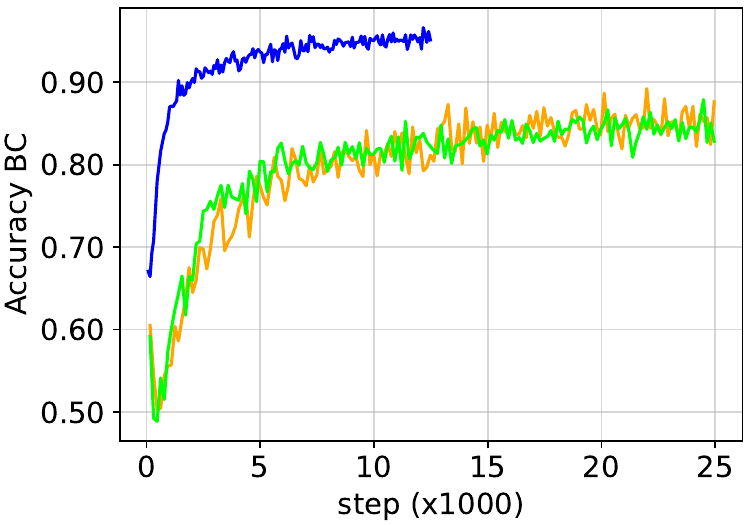}\tabularnewline
\multirow{1}{0.06\textwidth}[0.185\textwidth]{Train Xent BA} & \includegraphics[width=0.4\textwidth]{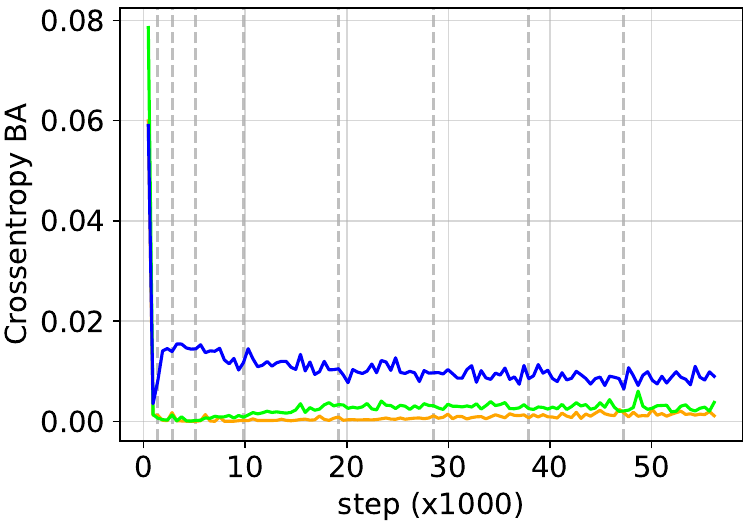} & \includegraphics[width=0.4\textwidth]{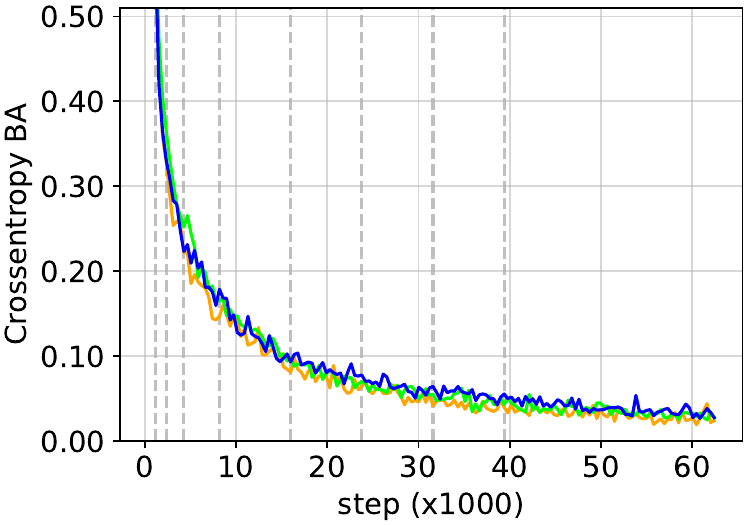} & \includegraphics[width=0.4\textwidth]{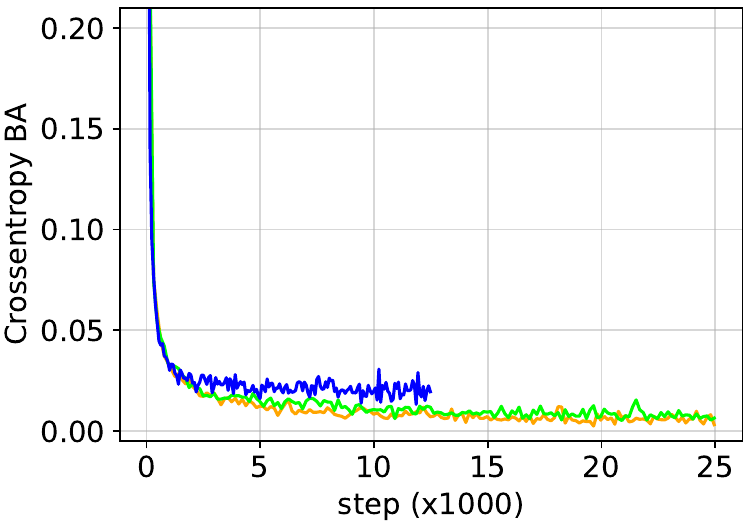}\tabularnewline
\multirow{1}{0.06\textwidth}[0.185\textwidth]{Train Xent BC} & \includegraphics[width=0.4\textwidth]{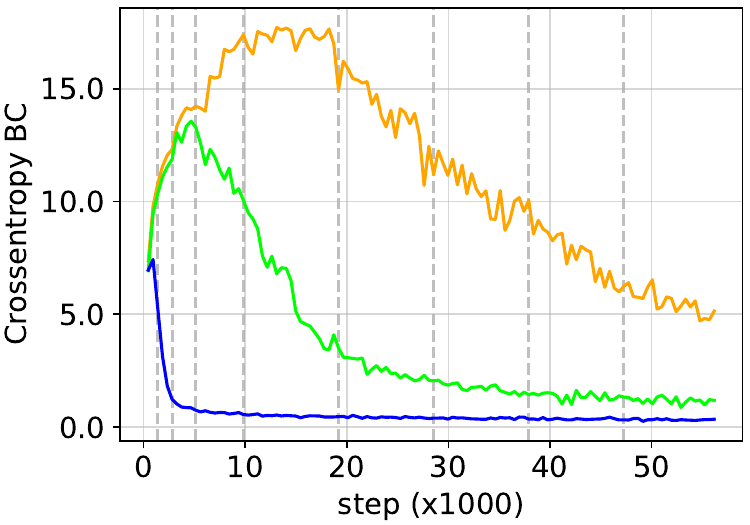} & \includegraphics[width=0.4\textwidth]{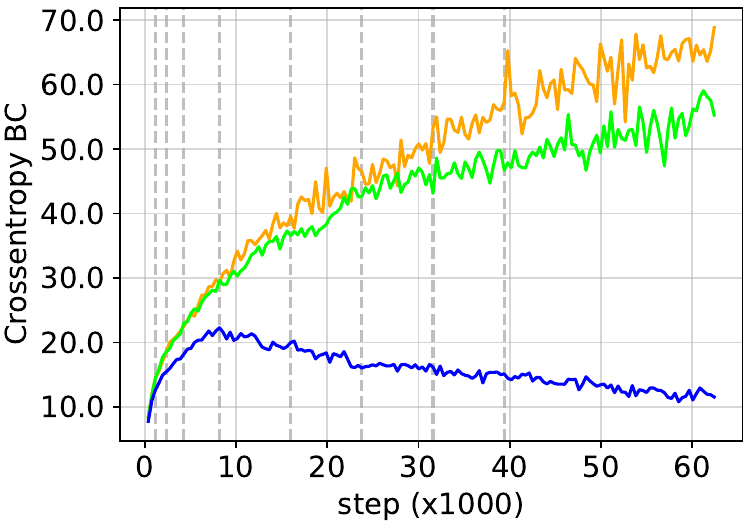} & \includegraphics[width=0.4\textwidth]{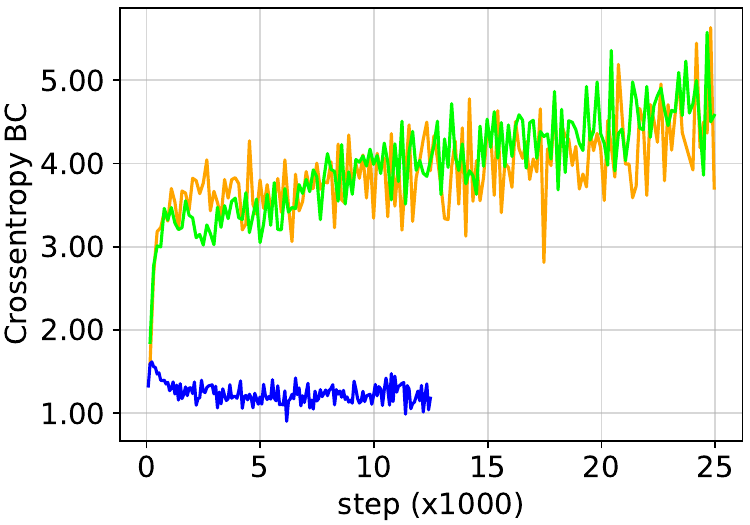}\tabularnewline
 & \multicolumn{3}{c}{\includegraphics[width=0.45\textwidth]{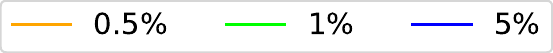}}\tabularnewline
\end{tabular}}
\par\end{centering}
\caption{Learning curves of the biased classifier trained on Colored MNIST,
Corrupted CIFAR10, and Biased CelebA with varying BC ratios of 0.5\%,
1\%, 2\%, and 5\%. The dashed lines in each plot, arranged from left
to right, indicate the training epochs of 2, 5, 10, 20, 40, 60, 80,
and 100.\label{fig:biased-classifier-curves}}
\end{figure*}

In Fig.~\ref{fig:biased-classifier-curves}, we show the learning
curves of the biased classifier $p_{\psi}(y|x)$ trained on Colored
MNIST, Corrupted CIFAR10, and Biased CelebA, considering various BC
ratios. Apparently, the biased classifier exhibits higher test accuracy
with prolonged training, indicating the necessity of employing an
early stopping strategy in some cases to prevent the biased classifier
from learning class semantics. For example, on Colored MNIST with
a BC ratio of 5\%, the biased classifier can achieve about 90\% test
accuracy after just 10 training epochs. Therefore, for Colored MNIST
with a BC ratio of 5\%, we need to set the training epoch of the biased
classifier to 1 to ensure that $p_{\psi}(y|x)$ is a good approximation
of $p_{\psi}(y|b)$ as shown in Table~\ref{tab:Optimal-settings}.
On Corrupted CIFAR10 with BC ratios of 0.5\% and 1\%, the test accuracy
of the biased classifier does not show significant improvement even
when trained until the final epoch (160). This clarifies why, in Table~\ref{tab:Optimal-settings},
the optimal number of training epochs for the biased classifier in
these settings falls within a range rather than being a specific value. 

Another interesting thing that can be seen from Fig.~\ref{fig:biased-classifier-curves}
is the \emph{positive correlation} between the training accuracy and
cross-entropy curves for BC samples on Corrupted CIFAR10, Biased CelebA,
and Colored MNIST\footnote{For Colored MNIST, the two curves are only positively correlated during
the early stage of training.}, while normally the two curves should be contradictory. Our conjecture
is that although \emph{more} BC samples are classified correctly by
the biased classifier, \emph{those that are misclassified are assigned
low probabilities}. Because the GCE loss encourages the biased classifier
to ignore samples with low probabilities, the probabilities of the
misclassified samples decrease gradually during training, resulting
in an increase in the cross-entropy loss. It should be noted that
even a single sample misclassified with a probability of zero can
lead to an extremely high cross-entropy loss, as the negative logarithm
of zero approaches infinity ($-\log0$ $\approx$ $\infty$). This
finding provides a better explanation for our observation about the
biased classifier's collapse during training discussed in Section~\ref{subsec:Additional-results-of-LfF}.

\subsection{Target Bias Adjustment - an extension of BiasBal to deal with unknown
bias\label{subsec:Target-Bias-Adjustment}}

Recalling that BiasBal \cite{Hong2021} incorporates the bias in training
data into its classifier via the following adjustment formula:
\[
\tilde{p}_{\theta}(y|x)=p_{\theta}(y|x)p(y|b)
\]
where $p_{\theta}(y|x)$ and $\tilde{p}_{\theta}(y|x)$ are the original
and adjusted outputs of the classifier, respectively. However, since
$\tilde{p}_{\theta}(y|x)$ is unnormalized under this adjustment formula
(i.e., $\sum_{c=1}^{C}\tilde{p}_{\theta}(y=c|x)\neq1$), we can rewrite
$\tilde{p}_{\theta}(y|x)$ in its normalized form as follows:
\[
\tilde{p}_{\theta}(y=c|x)=\frac{\exp\left(f_{\theta}(x)[c]+\log p(y=c|b)\right)}{\sum_{k=1}^{C}\exp\left(f_{\theta}(x)[k]+\log p(y=k|b)\right)}
\]

If we assume $p_{\theta}(y|x)$ to be an unbiased prediction, $\tilde{p}_{\theta}(y|x)$
will be the biased prediction associated with the bias in training
data. Therefore, to learn an unbiased classifier $p_{\theta}(y|x)$
given a biased training dataset, BiasBal maximizes the log-likelihood
of $\tilde{p}_{\theta}(y|x)$ (instead of $p_{\theta}(y|x)$) over
the training data. Its training loss is given below:
\[
\Loss_{\theta}^{\text{BiasBal}}:=\Expect_{p_{\mathcal{D}}(x_{n},y_{n})}\left[-\log\tilde{p}_{\theta}(y_{n}|x_{n})\right]
\]

\begin{figure*}
\begin{centering}
\resizebox{\textwidth}{!}{%
\par\end{centering}
\begin{centering}
\begin{tabular}{>{\raggedright}m{0.06\textwidth}ccc}
 & Colored MNIST (0.5\%) & Corrupted CIFAR10 (0.5\%) & Biased CelebA (0.5\%)\tabularnewline
\multirow{1}{0.06\textwidth}[0.175\textwidth]{Test Acc.} & \includegraphics[width=0.4\textwidth]{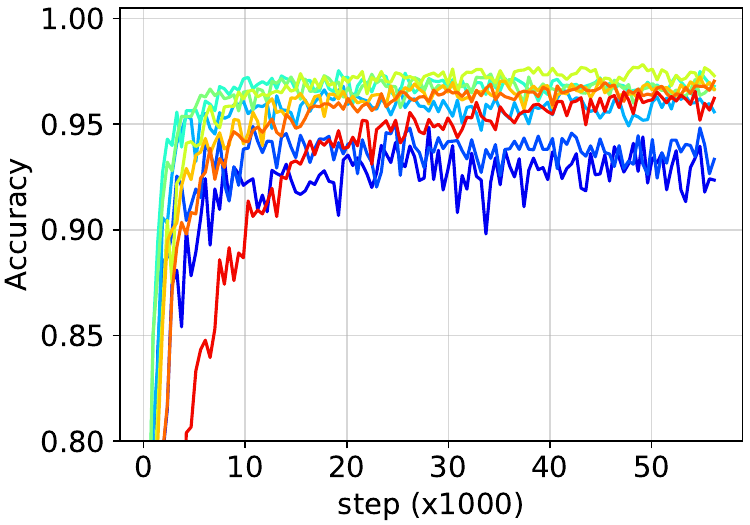} & \includegraphics[width=0.4\textwidth]{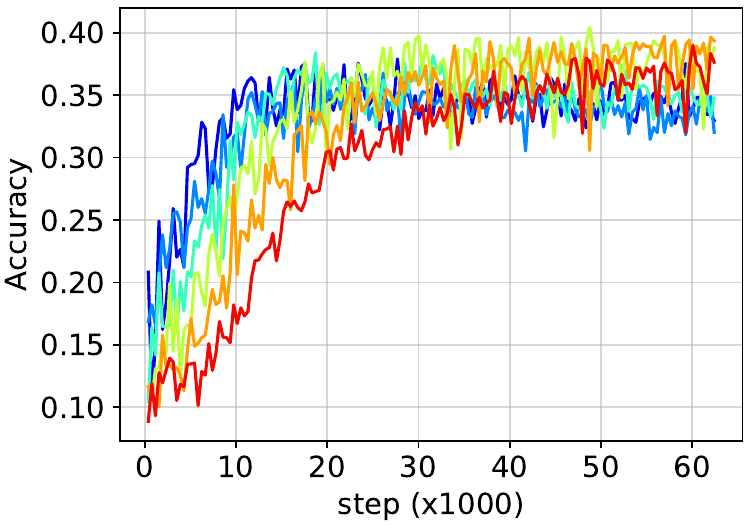} & \includegraphics[width=0.4\textwidth]{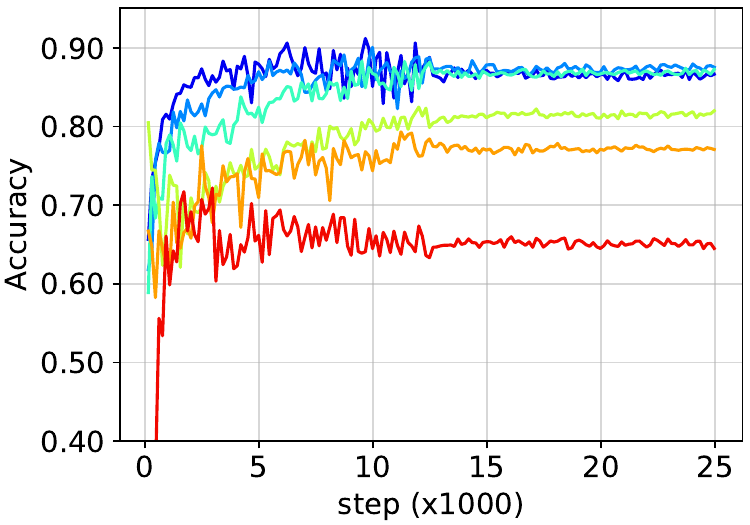}\tabularnewline
\multirow{1}{0.06\textwidth}[0.185\textwidth]{Test Acc. BA} & \includegraphics[width=0.4\textwidth]{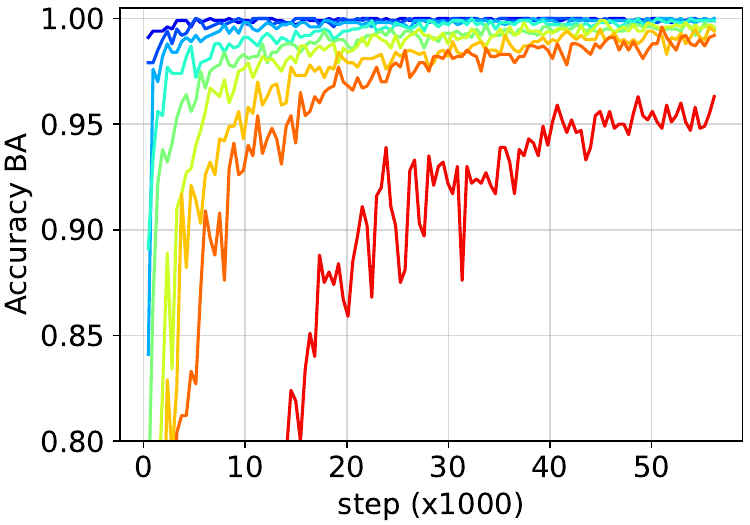} & \includegraphics[width=0.4\textwidth]{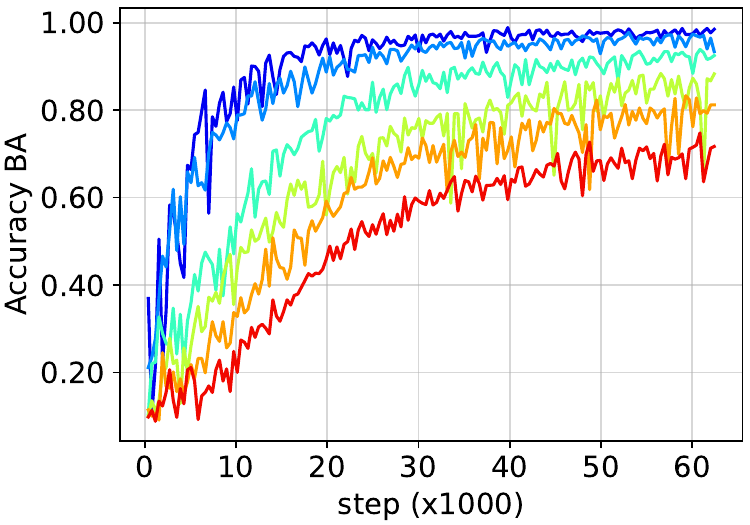} & \includegraphics[width=0.4\textwidth]{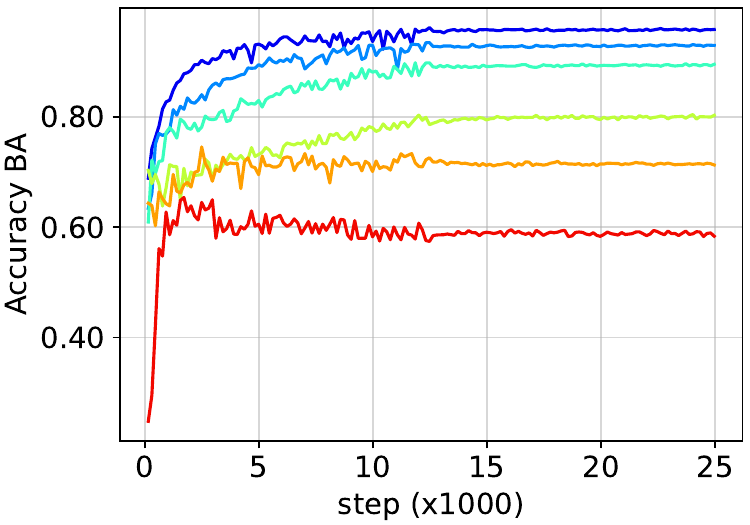}\tabularnewline
\multirow{1}{0.06\textwidth}[0.185\textwidth]{Test Acc. BC} & \includegraphics[width=0.4\textwidth]{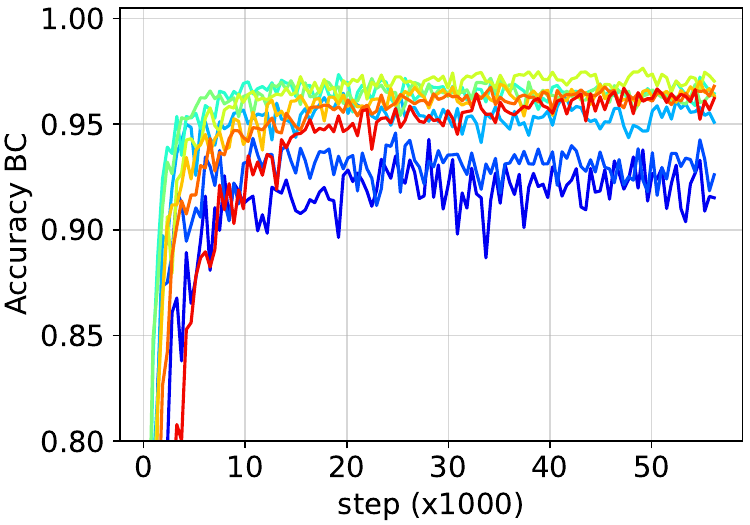} & \includegraphics[width=0.4\textwidth]{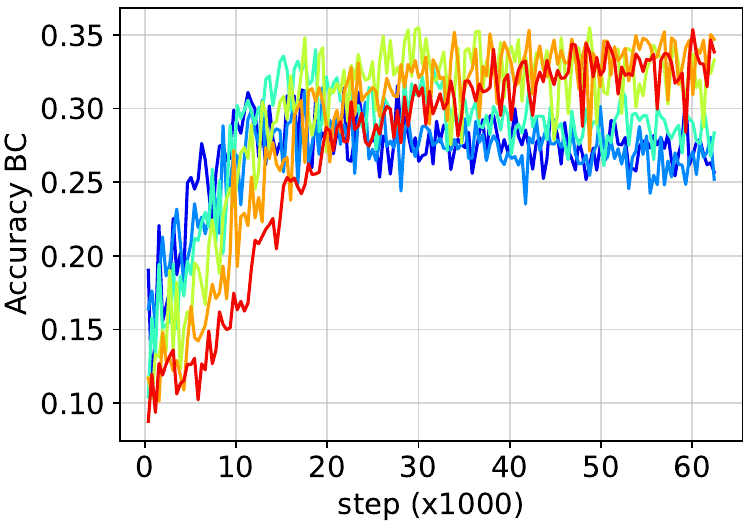} & \includegraphics[width=0.4\textwidth]{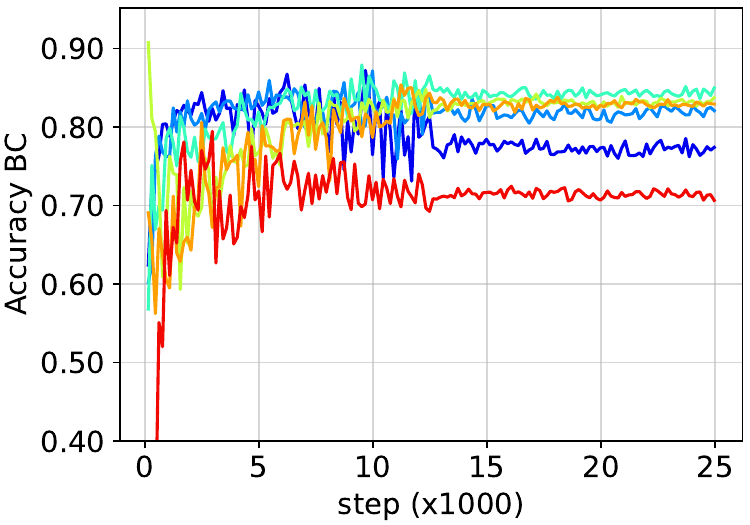}\tabularnewline
 & \includegraphics[width=0.4\textwidth]{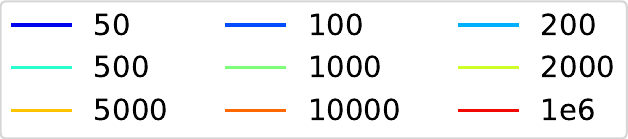} & \includegraphics[width=0.4\textwidth]{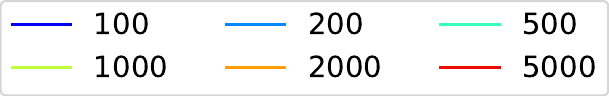} & \includegraphics[width=0.4\textwidth]{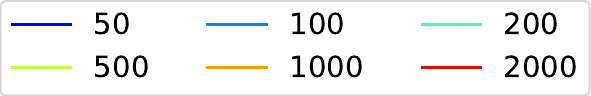}\tabularnewline
\end{tabular}}
\par\end{centering}
\caption{Test curves of Target Bias Adjustment (TBA) - our proposed extension
of BiasBal \cite{Hong2021} to deal with the unavailability of bias
labels - w.r.t. different values of $\gamma$. Considered datasets
are Colored MNIST, Corrupted CIFAR10, and Biased CelebA. All have
BC ratios of 0.5\%.\label{fig:TBA-on-3-datasets}}
\end{figure*}

Originally designed for discrete bias labels, BiasBal depended on
the direct estimation of $p(y|b)$ from training data. To expand its
applicability to unknown or continuous bias labels, we propose approximating
$p(y|b)$ using the technique detailed in Section~\ref{subsec:Bias-mitigation-based-on-p(u|b)}.
This involves training a biased classifier $p_{\psi}(y|x)$ with a
bias amplification loss, viewing it as an approximation of as an approximation
of $p(y|b)$. To ensure stability in computing $\tilde{p}_{\theta}(y|x)$,
$p_{\psi}(y|x)$ is clamped to values greater than $\frac{1}{\gamma}$,
where $\gamma>0$ (as in Eq.~\ref{eq:weight_clamp}). This results
in:
\[
\tilde{p}_{\theta}(y=c|x)=\frac{\exp\left(f_{\theta}(x)[c]+\log v[c]\right)}{\sum_{k=1}^{C}\exp\left(f_{\theta}(x)[k]+\log v[k]\right)}
\]
where $v=\max\left(p_{\psi}(y|x),\frac{1}{\gamma}\right)$. We call
this extension of BiasBal \emph{``Target Bias Adjustment''} (TBA)
and compare it with LW in Table~\ref{tab:Vanilla-TBA-LW}. TBA often
outperforms LW in debiasing, similar to the case where the bias label
is available (Table~\ref{tab:results-with-bias-labels}). This is
because TBA directly corrects the bias of the target $p_{\theta}(y|x)$
while LW does that indirectly as discussed in Section~\ref{subsec:Results-when-the-bias-is-available}.
Similar to LW, the performance of TBA also depends the choice of $\gamma$,
as show in Fig.~\ref{fig:TBA-on-3-datasets}. However, the optimal
values of $\gamma$ differ between the two methods (Fig.~\ref{fig:Diff-sample-weights-LW}
vs. Fig.~\ref{fig:TBA-on-3-datasets}).

\begin{table}
\begin{centering}
\begin{tabular}{ccccc}
\hline 
 & BC (\%) & Vanilla & TBA & LW\tabularnewline
\hline 
\hline 
Colored MNIST & 0.5 & 80.18$\pm$1.38 & \textbf{96.91$\pm$0.52} & 95.57$\pm$0.41\tabularnewline
Corrupted CIFAR10 & 0.5 & 28.00$\pm$1.15 & 38.35$\pm$1.66 & \textbf{45.76$\pm$1.49}\tabularnewline
Biased CelebA & 0.5 & 77.43$\pm$0.42 & \textbf{87.69$\pm$0.35} & 87.18$\pm$0.32\tabularnewline
\hline 
\end{tabular}
\par\end{centering}
\caption{Debiasing results of the vanilla classifier, Target Bias Adjustment
(TBA), and loss weighting (LW) on Colored MNIST, Corrupted CIFAR10,
and Biased CelebA with the BC ratio of 0.5\%. The bias label is assumed
to be unavailable in this case.\label{tab:Vanilla-TBA-LW}}
\end{table}

\section{Bias mitigation based on $p(b|u)$\label{sec:Bias-mitigation-based-on-p(b|u)}}

\subsection{Idea}

\begin{figure*}
\begin{centering}
\begin{tabular}[t]{ccc}
\multirow{2}{*}{\includegraphics[width=0.15\textwidth]{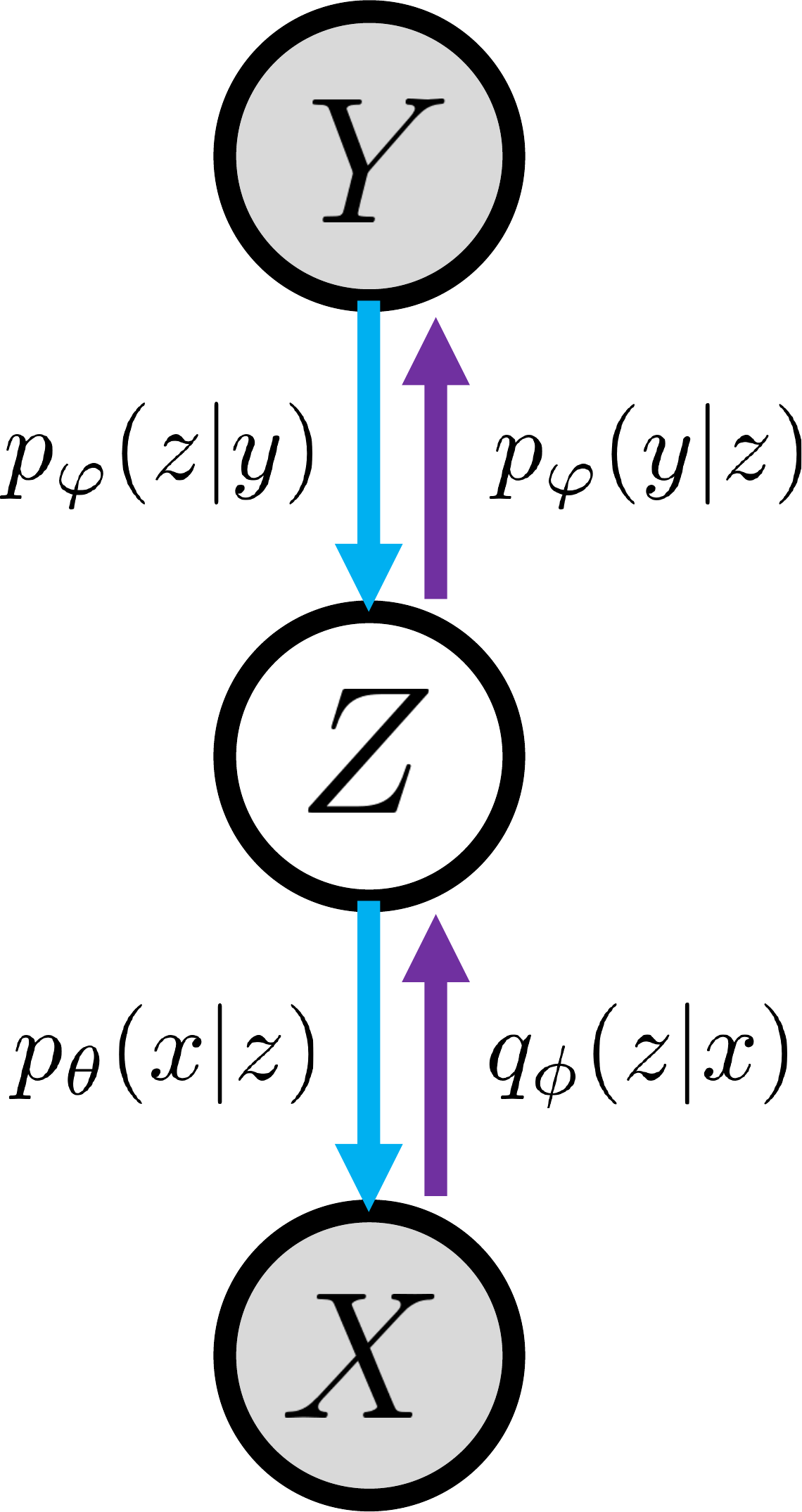}\hspace{0.015\textwidth}} & \includegraphics[width=0.33\textwidth]{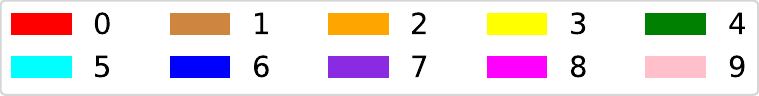} & \multirow{2}{*}{\hspace{0.015\textwidth}\includegraphics[width=0.4\textwidth]{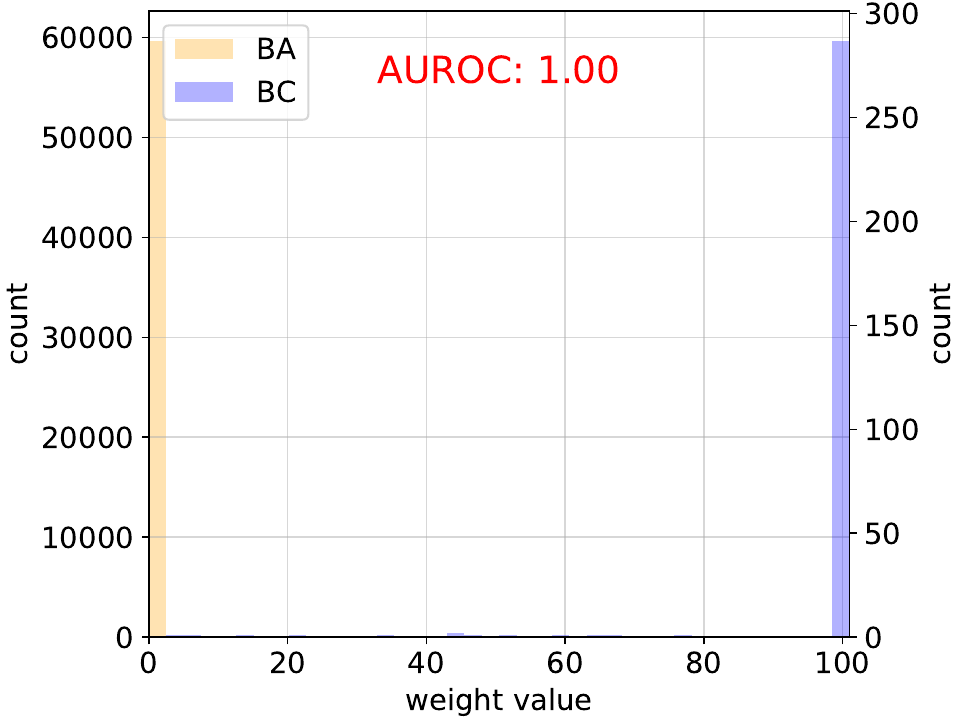}}\tabularnewline
 & \includegraphics[width=0.33\textwidth]{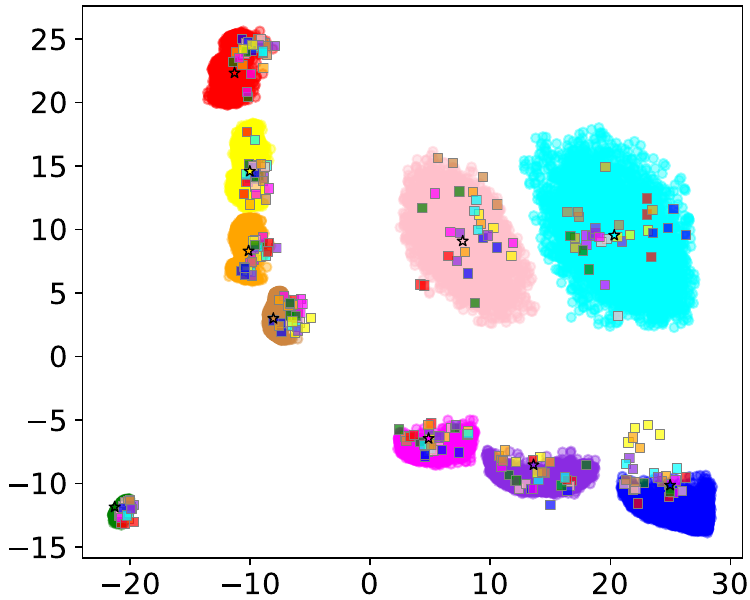} & \tabularnewline
(a) & (b) & (c)\tabularnewline
\end{tabular}
\par\end{centering}
\caption{\textbf{(a)} Graphical representation of the Variational Clustering
AutoEncoder. Observed variables are shaded. \textbf{(b)} Latent space
of a VCAE with $\text{dim}(z)=2$ trained on the biased ColoredMNIST
dataset (bias ratio $=$ 99.5\%). Samples in the figure are colorized
according to their \emph{class labels}. Bias-aligned (BA) and bias-conflicting
(BC) samples are distinguished by round and bordered square markers,
respectively. \textbf{(c)} Histograms of the sample weights computed
via VCAE for BA and BC samples. The formula for the weight $w_{n}$
of a sample $(x_{n},y_{n})$ is $w_{n}$=$\min\left(\frac{1}{p_{\varphi}(y_{n}|z_{n})},100\right)$.\label{fig:VCAE_model}}
\end{figure*}

In this section, we present an idea for mitigating bias by utilizing
$p(b|u)$. We can estimate $p(b|u)$ via its proxy $p(b|y)$. When
the bias attribute $b$ is discrete and the bias label is available,
$p(b|y)$ can be estimated from the training data. However, in the
case where $b$ is continuous or unknown, analytical computation of
$p(b|y)$ become challenging. A possible workaround is computing $p(b_{n},u_{n}|y_{n})$
as an \emph{approximation} of $p(b_{n}|y_{n})$ instead. This stems
from the following equation:
\begin{equation}
p(b_{n},u_{n}|y_{n})=p(b_{n}|y_{n})p(u_{n}|b_{n},y_{n})\propto p(b_{n}|y_{n})\label{eq:approx_of_cond_bias}
\end{equation}
The last equation in Eq.~\ref{eq:approx_of_cond_bias} is based on
the assumption $p(u_{n}|b_{n},y_{n})\approx p(u_{n}|y_{n})\propto1$.
Intuitively, this assumption implies that there is only one class-characteristic
attribute associated with each class. As a result, once we know $y_{n}$,
$u_{n}$ becomes deterministic and the randomness in $(b_{n},u_{n})$
is solely due to the randomness in $b_{n}$. While this assumption
may be strict in general, it allows us to represent $(u_{n},b_{n})$
as a whole instead of extracting the bias attribute $b_{n}$ from
$(u_{n},b_{n})$. If we consider $(u_{n},b_{n})$ as the latent representation
$z_{n}$ of $x_{n}$ rather than $x_{n}$, then $p(u_{n},b_{n}|y_{n})$
becomes $p(z_{n}|y_{n})$. To compute $p(z|y)$, we design a conditional
generative model that encompasses $p(z|y)$ as part of its generative
process. This model is depicted in Fig.~\ref{fig:VCAE_model} (a)
and has the joint distribution factorized as follows:

\begin{equation}
p_{\omega,\varphi}(x,z,y)=p_{\omega}(x|z)p_{\varphi}(z|y)p_{\Data}(y)\label{eq:VCAE_gen_process}
\end{equation}
where $p_{\varphi}(z|y)$ is modeled as an isotropic Gaussian distribution
$\Normal(\mu_{y},\sigma_{y}^{2}\mathrm{I})$ with the \emph{learnable}
mean $\mu_{y}$ and standard deviation $\sigma_{y}$; $p_{\omega}(x|z)$
is modeled via a decoder network $D$ that maps $z$ to $x$. We learn
this generative model by maximizing the \emph{sum} of i) the variational
lower bound (VLB) of $\log p_{\omega,\varphi}(x|y)$ w.r.t. the variational
posterior distribution $q_{\phi}(z|x)$\footnote{We assume that $q_{\phi}(z|x)\approx p_{\theta,\varphi}(z|x,y)$,
i.e., $x$ contains all information of $y$ necessary for predicting
$z$.} and ii) the lower bound of $\log q_{\varphi,\phi}(y|x)$. $q_{\phi}(z|x)$
is modeled as an isotropic Gaussian $\Normal(\mu_{x},\sigma_{x}^{2}\mathrm{I})$
with the mean $\mu_{x}$ and standard deviation $\sigma_{x}$ being
outputs of an encoder network $E$. $q_{\varphi,\phi}(y|x)$ computed
as follows:
\begin{align}
q_{\varphi,\phi}(y|x) & =\int_{z}q_{\varphi,\phi}(y,z|x)\ dz\\
q_{\varphi,\phi}(y,z|x) & =p_{\varphi}(y|z)q_{\phi}(z|x)
\end{align}
where $p_{\varphi}(y|z)$ is derived from $p_{\varphi}(z|y)$ as $p_{\varphi}(y|z)=\frac{p_{\varphi}(z|y)p_{\Data}(y)}{\sum_{c=1}^{C}p_{\varphi}(z|c)p_{\Data}(c)}$\footnote{$p_{\varphi}(z|y)=\frac{p(z,y)}{p(z)}=\frac{p(z,y)}{\sum_{c=1}^{C}p(z,c)}=\frac{p_{\varphi}(z|y)p_{\Data}(y)}{\sum_{c=1}^{C}p_{\varphi}(z|c)p_{\Data}(c)}$}.

We name our proposed generative model \emph{``}\textbf{\emph{V}}\emph{ariational
}\textbf{\emph{C}}\emph{lustering }\textbf{\emph{A}}\emph{uto-}\textbf{\emph{E}}\emph{ncoder''}
(VCAE). The loss function for training VCAE is given below:
\begin{align}
\mathcal{L}_{\omega,\varphi,\phi}^{\text{VCAE}}(x,y):=\  & \lambda_{0}\underbrace{\Expect_{q_{\phi}(z|x)}\left[-\log p_{\omega}(x|z)\right]}_{\text{reconstruction loss w.r.t. }x}+\nonumber \\
 & \lambda_{1}\underbrace{D_{\text{KL}}\left(q_{\phi}(z|x)\|p_{\varphi}(z|y)\right)}_{\text{KL divergence w.r.t. }z}+\nonumber \\
 & \lambda_{2}\underbrace{\Expect_{q_{\phi}(z|x)}\left[-\log p_{\varphi}(y|z)\right]}_{\text{cross-entropy loss w.r.t. }y}\label{eq:VCAE_neg_elbo}
\end{align}
where $\lambda_{0}$, $\lambda_{1}$, $\lambda_{2}$ $\geq0$ are
coefficients. When $\lambda_{0}$=$\lambda_{1}$=$\lambda_{2}$= 1,
the first two terms in Eq.~\ref{eq:VCAE_neg_elbo} form the negative
of the VLB of $\log p_{\omega,\varphi}(x|y)$ while the last term
is the negative of the lower bound of $\log q_{\varphi,\phi}(y|x)$.
We provide the detailed mathematical derivations of these bounds in
Appdx.~\ref{sec:Derivation-of-VLB}. The reconstruction loss encourages
the encoder $E$ to capture both class and non-class attributes in
$x$, the KL divergence forces latent representations of the same
class to form a cluster that somewhat follows the Gaussian distribution,
and the cross-entropy loss ensures that these clusters are well separate. 

Once we have learned the parameters $\omega$, $\varphi$, $\phi$
by minimizing $\Expect_{p_{\Data}(x,y)}\left[\mathcal{L}_{\omega,\varphi,\phi}^{\text{VCAE}}(x,y)\right]$,
given an input-label pair $(x_{n},y_{n})$, we can sample $z_{n}$
from $q_{\phi}(z_{n}|x_{n})$ or simply set it to the mean $\mu_{x_{n}}$
of $q_{\phi}(z_{n}|x_{n})$. Then, we can compute $p_{\varphi}(z_{n}|y_{n})$
easily because it is the density of a Gaussian distribution. Intuitively,
if we view $p_{\varphi}(z|y)$ as the distribution of points in the
cluster centered at $\mu_{y}$, we would expect that the latent representations
$z$ of bias-aligned (BA) samples should \emph{locate near} $\mu_{y}$
and have \emph{large} $p_{\varphi}(z|y)$ as these samples are abundant
in the training data, while those of bias-conflicting (BC) samples
should \emph{be far away from} $\mu_{y}$ and have \emph{small} $p_{\varphi}(z|y)$
due to the scarcity of BC samples. In the extreme case, the latent
representations of BC samples can even be situated in the cluster
of other classes if the BC samples contain non-class attributes $b$
that are highly correlated with those classes. This intuition is clearly
depicted in Fig.~\ref{fig:VCAE_model} (b). From the figure, we can
see that, the learned latent representations of BC samples of, for
example, digit \textcolor{red}{0} (marked as \textcolor{red}{red}
bordered squares) are not located within the cluster w.r.t. digit
\textcolor{red}{0} (the \textcolor{red}{red} cluster) but within the
clusters w.r.t. other digits (clusters with different colors) according
to the background color of these BC samples.

However, although we can explicitly model and compute $p(z|y)$, it
is still not straight-forward to use $p(z|y)$ for weighting. The
main reason is that when $z$ is continuous, $p(z|y)$ is a probabilistic
density function, and thus, is unnormalized (unlike the probability
mass function). Another reason is that $p(z|y)$ becomes exponentially
small when the dimension of $z$ increases, causing $\frac{1}{p(z|y)}$
to be exponentially large. These limitations of $p(z|y)$ are inherent,
implying that we need to find alternative representations of $z$
that are well-normalized and stable. One possible solution to address
this issue is using a collection of discrete latent codes (i.e., a
codebook) to represent the latent space as in discrete VAE \cite{rolfe2016discrete}
or VQ-VAE \cite{van2017neural}. In this case, $p(z|y)$ becomes a
categorical distribution, and thus, is well-normalized and stable.

\subsection{Using $p_{\varphi}(y|z)$ instead of $p_{\varphi}(z|y)$}

\begin{figure*}
\begin{centering}
\resizebox{\textwidth}{!}{%
\par\end{centering}
\begin{centering}
\begin{tabular}{cc}
\multicolumn{2}{c}{\includegraphics[width=0.33\textwidth]{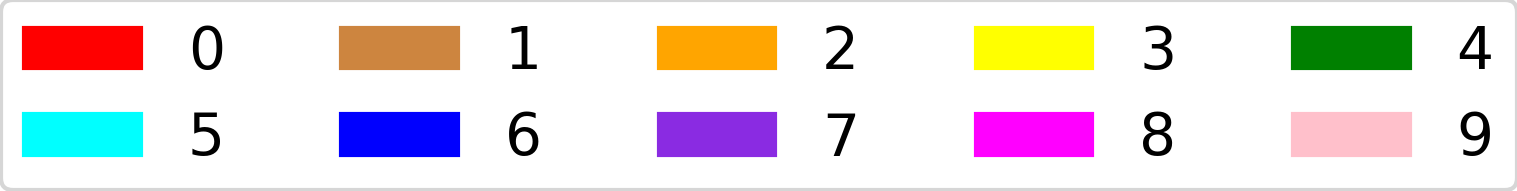}}\tabularnewline
\includegraphics[width=0.5\textwidth]{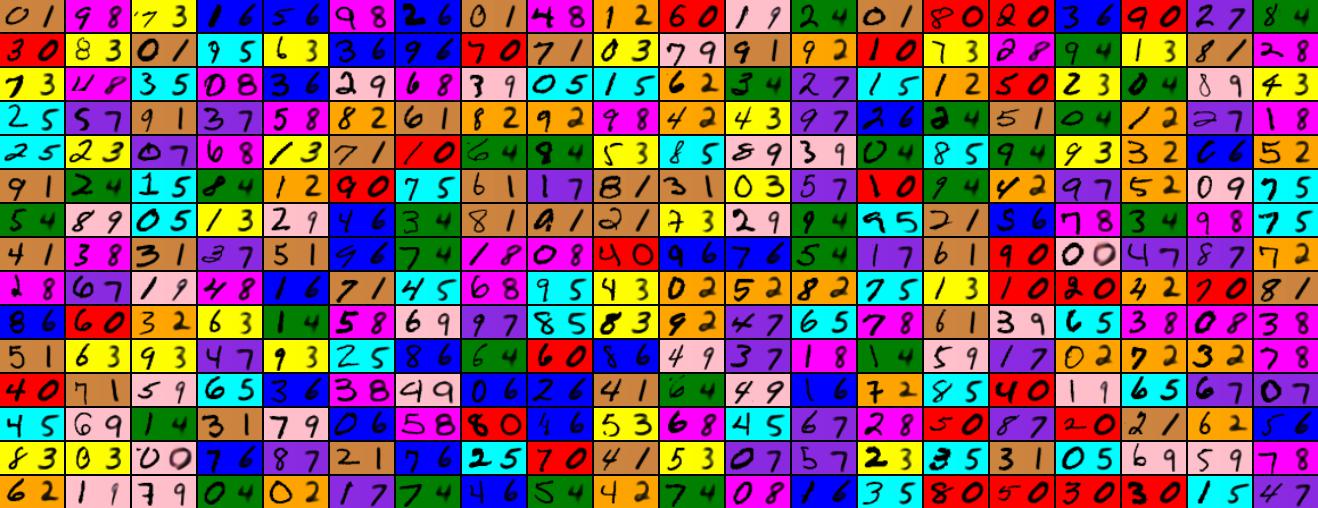} & \includegraphics[width=0.5\textwidth]{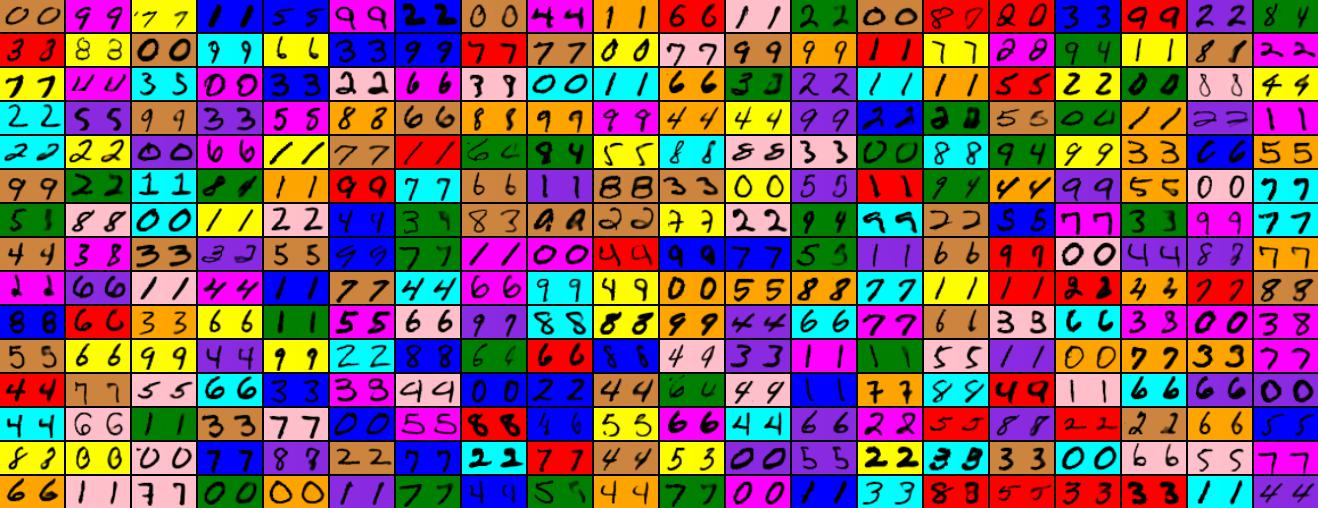}\tabularnewline
(a) dim($z$)=2 & (b) dim($z$)=100\tabularnewline
\end{tabular}}
\par\end{centering}
\caption{Pairs of original and reconstructed test BC images on Colored MNIST
produced by VCAE. It is obvious that when dim($z$) = 2, the latent
variable $z$ captures only the bias attribute (i.e., background color)
since most of the reconstructed images contain incorrect class information.\label{fig:rec_x_VCAE}}
\end{figure*}

We may consider using $p_{\varphi}(y|z)$ for weighting instead of
$p_{\varphi}(z|y)$ since $p_{\varphi}(y|z)$ is well-normalized and
stable. However, this technique is no longer associated with $p(b|u)$
but $p(u|b)$ (Section~\ref{subsec:Bias-mitigation-based-on-p(u|b)}).
It means that $z$, as a representation of $x$, should capture only
the bias information $b$ in order to achieve good debiasing performance.
In our experiments, we empirically observed that using $p(y|z)$ for
bias mitigation, with $p(y|z)$ being modeled via a VCAE, \emph{only}
performs well on Colored MNIST when the VCAE has a small enough latent
vector (e.g., dim($z$) < 10) and a large enough KL divergence coefficient
(e.g., $\lambda_{1}\approx1$) (Table~\ref{tab:Results-of-VCAE}
and Fig.~\ref{fig:rec_x_VCAE}). This is because on Colored MNIST,
the bias attribute (i.e., background color) accounts for most information
in the input $x$, yet can be captured easily. Therefore, by using
a small latent vector and a large KL divergence, VCAE is forced to
learn a latent representation $z$ that focuses on capturing $b$
rather than $u$ in order to achieve a better reconstruction of $x$.

\begin{table*}
\begin{centering}
\begin{tabular}{>{\raggedright}p{0.08\textwidth}cccccc}
\hline 
\multirow{2}{0.08\textwidth}{Dataset} & \multirow{2}{*}{BC (\%)} & \multirow{2}{*}{Vanilla} & \multirow{2}{*}{LfF} & \multirow{2}{*}{PGD} & \multicolumn{2}{c}{LW}\tabularnewline
\cline{6-7} 
 &  &  &  &  & $p(y|b)$ & $p(y|z)$\tabularnewline
\hline 
\hline 
\multirow{3}{0.08\textwidth}{Colored MNIST} & 0.5 & 80.18$\pm$1.38 & 93.38$\pm$0.52 & \textcolor{gray}{96.15$\pm$0.28} & 95.57$\pm$0.41 & \textbf{96.85$\pm$0.16}\tabularnewline
 & 1.0 & 87.48$\pm$1.75 & 94.09$\pm$0.78 & \textcolor{gray}{97.93$\pm$0.19} & 97.18$\pm$0.34 & \textbf{97.93$\pm$0.12}\tabularnewline
 & 5.0 & 97.04$\pm$0.21 & 97.40$\pm$0.25 & \textbf{98.74$\pm$0.12} & \textcolor{gray}{98.61$\pm$0.09} & 98.18$\pm$0.11\tabularnewline
\hline 
\end{tabular}
\par\end{centering}
\caption{Results of LW with the sample weight computed based on $p(y|z)$ which
is modeled by a VCAE in comparison with other baselines on Colored
MNIST. Regarding the VCAE's settings, $\text{dim}(z)=2$ and $(\lambda_{0},\lambda_{1},\lambda_{2})$
is set to (1, 1, 1).\label{tab:Results-of-VCAE}}
\end{table*}

\subsection{Derivations of the lower bounds of $\log p_{\theta,\varphi}(x|y)$
and $\log q_{\varphi,\phi}(y|x)$\label{sec:Derivation-of-VLB}}

Given that $p_{\theta,\varphi}(x,z,y)=p_{\theta}(x|z)p_{\varphi}(z|y)p_{\Data}(y)$
and $q_{\phi}(z|x)$ is the variational approximation of $p_{\theta,\varphi}(z|x,y)$,
we can compute the variational lower bound of $p_{\theta,\varphi}(x|y)$
as follows:
\begin{align}
 & \log p_{\theta,\varphi}(x|y)\nonumber \\
=\  & \Expect_{q_{\phi}(z|x)}\left[\log p_{\theta,\varphi}(x|y)\right]\\
=\  & \Expect_{q_{\phi}(z|x)}\left[\log\frac{p_{\theta,\varphi}(x,z|y)}{p_{\theta,\varphi}(z|x,y)}\right]\\
=\  & \Expect_{q_{\phi}(z|x)}\left[\log\frac{p_{\theta,\varphi}(x,z|y)}{q_{\phi}(z|x)}\right]+\Expect_{q_{\phi}(z|x)}\left[\log\frac{q_{\phi}(z|x)}{p_{\theta,\varphi}(z|x,y)}\right]\\
=\  & \Expect_{q_{\phi}(z|x)}\left[\log\frac{p_{\theta,\varphi}(x,z|y)}{q_{\phi}(z|x)}\right]+D_{\text{KL}}\left(q_{\phi}(z|x)\|p_{\theta,\varphi}(z|x,y)\right)\\
\geq\  & \Expect_{q_{\phi}(z|x)}\left[\log\frac{p_{\theta,\varphi}(x,z|y)}{q_{\phi}(z|x)}\right]\label{eq:VLB_p_x_cond_y}\\
=\  & \Expect_{q_{\phi}(z|x)}\left[\log\frac{p_{\theta}(x|z)p_{\varphi}(z|y)}{q_{\phi}(z|x)}\right]\\
=\  & \Expect_{q_{\phi}(z|x)}\left[\log p_{\theta}(x|z)+\log p_{\varphi}(z|y)-\log q_{\phi}(z|x)\right]\\
=\  & \Expect_{q_{\phi}(z|x)}\left[\log p_{\theta}(x|z)\right]-\Expect_{q_{\phi}(z|x)}\left[\log q_{\phi}(z|x)-\log p_{\varphi}(z|y)\right]\\
=\  & \Expect_{q_{\phi}(z|x)}\left[\log p_{\theta}(x|z)\right]-D_{\text{KL}}\left(q_{\phi}(z|x)\|p_{\varphi}(z|y)\right)
\end{align}
where the inequality in Eq.~\ref{eq:VLB_p_x_cond_y} is due to the
fact that $D_{\text{KL}}\left(q_{\phi}(z|x)\|p_{\theta,\varphi}(z|x,y)\right)\geq0$.

We can also derive the lower bound of $\log q_{\varphi,\phi}(y|x)$
as follows:
\begin{align}
\log q_{\varphi,\phi}(y|x) & =\log\left(\int_{z}q_{\varphi,\phi}(y,z|x)\ dz\right)\\
 & =\log\left(\int_{z}p_{\varphi}(y|z)q_{\phi}(z|x)\ dz\right)\\
 & =\log\Expect_{q_{\phi}(z|x)}\left[p_{\varphi}(y|z)\right]\\
 & \geq\Expect_{q_{\phi}(z|x)}\left[\log p_{\varphi}(y|z)\right]\label{eq:LB_q_y_cond_x_4}
\end{align}
where the inequality in Eq.~\ref{eq:LB_q_y_cond_x_4} is the Jensen's
inequality for the concave log function.


\begin{thebibliography}{}

\end{thebibliography}


\begin{thebibliography}{10}

\bibitem{agrawal2016analyzing}
Aishwarya Agrawal, Dhruv Batra, and Devi Parikh.
\newblock Analyzing the behavior of visual question answering models.
\newblock In {\em Proceedings of the 2016 Conference on Empirical Methods in
  Natural Language Processing}, pages 1955--1960, 2016.

\bibitem{agrawal2018don}
Aishwarya Agrawal, Dhruv Batra, Devi Parikh, and Aniruddha Kembhavi.
\newblock Dont just assume; look and answer: Overcoming priors for visual
  question answering.
\newblock In {\em Proceedings of the IEEE conference on computer vision and
  pattern recognition}, pages 4971--4980, 2018.

\bibitem{Ahn2023}
Sumyeong Ahn, Seongyoon Kim, and Se-young Yun.
\newblock Mitigating dataset bias by using per-sample gradient.
\newblock {\em International Conference on Learning Representations}, 2023.

\bibitem{Alvi2018}
Mohsan Alvi, Andrew Zisserman, and Christoffer Nell{\aa}ker.
\newblock Turning a blind eye: Explicit removal of biases and variation from
  deep neural network embeddings.
\newblock In {\em Proceedings of the European Conference on Computer Vision
  (ECCV) Workshops}, pages 0--0, 2018.

\bibitem{Bahng2020}
Hyojin Bahng, Sanghyuk Chun, Sangdoo Yun, Jaegul Choo, and Seong~Joon Oh.
\newblock Learning de-biased representations with biased representations.
\newblock In {\em International Conference on Machine Learning}, pages
  528--539. PMLR, 2020.

\bibitem{BenTal2013}
Aharon Ben-Tal, Dick den Hertog, Anja De~Waegenaere, Bertrand Melenberg, and
  Gijs Rennen.
\newblock Robust solutions of optimization problems affected by uncertain
  probabilities.
\newblock {\em Management Science}, 59(2):341--357, 2013.

\bibitem{buda2018systematic}
Mateusz Buda, Atsuto Maki, and Maciej~A Mazurowski.
\newblock A systematic study of the class imbalance problem in convolutional
  neural networks.
\newblock {\em Neural networks}, 106:249--259, 2018.

\bibitem{Cadene2019}
Remi Cadene, Corentin Dancette, Matthieu Cord, Devi Parikh, et~al.
\newblock Rubi: Reducing unimodal biases for visual question answering.
\newblock {\em Advances in neural information processing systems}, 32, 2019.

\bibitem{cao2019learning}
Kaidi Cao, Colin Wei, Adrien Gaidon, Nikos Arechiga, and Tengyu Ma.
\newblock Learning imbalanced datasets with label-distribution-aware margin
  loss.
\newblock {\em Advances in neural information processing systems}, 32, 2019.

\bibitem{chawla2002smote}
Nitesh~V Chawla, Kevin~W Bowyer, Lawrence~O Hall, and W~Philip Kegelmeyer.
\newblock Smote: synthetic minority over-sampling technique.
\newblock {\em Journal of artificial intelligence research}, 16:321--357, 2002.

\bibitem{clark2019don}
Christopher Clark, Mark Yatskar, and Luke Zettlemoyer.
\newblock Dont take the easy way out: Ensemble based methods for avoiding known
  dataset biases.
\newblock In {\em Proceedings of the 2019 Conference on Empirical Methods in
  Natural Language Processing and the 9th International Joint Conference on
  Natural Language Processing (EMNLP-IJCNLP)}, pages 4069--4082, 2019.

\bibitem{cui2019class}
Yin Cui, Menglin Jia, Tsung-Yi Lin, Yang Song, and Serge Belongie.
\newblock Class-balanced loss based on effective number of samples.
\newblock In {\em Proceedings of the IEEE/CVF conference on computer vision and
  pattern recognition}, pages 9268--9277, 2019.

\bibitem{Duchi2016}
John Duchi, Peter Glynn, and Hongseok Namkoong.
\newblock Statistics of robust optimization: A generalized empirical likelihood
  approach.
\newblock {\em arXiv preprint arXiv:1610.03425}, 2016.

\bibitem{elkan2001foundations}
Charles Elkan.
\newblock The foundations of cost-sensitive learning.
\newblock In {\em International joint conference on artificial intelligence},
  volume~17, pages 973--978. Lawrence Erlbaum Associates Ltd, 2001.

\bibitem{estabrooks2004multiple}
Andrew Estabrooks, Taeho Jo, and Nathalie Japkowicz.
\newblock A multiple resampling method for learning from imbalanced data sets.
\newblock {\em Computational intelligence}, 20(1):18--36, 2004.

\bibitem{ganin2016domain}
Yaroslav Ganin, Evgeniya Ustinova, Hana Ajakan, Pascal Germain, Hugo
  Larochelle, Fran{\c{c}}ois Laviolette, Mario Marchand, and Victor Lempitsky.
\newblock Domain-adversarial training of neural networks.
\newblock {\em Journal of Machine Learning Research}, 17:1--35, 2016.

\bibitem{geirhos2018imagenet}
Robert Geirhos, Patricia Rubisch, Claudio Michaelis, Matthias Bethge, Felix~A
  Wichmann, and Wieland Brendel.
\newblock Imagenet-trained cnns are biased towards texture; increasing shape
  bias improves accuracy and robustness.
\newblock {\em arXiv preprint arXiv:1811.12231}, 2018.

\bibitem{he2009learning}
Haibo He and Edwardo~A Garcia.
\newblock Learning from imbalanced data.
\newblock {\em IEEE Transactions on Knowledge and Data Engineering},
  21(9):1263--1284, 2009.

\bibitem{he2016deep}
Kaiming He, Xiangyu Zhang, Shaoqing Ren, and Jian Sun.
\newblock Deep residual learning for image recognition.
\newblock In {\em Proceedings of the IEEE conference on computer vision and
  pattern recognition}, pages 770--778, 2016.

\bibitem{hirano2001estimation}
Keisuke Hirano and Guido~W Imbens.
\newblock Estimation of causal effects using propensity score weighting: An
  application to data on right heart catheterization.
\newblock {\em Health Services and Outcomes research methodology}, 2:259--278,
  2001.

\bibitem{hirano2003efficient}
Keisuke Hirano, Guido~W Imbens, and Geert Ridder.
\newblock Efficient estimation of average treatment effects using the estimated
  propensity score.
\newblock {\em Econometrica}, 71(4):1161--1189, 2003.

\bibitem{Hong2021}
Youngkyu Hong and Eunho Yang.
\newblock Unbiased classification through bias-contrastive and bias-balanced
  learning.
\newblock {\em Advances in Neural Information Processing Systems},
  34:26449--26461, 2021.

\bibitem{Hu2018}
Weihua Hu, Gang Niu, Issei Sato, and Masashi Sugiyama.
\newblock Does distributionally robust supervised learning give robust
  classifiers?
\newblock In {\em International Conference on Machine Learning}, pages
  2029--2037. PMLR, 2018.

\bibitem{Hwang2022}
Inwoo Hwang, Sangjun Lee, Yunhyeok Kwak, Seong~Joon Oh, Damien Teney, Jin-Hwa
  Kim, and Byoung-Tak Zhang.
\newblock Selecmix: Debiased learning by contradicting-pair sampling.
\newblock In {\em Proceedings of the 36th Advances in Neural Information
  Processing Systems}, 2022.

\bibitem{ioffe2015batch}
Sergey Ioffe and Christian Szegedy.
\newblock Batch normalization: Accelerating deep network training by reducing
  internal covariate shift.
\newblock In {\em International conference on machine learning}, pages
  448--456. pmlr, 2015.

\bibitem{khan2019striking}
Salman Khan, Munawar Hayat, Syed~Waqas Zamir, Jianbing Shen, and Ling Shao.
\newblock Striking the right balance with uncertainty.
\newblock In {\em Proceedings of the IEEE/CVF Conference on Computer Vision and
  Pattern Recognition}, pages 103--112, 2019.

\bibitem{khan2017cost}
Salman~H Khan, Munawar Hayat, Mohammed Bennamoun, Ferdous~A Sohel, and Roberto
  Togneri.
\newblock Cost-sensitive learning of deep feature representations from
  imbalanced data.
\newblock {\em IEEE transactions on neural networks and learning systems},
  29(8):3573--3587, 2017.

\bibitem{Khosla2012}
Aditya Khosla, Tinghui Zhou, Tomasz Malisiewicz, Alexei~A Efros, and Antonio
  Torralba.
\newblock Undoing the damage of dataset bias.
\newblock In {\em Computer Vision--ECCV 2012: 12th European Conference on
  Computer Vision, Florence, Italy, October 7-13, 2012, Proceedings, Part I
  12}, pages 158--171. Springer, 2012.

\bibitem{Kim2019}
Byungju Kim, Hyunwoo Kim, Kyungsu Kim, Sungjin Kim, and Junmo Kim.
\newblock Learning not to learn: Training deep neural networks with biased
  data.
\newblock In {\em Proceedings of the IEEE/CVF Conference on Computer Vision and
  Pattern Recognition}, pages 9012--9020, 2019.

\bibitem{Kim2021}
Eungyeup Kim, Jihyeon Lee, and Jaegul Choo.
\newblock Biaswap: Removing dataset bias with bias-tailored swapping
  augmentation.
\newblock In {\em Proceedings of the IEEE/CVF International Conference on
  Computer Vision}, pages 14992--15001, 2021.

\bibitem{Kim2022}
Nayeong Kim, Sehyun Hwang, Sungsoo Ahn, Jaesik Park, and Suha Kwak.
\newblock Learning debiased classifier with biased committee.
\newblock In {\em Proceedings of the 36th Advances in Neural Information
  Processing Systems}, 2022.

\bibitem{krizhevsky2009learning}
Alex Krizhevsky, Geoffrey Hinton, et~al.
\newblock Learning multiple layers of features from tiny images.
\newblock 2009.

\bibitem{kukar1998cost}
Matjaz Kukar, Igor Kononenko, et~al.
\newblock Cost-sensitive learning with neural networks.
\newblock In {\em ECAI}, volume~15, pages 88--94, 1998.

\bibitem{lawrence2012neural}
Steve Lawrence, Ian Burns, Andrew Back, Ah~Chung Tsoi, and C~Lee Giles.
\newblock Neural network classification and prior class probabilities.
\newblock {\em Neural Networks: Tricks of the Trade: Second Edition}, pages
  295--309, 2012.

\bibitem{LeBras2020}
Ronan Le~Bras, Swabha Swayamdipta, Chandra Bhagavatula, Rowan Zellers, Matthew
  Peters, Ashish Sabharwal, and Yejin Choi.
\newblock Adversarial filters of dataset biases.
\newblock In {\em International conference on machine learning}, pages
  1078--1088. PMLR, 2020.

\bibitem{lecun2010mnist}
Yann LeCun, Corinna Cortes, Chris Burges, et~al.
\newblock Mnist handwritten digit database, 2010.

\bibitem{Lee2021}
Jungsoo Lee, Eungyeup Kim, Juyoung Lee, Jihyeon Lee, and Jaegul Choo.
\newblock Learning debiased representation via disentangled feature
  augmentation.
\newblock {\em Advances in Neural Information Processing Systems},
  34:25123--25133, 2021.

\bibitem{Lee2023}
Jungsoo Lee, Jeonghoon Park, Daeyoung Kim, Juyoung Lee, Edward Choi, and Jaegul
  Choo.
\newblock Biasensemble: Revisiting the importance of amplifying bias for
  debiasing.
\newblock {\em Proceedings of the 37th AAAI Conference on Artificial
  Intelligence}, 2023.

\bibitem{Li2019}
Yi~Li and Nuno Vasconcelos.
\newblock Repair: Removing representation bias by dataset resampling.
\newblock In {\em Proceedings of the IEEE/CVF conference on computer vision and
  pattern recognition}, pages 9572--9581, 2019.

\bibitem{Li2018}
Yingwei Li, Yi~Li, and Nuno Vasconcelos.
\newblock Resound: Towards action recognition without representation bias.
\newblock In {\em Proceedings of the European Conference on Computer Vision
  (ECCV)}, pages 513--528, 2018.

\bibitem{lin2017focal}
Tsung-Yi Lin, Priya Goyal, Ross Girshick, Kaiming He, and Piotr Doll{\'a}r.
\newblock Focal loss for dense object detection.
\newblock In {\em Proceedings of the IEEE international conference on computer
  vision}, pages 2980--2988, 2017.

\bibitem{liu2021just}
Evan~Z Liu, Behzad Haghgoo, Annie~S Chen, Aditi Raghunathan, Pang~Wei Koh,
  Shiori Sagawa, Percy Liang, and Chelsea Finn.
\newblock Just train twice: Improving group robustness without training group
  information.
\newblock In {\em International Conference on Machine Learning}, pages
  6781--6792. PMLR, 2021.

\bibitem{liu2015faceattributes}
Ziwei Liu, Ping Luo, Xiaogang Wang, and Xiaoou Tang.
\newblock Deep learning face attributes in the wild.
\newblock In {\em Proceedings of International Conference on Computer Vision
  (ICCV)}, December 2015.

\bibitem{lu2021invariant}
Chaochao Lu, Yuhuai Wu, Jos{\'e}~Miguel Hern{\'a}ndez-Lobato, and Bernhard
  Sch{\"o}lkopf.
\newblock Invariant causal representation learning for out-of-distribution
  generalization.
\newblock In {\em International Conference on Learning Representations}, 2021.

\bibitem{mitrovic2020representation}
Jovana Mitrovic, Brian McWilliams, Jacob Walker, Lars Buesing, and Charles
  Blundell.
\newblock Representation learning via invariant causal mechanisms.
\newblock {\em arXiv preprint arXiv:2010.07922}, 2020.

\bibitem{Nam2020}
Junhyun Nam, Hyuntak Cha, Sungsoo Ahn, Jaeho Lee, and Jinwoo Shin.
\newblock Learning from failure: De-biasing classifier from biased classifier.
\newblock {\em Advances in Neural Information Processing Systems},
  33:20673--20684, 2020.

\bibitem{nguyen2022front}
Toan Nguyen, Kien Do, Duc~Thanh Nguyen, Bao Duong, and Thin Nguyen.
\newblock Front-door adjustment via style transfer for out-of-distribution
  generalisation.
\newblock {\em arXiv preprint arXiv:2212.03063}, 2022.

\bibitem{Park2020}
Taesung Park, Jun-Yan Zhu, Oliver Wang, Jingwan Lu, Eli Shechtman, Alexei
  Efros, and Richard Zhang.
\newblock Swapping autoencoder for deep image manipulation.
\newblock {\em Advances in Neural Information Processing Systems},
  33:7198--7211, 2020.

\bibitem{pearl2000models}
Judea Pearl et~al.
\newblock Models, reasoning and inference.
\newblock {\em Cambridge, UK: CambridgeUniversityPress}, 19(2):3, 2000.

\bibitem{ramakrishnan2018overcoming}
Sainandan Ramakrishnan, Aishwarya Agrawal, and Stefan Lee.
\newblock Overcoming language priors in visual question answering with
  adversarial regularization.
\newblock {\em Advances in Neural Information Processing Systems}, 31, 2018.

\bibitem{rolfe2016discrete}
Jason~Tyler Rolfe.
\newblock Discrete variational autoencoders.
\newblock {\em arXiv preprint arXiv:1609.02200}, 2016.

\bibitem{rubin1974estimating}
Donald~B Rubin.
\newblock Estimating causal effects of treatments in randomized and
  nonrandomized studies.
\newblock {\em Journal of educational Psychology}, 66(5):688, 1974.

\bibitem{sagawa2020distributionally}
Shiori Sagawa, Pang~Wei Koh, Tatsunori~B Hashimoto, and Percy Liang.
\newblock Distributionally robust neural networks for group shifts: On the
  importance of regularization for worst-case generalization.
\newblock 2020.

\bibitem{Shrestha2022}
Robik Shrestha, Kushal Kafle, and Christopher Kanan.
\newblock An investigation of critical issues in bias mitigation techniques.
\newblock In {\em Proceedings of the IEEE/CVF Winter Conference on Applications
  of Computer Vision}, pages 1943--1954, 2022.

\bibitem{Tartaglione2021}
Enzo Tartaglione, Carlo~Alberto Barbano, and Marco Grangetto.
\newblock End: Entangling and disentangling deep representations for bias
  correction.
\newblock In {\em Proceedings of the IEEE/CVF conference on computer vision and
  pattern recognition}, pages 13508--13517, 2021.

\bibitem{Torralba2011}
A~Torralba and AA~Efros.
\newblock Unbiased look at dataset bias.
\newblock In {\em Proceedings of the 2011 IEEE Conference on Computer Vision
  and Pattern Recognition}, pages 1521--1528, 2011.

\bibitem{van2017neural}
Aaron Van Den~Oord, Oriol Vinyals, et~al.
\newblock Neural discrete representation learning.
\newblock {\em Advances in neural information processing systems}, 30, 2017.

\bibitem{verma2019manifold}
Vikas Verma, Alex Lamb, Christopher Beckham, Amir Najafi, Ioannis Mitliagkas,
  David Lopez-Paz, and Yoshua Bengio.
\newblock Manifold mixup: Better representations by interpolating hidden
  states.
\newblock In {\em International conference on machine learning}, pages
  6438--6447. PMLR, 2019.

\bibitem{wang2019learning}
Haohan Wang, Eric~P Xing, Zexue He, and Zachary~C Lipton.
\newblock Learning robust representations by projecting superficial statistics
  out.
\newblock In {\em International Conference on Learning Representations, ICLR
  2019}, 2019.

\bibitem{wang2022out}
Ruoyu Wang, Mingyang Yi, Zhitang Chen, and Shengyu Zhu.
\newblock Out-of-distribution generalization with causal invariant
  transformations.
\newblock In {\em Proceedings of the IEEE/CVF Conference on Computer Vision and
  Pattern Recognition}, pages 375--385, 2022.

\bibitem{yao2021survey}
Liuyi Yao, Zhixuan Chu, Sheng Li, Yaliang Li, Jing Gao, and Aidong Zhang.
\newblock A survey on causal inference.
\newblock {\em ACM Transactions on Knowledge Discovery from Data (TKDD)},
  15(5):1--46, 2021.

\bibitem{yue2021transporting}
Zhongqi Yue, Qianru Sun, Xian-Sheng Hua, and Hanwang Zhang.
\newblock Transporting causal mechanisms for unsupervised domain adaptation.
\newblock In {\em Proceedings of the IEEE/CVF International Conference on
  Computer Vision}, pages 8599--8608, 2021.

\bibitem{zhang2022correct}
Michael Zhang, Nimit~S Sohoni, Hongyang~R Zhang, Chelsea Finn, and Christopher
  Re.
\newblock Correct-n-contrast: a contrastive approach for improving robustness
  to spurious correlations.
\newblock In {\em International Conference on Machine Learning}, pages
  26484--26516. PMLR, 2022.

\bibitem{zhang2018generalized}
Zhilu Zhang and Mert Sabuncu.
\newblock Generalized cross entropy loss for training deep neural networks with
  noisy labels.
\newblock {\em Advances in neural information processing systems}, 31, 2018.

\bibitem{zhou2016learning}
Bolei Zhou, Aditya Khosla, Agata Lapedriza, Aude Oliva, and Antonio Torralba.
\newblock Learning deep features for discriminative localization.
\newblock In {\em Proceedings of the IEEE conference on computer vision and
  pattern recognition}, pages 2921--2929, 2016.

\bibitem{Zhu2021}
Wei Zhu, Haitian Zheng, Haofu Liao, Weijian Li, and Jiebo Luo.
\newblock Learning bias-invariant representation by cross-sample mutual
  information minimization.
\newblock In {\em Proceedings of the IEEE/CVF International Conference on
  Computer Vision}, pages 15002--15012, 2021.

\end{thebibliography}
\end{document}